\documentclass[runningheads]{llncs}

\usepackage{eccv}

\usepackage{eccvabbrv}

\usepackage{graphicx}
\usepackage{booktabs}

\usepackage{booktabs}
\usepackage{float}
\usepackage{caption}
\usepackage{subcaption}
\usepackage{etoolbox}
\usepackage{multirow}
\usepackage{multicol}
\usepackage{rotating}
\usepackage{adjustbox}
\usepackage{nicefrac, xfrac}
\usepackage{wrapfig}
\usepackage{placeins}

\usepackage{wrapfig} 

\AtEndEnvironment{section}{\vspace{0.0cm}}
\AtEndEnvironment{table}{\vspace{0.0cm}}
\AtEndEnvironment{figure}{\vspace{0.0cm}} 
\AtEndEnvironment{figure*}{\vspace{0.0cm}} 
\usepackage{colortbl}
\usepackage[font=scriptsize,skip=0pt]{caption}
\usepackage{comment}
\newcommand{\tcolor}[1]{\textcolor{magenta}{#1}}

\usepackage{hyperref}

\usepackage{lmodern}

\begin{document}

\title{\tcolor{ViscoNet}: Bridging and Harmonizing \tcolor{Vis}ual and Textual Conditioning for \tcolor{Co}ntrol\tcolor{Net}}

\titlerunning{ViscoNet}

\author{Soon Yau Cheong \and
Armin Mustafa \and
Andrew Gilbert}

\authorrunning{SY.Cheong et al.}

\institute{\email{s.cheong, armin.mustafa, a.gilbert@surrey.ac.uk} \\
University of Surrey, UK \\}

\maketitle

\begin{abstract}
This paper introduces ViscoNet, a novel one-branch-adapter architecture for concurrent spatial and visual conditioning. Our lightweight model requires trainable parameters and dataset size multiple orders of magnitude smaller than the current state-of-the-art IP-Adapter. However, our method successfully preserves the generative power of the frozen text-to-image (T2I) backbone. Notably, it excels in addressing mode collapse, a pervasive issue previously overlooked. Our novel architecture demonstrates outstanding capabilities in achieving a harmonious visual-text balance, unlocking unparalleled versatility in various human image generation tasks, including pose re-targeting, virtual try-on, stylization, person re-identification, and textile transfer.Demo and code are available from project page  \url{https://soon-yau.github.io/visconet/}.

\end{abstract}

\begin{figure}
    \centering
    \begin{subfigure}[b]{1\linewidth}
        \includegraphics[width=1.0\linewidth]{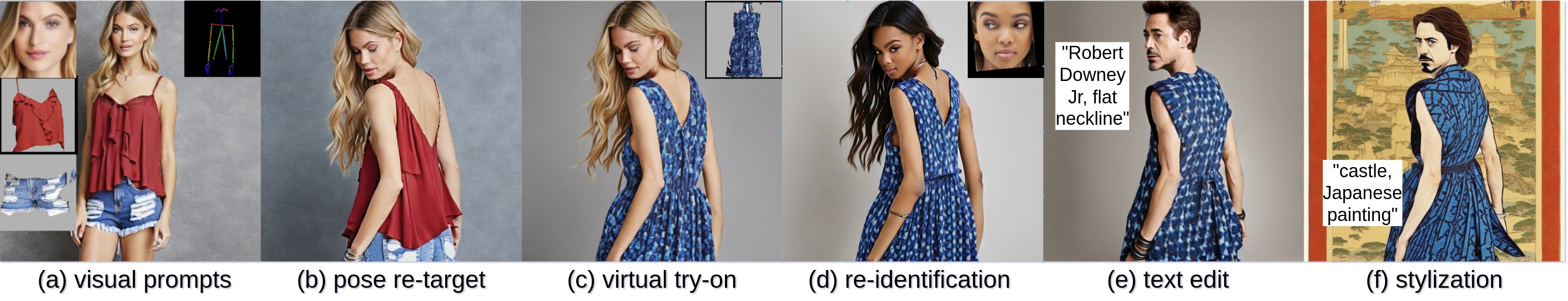}    
    \end{subfigure}
    \begin{subfigure}[b]{1\linewidth}
        \includegraphics[width=1.0\linewidth]{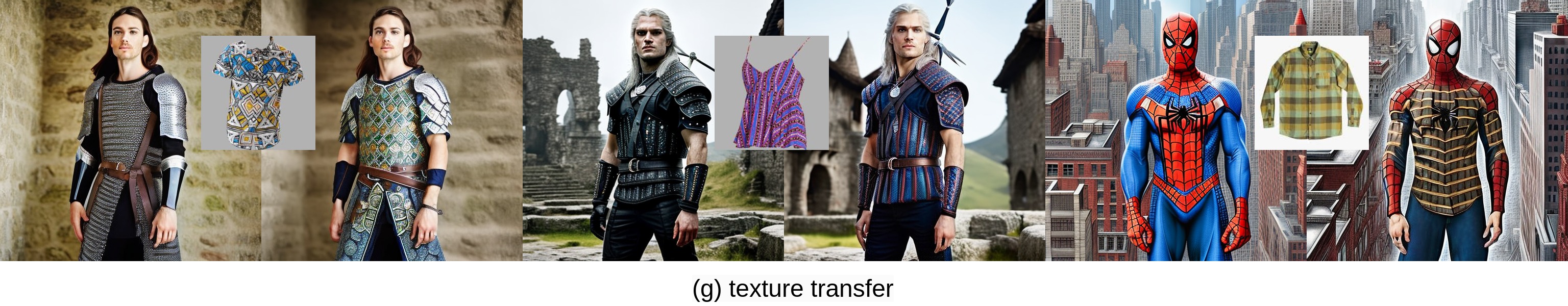}    
    \end{subfigure}  
    \begin{subfigure}[b]{1\linewidth}
        \includegraphics[width=1.0\linewidth]{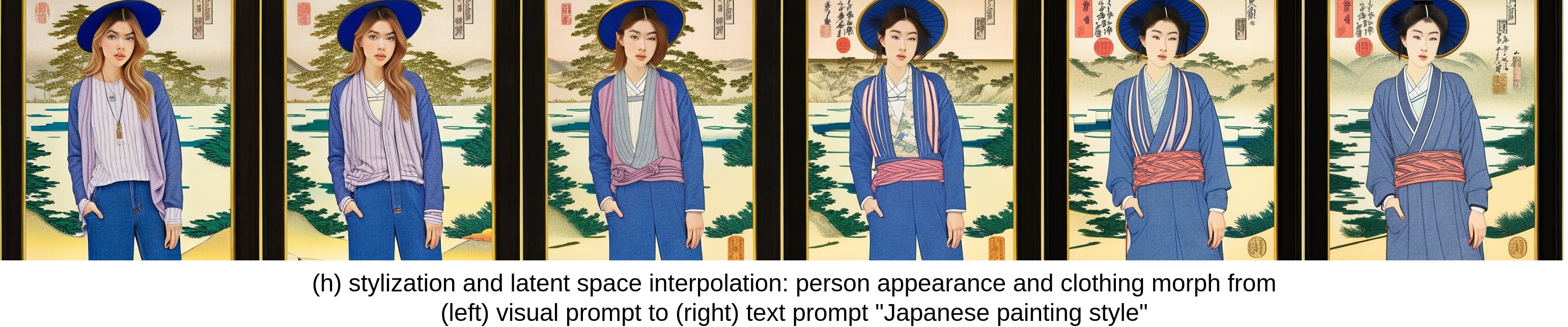}    
    \end{subfigure}            
    \caption{Our proposed \textbf{Visconet} demonstrates broad versatility in multimodal human image tasks including visual prompts, pose re-target, virtual try-on, re-identification using either text or visual prompt, text prompt, texture transfer, stylization and latent space interpolation to perform human morphing.}
    \label{fig:intro:visual}
\end{figure}

\section{Introduction}
\label{sec:intro}

Diffusion models \cite{diffusion_model, dalle2, imagen, glide} are powerful tools for generating realistic and diverse images and videos from various inputs. Among them, latent diffusion models (LDM)\cite{ldm}, more notably Stable Diffusion (SD)~\cite{ldm}, have shown impressive results in text-to-image (T2I) synthesis, thanks to their high quality and open-source availability. However, relying solely on text as the input condition introduces several limitations, notably the challenge of providing a comprehensive description of an image. Furthermore, concept bleeding is a prevalent issue in T2I, as highlighted by works such as \cite{attend-and-excite, divide-and-bind}, where the text becomes erroneously associated with incorrect subjects in the generated images. In human image generation (HIG), this misassociation may manifest in inaccuracies such as assigning the wrong clothing color or experiencing color spillover between clothing and the background, and vice versa.

\begin{figure}[!htb]
\centering  
    \includegraphics[width=0.9\linewidth]{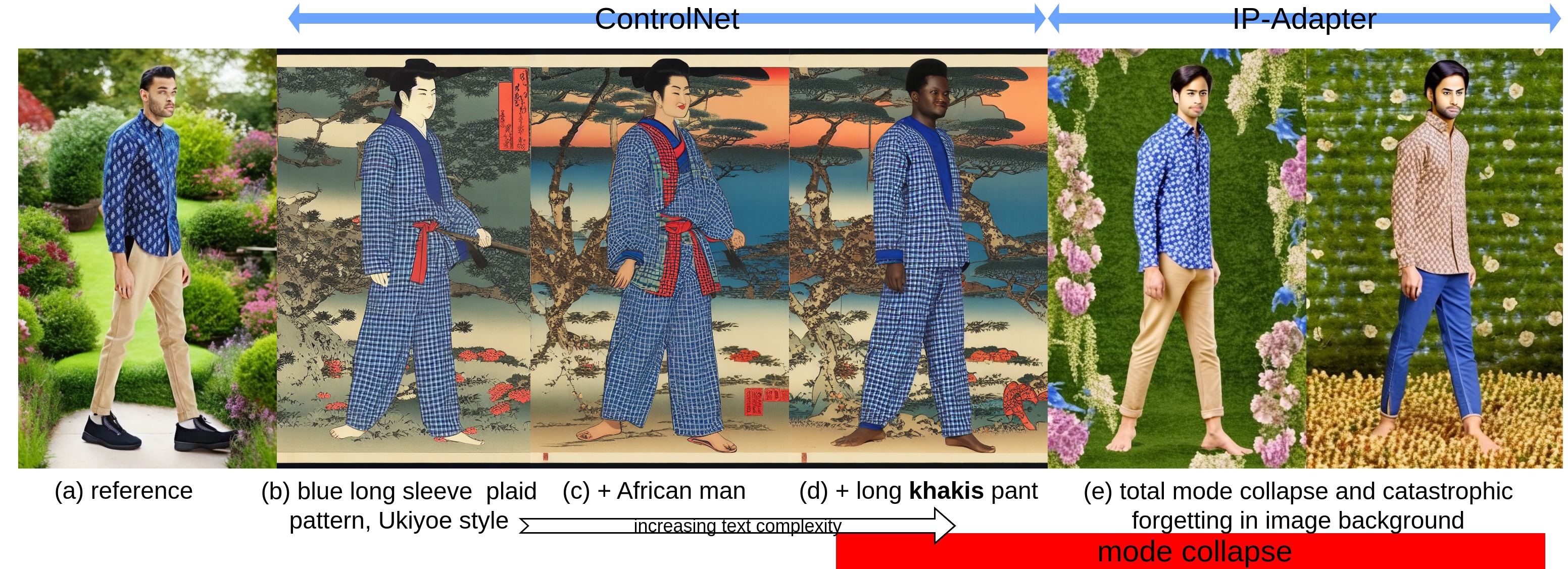}    
    \caption{To motivate our work, this figure illustrates how increasing text complexity in ControlNet~\cite{controlnet} can expose (c) domain gap and eventually lead to mode collapse in (d). IP-Adapter~\cite{ip-adapter} also exhibits (e) catastrophic forgetting, resulting in the inability to generate a rich background. Both show the concept of bleeding by assigning the wrong color to clothing garments.}
\label{fig:intro:collapse}
\end{figure}
\begin{figure}[!htb]
    \centering  
    \begin{subfigure}[b]{0.9\linewidth}
        \includegraphics[width=1.0\linewidth]{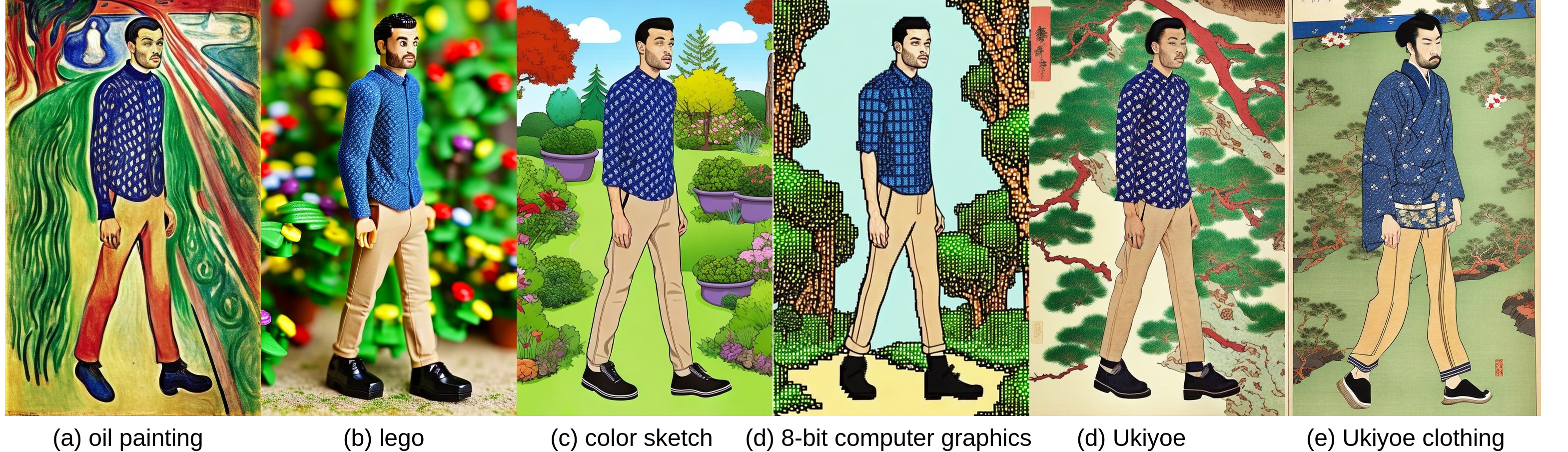}    
    \end{subfigure}        
    \caption{Our method retains generative power of the T2I backbone in (a)-(d) various image styles and rich backgrounds while maintaining the person and clothing appearance, assigning correct clothing colors. In (e), we can control the level of stylization to expand it to the clothing styles.}
    \label{fig:intro:morph}
\end{figure}

While incorporating new conditioning factors, such as pose information, into a model for novel tasks necessitates retraining from scratch ~\cite{humansd, upgpt, human_diffusion, hyperhuman}, demanding substantial datasets and computational resources. Recognizing this challenge, recent methodologies like T2I-Adapter\cite{t2i_adapter} and ControlNet\cite{controlnet} have introduced a pragmatic approach. They integrate a lightweight adapter branch\cite{adapter} to encode spatial conditioning information, such as pose or segmentation maps, onto a frozen pre-trained T2I LDM backbone. In a more recent development, methods such as IP-Adapter\cite{ip-adapter} and MasaCtrl\cite{masactrl} extend this concept by introducing visual conditioning capabilities. However, they cannot control pose independently and require a separate spatial adapter, introducing additional computational complexity to the overall architecture. However, training adapters on smaller and disparate datasets may introduce a domain gap with the frozen LDM model. As noted by \cite{humansd, masactrl}, this conflict between branches can manifest in the model's inability to generate people following specified text and pose conditions. The situation may be exacerbated when multiple adapter branches are employed. Our work addresses this issue by striving to develop a lightweight adapter that accommodates both pose and visual conditioning. This singular adapter aims to excel in a spectrum of human image generation tasks, unifying functionalities currently achievable through utilizing distinct models.

We illustrate the domain gap and conflict of adapters in Figure \ref{fig:intro:collapse} where ControlNet
attempts to reconstruct reference images with increasing text complexity from (b) to (d). Figure \ref{fig:intro:collapse}c shows a sign of domain gap as dark-skinned people were not typical in ancient Japanese drawing (Ukiyoe style). We continue adding ``khaki'', a more modern term, into the text prompt. The complexity eventually exposes the domain gap between ControlNet and T2I. As a result, ControlNet resorts to generating realistic people that it learned from its small training data (Figure \ref{fig:intro:collapse}d). This is a phenomenon we call \textbf{mode collapse (MC)}. Mode collapse has existed since GANs \cite{gan} but has not been discussed recently despite widely affecting recent diffusion model-based adapters. We are the first to study mode collapse in an adapter-based diffusion model systematically. There is currently no effective mechanism to control and manage this conflict; only when one of the conflicting texts, i.e., khaki or Ukiyoe style, is removed will it escape the stuck mode. Unfortunately, this restricts the image content that can be generated. The general solution is to train a more extensive dataset to close the domain gap. HumanSD\cite{humansd} compiled a 1M image dataset, up from ControlNet's 200k, while IP-Adapter uses 10M\cite{ip-adapter} and more recent Hyperhuman \cite{hyperhuman} ballooned to 340M! This is an inefficient use of computing resources, and as we will show, this is insufficient to eradicate mode collapse completely. Conversely, training on a limited dataset may lead to overfitting and, consequently, catastrophic forgetting. This is evident in the model's diminished ability to generate diverse individuals, varied image backgrounds, or encompassing artistic styles as depicted by the input prompts. In contrast, our method trains only on about 50K images, many orders of magnitudes smaller than reference methods.

In this paper, we propose a novel architecture extending ControlNet~\cite{controlnet}, which we call \textbf{ViscoNet} (\textbf{Vis}ual \textbf{Co}ntrol\textbf{N}et), bridging and harmonizing visual and text conditioning. Our method's ability to fuse and control the balance of both text and visual conditioning unlocks unparalleled versatility in HIG, which includes pose re-target (transfer), virtual try-on, person re-identification (face swap) with both text and visual, image stylization, textile transfer, and visual-text latent space interpolation to achieve morphing as shown in Figure \ref{fig:intro:visual}. The summary of our contributions:
\begin{enumerate}
\vspace{-3mm}
    \item A lightweight one-branch adapter architecture for spatial and visual conditioning.
    \item Excellent ability to control and harmonize text and visual prompts, significantly mitigating mode collapse and empowering various HIG capabilities.
    \item Our training with feature masking effectively preserves the backbone model's generative capabilities on a small dataset, mitigating catastrophic forgetting.
\end{enumerate}

\section{Related Works}

\textbf{Human Image Generation} in early days uses GANs \cite{Ma2017, pise, casd, dptn, nted} predominately taking pose and reference image as input conditions to perform pose re-target and virtual try-on. Later, architectures based on transformer \cite{transformer} e.g., \cite{vqgan, dalle} and notably diffusion models \cite{diffusion_model, ldm, glide, imagen, dalle2, persion_dm, humansd} increasingly became the mainstream image generation methods. However, they used only either text or image but not both as input modality, limiting the controllability. Therefore, text prompt is added to specialist HIG models \cite{kpe, upgpt, text2human} to enrich the finer level of control. These models are typically trained from scratch on small datasets resulting in overfitting and an inability to generalize to generate realistic images in diverse, real-world scenarios. 

\textbf{Visual Conditioning.} 
Image personalization methods \cite{ruiz2023dreambooth, textual_inversion} explore finetuning text vocabularies to define specific identities. \cite{chen2023disenbooth,gal2022imageisworth} follow the same idea, while \cite{shi2023instantbooth,jia2023taming,chen2023subjectdriven} leverage large-scale upstream training to eliminate the need for test-time finetuning. These methods use text to control visual aspects rather than images as input conditioning. In HIG, UPGPT \cite{upgpt} pioneered visual conditioning in the T2I diffusion model by concatenating visual tokens alongside text tokens and pose tokens. However, it changes the model architecture and unable to re-use the pre-trained model weights.

\textbf{Adapter.} More recently, adapter modules and lightweight models have been added to pre-trained, frozen diffusion models for faster finetuning requiring less data; among them are ControlNet \cite{controlnet}, T2I-Adapter\cite{t2i_adapter}. However, as they add the learned feature spatially to the UNet's multi-resolution layers in the diffusion model, the control signals are limited to the spatial dimension. Although the T2I-Adapter demonstrates the use of reference images for visual conditioning, it is constrained to the overall artistic style of the image. MasaCtrl\cite{masactrl} is a tuning-free method that injects masked self-attention features from a reference image in the T2I denoising step. IP-Adapter\cite{ip-adapter} uses a separate cross-attention map for image conditioning to be added to the textual attention map. The balance can be adjusted using a weighted average between the two attention maps. Both IP-Adapter and MasaCtrl are conditioned on a single image for a global image, lacking fine-grained visual conditioning. Uni-ControlNet\cite{unicontrolnet} supports both global and local image but still employs dual-branch design.   InstantID\cite{instantid} is based on IP-Adapter's architecture, with the main difference being swapping the CLIP image encoder with a specialized face encoder. While they focus on human face, our method exhibits a broader capacity, generating full human body with higher complexity.

\textbf{Dancing Avatar.} This group of models re-purposes T2I into image-only-conditioning to reconstruct humans for dancing avatar videos faithfully. They sacrifice the T2I's text capability and are not directly comparable to our method. Nevertheless, we scrutinize their pose-and-visual methods. Disco~\cite{disco} uses ControlNet to inject static image background signal. To ensure visual consistency of the moving foreground person, it applies a visual signal to cross-attention of UNet in image-to-image SD variant \cite{sdiv}, which requires re-training. MagicAnimate~\cite{magicanimate} and AnimateAnyone~\cite{animateanyone} use a dedicated adapter branch to encode visual information to be fused with UNet using cross-attention. 

\textbf{Overall}, existing methods \cite{controlnet, t2i_adapter, ip-adapter, unicontrolnet, disco, magicanimate, animateanyone, masactrl} employs multiple adapters for simultaneous pose (e.g. ControlNet) and visual control (e.g. IP-Adapter). Our method introduces improvements over a single ControlNet to offer both pose and visual control, saving computational requirements and potentially mitigating conflicts introduced by multiple branches.



\section{Method}

\subsection{Preliminaries}

\begin{figure}[!htb]
\centering
\includegraphics[width=0.8\linewidth]{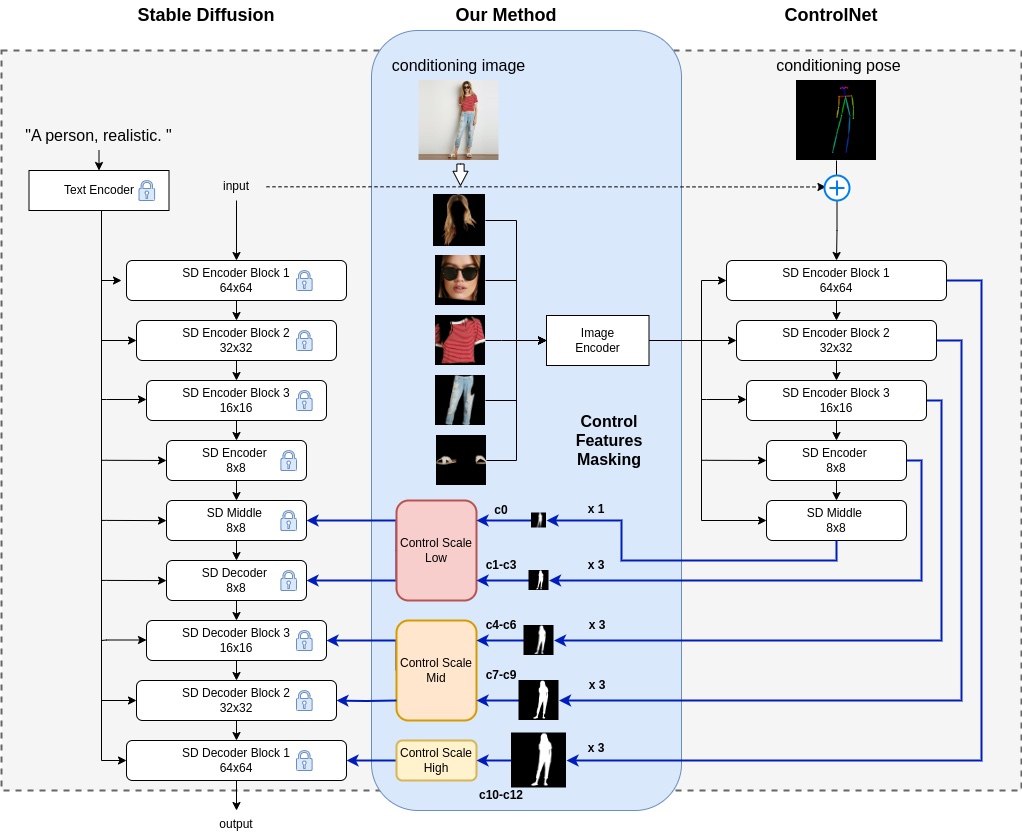}
\caption{Architectural diagram showing our contribution concerning backbone LDM and ControlNet layers. We omit time embedding, zero convolution, and some blocks from the ControlNet diagram \cite{controlnet} for simplicity.}
\label{fig:method}
\end{figure}

Stable Diffusion (SD), a backbone LDM \cite{ldm}, and a ControlNet model~\cite{controlnet} are shown in the left and right block in Figure \ref{fig:method}. SD uses a UNet\cite{unet} as the denoising network and progressively refines the input noise into latent variables that can be reconstructed into realistic synthetic images, relying on understanding intricate image distributions. The words within a text prompt are decomposed into smaller subword units and tokenized and encoded with a CLIP \cite{clip} text transformer \cite{transformer}. The text embedding is injected into the cross-attention layers in UNet, serving as the sole conditioning in image generation. The loss function of the LDM is:
\begin{equation}
    \mathcal{L_{MSE}}  := \mathbb{E}_{z,c,t,\epsilon \sim \mathcal{N}(0,1)} \left[\| \epsilon -\epsilon_{\theta}(z_t, t, c) \|^2_2\right]
    \label{eq:sd}
\end{equation}
where $c$ is the text conditioning token, $t$ is the diffusion time step, and  $z$ is the latent variable (denoted as input in Figure \ref{fig:method}).

Instead of training from scratch, ControlNet~\cite{controlnet} adds a learnable branch parallel with a now frozen pre-trained LDM, as shown on the block on the right in Figure \ref{fig:method}. The branch consists of an identical LDM UNet encoder copy, sharing the same latent noise input and text embedding. It learns to control pose conditions by adding skeleton image features into the latent noise input at the branch input. ControlNet generates spatial control signals and adds them to the SD decoder across multiple spatial resolutions. 

\subsection{Replace Text with Visual Prompt}

That ControlNet's sharing of the exact text embedding with the LDM is unnecessary when learning the spatial condition. Their mandatory use of text-image pairs in training places an excessive burden on data collection and annotation to the specific image and text styles. The text entanglement also increases potential conflict between the branch and LDM \cite{humansd}. Therefore, in our architecture, we remove the text prompt from ControlNet to sever the entanglement and replace it with a visual prompt. Unlike \cite{disco, ip-adapter} that use a single reference image for overall visual conditioning, we use multiple images consisting of segmented body parts (e.g. hair, face, top clothing, bottom clothing) for fine-grained visual control on individual clothing garment pieces (Figure \ref{fig:intro:visual}a-e).

The de-facto image encoding method for the diffusion model uses a CLIP image encoder to extract a global image embedding, but this is insufficient in capturing intricate image details. Therefore,  we utilize the larger dimension local CLIP embedding before the pooling layer. We project the local CLIP embedding of individual images using a linear layer into length \textit{N} and concatenate them like the text tokens they replace. \textit{N} can be adjusted based on the number of conditioning images and the text token length limit of ControlNet, and we use \textit{N}=8 in this work. The linear layer consists of only 2K parameters. It is the only additional trainable parameter introduced by our method, 10,000 times fewer than IP-Adapter's 22M parameters. Controlling all the human information (pose and visual appearance) within the single adapter branch creates effective disentanglement from the LDM to avoid conflict. For example,  ``ripped jeans'' may conflict with ``Picasso painting''; having both in a text prompt could trigger mode collapse in ControlNet. Instead, we can avoid this by removing ``ripped jeans'' from the text prompt and replacing it with an image in the visual prompts. 

\subsection{Control Feature Masking}
DeepFashion\cite{deepfashion} is a popular and de-facto dataset in many HIG tasks in machine vision literature. However, all the images consist of plain studio backgrounds, which will overfit and severely restrain LDM generative capability, resulting in dull and bland image backgrounds. This phenomenon is observed with IP-Adapter~\cite{ip-adapter} in Figure \ref{fig:intro:collapse}e despite it being trained on a large dataset of 10M images. To tackle this issue, we apply a binary human silhouette mask to the control signals originating from our adapter branch before injecting it into the LDM. This eliminates unwanted image backgrounds leaking into and causing overfitting in the LDM. The disentanglement between foreground people and background contributes to reducing the image domain gap with the LDM. This improvement empowers our method to harness LDM text capability to generate vibrant backgrounds in various artistic image styles despite training only on a relatively tiny dataset (only 52K) of images with plain backgrounds. In Section \ref{sec:ablation:mask},  we delve into further analysis, demonstrating that feature masking is essential during training rather than used solely during image sampling.

The feature mask is also applied to the LDM loss function (Equation \ref{eq:sd}) at the \textit{output} in Figure \ref{fig:method}. The training loss backpropagates via the frozen LDM to train the model. This approach is akin to \cite{upgpt,humansd}, although they use it to assign weight loss to different body segmentation parts rather than masking a region entirely. We add masking to the LDM loss function \ref{eq:sd}:  
\begin{equation}
    \mathcal{L_{MSE}}  := \mathbb{E}_{z, c, v, t,\epsilon \sim \mathcal{N}(0,1)} \left[\| \mathcal{M}\odot(\epsilon -\epsilon_{\beta}(z_t, t, c, v) \|^2_2)\right]
    \label{eq:mse}
\end{equation}
where $\epsilon_{\beta}$ is the model, $v$ is the image embedding, $\odot$ is the element-wise multiplication, and $\mathcal{M} \in \mathbb{R}^{H, W} $ is the binary mask resized to resolution (H, W) of the LDM output. Although text conditions are not used in the model, they are used by LDM in training and, thus, are included in the equation. 

\subsection{Harmonizing Text and Visual Influence}
\label{sec:control_scale}
The versatility of our approach in performing diverse human image generation tasks arises from its ability to seamlessly integrate and regulate the balance between visual and textual conditioning. We achieve this by multiplying scalar values $[0.0, 1.0]$ with the control features. Scaling control signals is commonplace in adapter-based approaches, but our novel model architecture unlocks unprecedented effects not observed in existing methods. Despite innovations adopted to reduce data conflict between the adapter and the LDM, mode collapse can still happen in challenging image styles. In this scenario, we can decrease the control signal strength to weaken the visual prompt strength to escape the mode collapse. As we will show, the application of this approach has no discernible impact on mitigating mode collapse in ControlNet~\cite{controlnet} and other spatial conditioned models~\cite{t2i_adapter, humansd}. This is attributed to the fact that its control signals exclusively influence pose conditioning, whereas the root causes of the conflict lie in the image domain gap and text entanglement. In contrast, our innovative architecture, which involves the separation and subsequent bridging of text and visual conditioning, empowers us to dynamically adjust their balance, thereby enabling latent space interpolation (Figure \ref{fig:intro:visual}h) and eliminating mode collapse.

On the other hand, IP-Adapter~\cite{ip-adapter} supports visual prompts, and it can adjust the text-visual balance by changing the scales of the respective cross-attention map, but the effect is global to the image. For example, tipping the balance away from the visual prompt of a realistic photo of a man towards the text prompt ``a girl, Chinese ink painting'' would result in the global transformation of a modern man towards a Chinese girl wearing period Chinese clothing in Chinese ink painting style. Our method can apply different scaling at each multi-spatial resolution to customize at different image levels. This is demonstrated in Figure \ref{fig:intro:visual}f, which depicts only the image's artistic style while retaining the person's identity and appearance, and Figure \ref{fig:intro:visual}h, which shows the morphing only of the person, leaving the background essentially unchanged. 

For the sake of discussion, the 13 individual control strength scales (c0-c12) can be roughly grouped into three blocks - Low Blocks (LB), Mid Blocks (MB), and High Blocks (HB) arranged hierarchically from low to high spatial resolution. We can adjust their values separately to create different effects. Through experimentation, we observed that LB exerts negligible influence and can effectively be set to 0. The MB is the most influential in overall visual appearance styles among them. HB regulates fine image texture, aligning with our expectations for image hierarchy control. Setting HB alone yields the notable outcome of transferring only the texture of the visual prompt (Figure \ref{fig:intro:visual}g). We can also constrain our control to pose only by setting c4 to 0.5 while leaving others 0.0, allowing using text prompts to control the whole person's appearance.

\subsection{Training Setup}
To train the model, we employ 52K-images  DeepFashion In-shop Clothes Retrieval dataset \cite{deepfashion} and adopt the train-test split proposed by \cite{patn} for the pose transfer task, padding the images to the size of $512\times512$. Pose information is extracted using OpenPose \cite{openpose} to create body-and-hand skeleton images, and we use pre-segmented fashion images from \cite{upgpt}. We employ a simple text prompt of ``a person'' for all the images. This serves two purposes: first, the neutral description avoids potential conflicts with the LDM, proving our method does not need to annotate text to match the style of the LDM carefully. Secondly, it acts as an unconditional text embedding, enabling users to amplify the desired visual effect using positive prompts, negative prompts, and guidance scales\cite{Dhariwal2021}. 

Many adapter models are based on pre-trained SD or similar-sized models. Thus, we also performed our experiments using SD2.1\cite{sd2} for a fair comparison. We initialize our adapter branch by copying frozen weights from the SD. However, since the cross-attention input has shifted from global CLIP text embedding to local CLIP image embedding, we re-initialize the weights in the cross-attention layer at the start of training. All weights in the SD, CLIP text, and image encoders are frozen. We use CLIP image encoder \textit{clip-vit-large-patch14}\cite{huggingface_llm}. We trained the model on a single desktop GPU GTX 3090 for 2 epochs, using a batch size of 4 with four gradient accumulations per batch, resulting in an effective batch size of 16. We retained the remaining configurations from~\cite{controlnet}. 

\subsection{Image Resolution}
We use an image resolution of $512\times512$ throughout the paper. In our experiment, we utilized $\nicefrac{3}{4}$ length to full-body images, resulting in smaller human faces within the images. The stringent demand for high pixel density per latent variable can lead to suboptimal face construction\cite{upgpt} compared to high resolution face images generated by \cite{ip-adapter, instantid}. This inherent limitation is a characteristic drawback of the LDM rather than a weakness of our method.

\section{Experiments}
In Section \ref{sec:exp:1}, we perform an in-depth study of the effect of control strength on mode collapse as observed in image artistic styles. We show that visual prompt methods (IP-Adapter and ours) effectively reduce mode collapse compared to ControlNet. Then, in Section \ref{sec:exp:2}, we perform further, more challenging experiments in person re-identification to show our method has superior text-visual harmonization capability compared to IP-Adapter. Lastly,  we performed large-scale human evaluation in Section \ref{sec:exp:3} to prove our image quality over SOTA HIG models. 

\subsection{Mode Collapse and Control Strength}
\label{sec:exp:1}

\begin{figure}[!ht]
\centering
\begin{subfigure}[b]{1\linewidth}
    \begin{turn}{90}
        \begin{minipage}{0.15\textwidth}
            \centering
            ControlNet
        \end{minipage}
    \end{turn}
    \begin{subfigure}[b]{0.14\columnwidth}
        \includegraphics[width=1.0\linewidth]{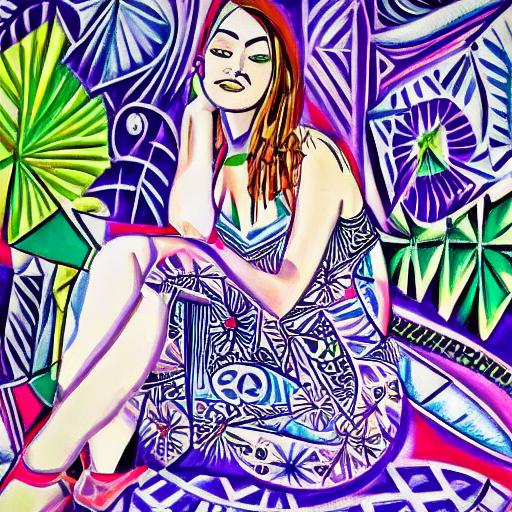}
    \end{subfigure}
    \begin{subfigure}[b]{0.14\columnwidth}
        \includegraphics[width=1.0\linewidth]{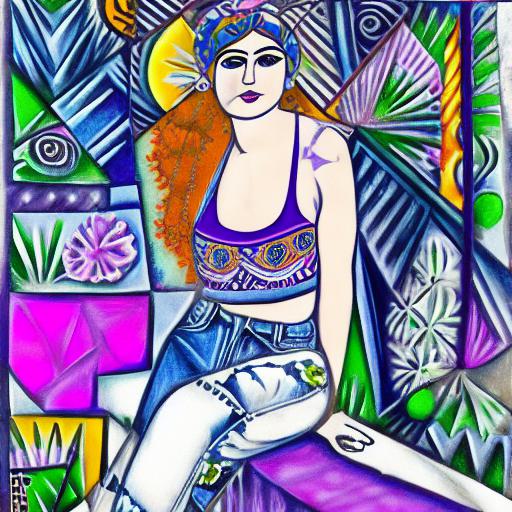}
    \end{subfigure}
    \begin{subfigure}[b]{0.14\columnwidth}
        \includegraphics[width=1.0\linewidth]{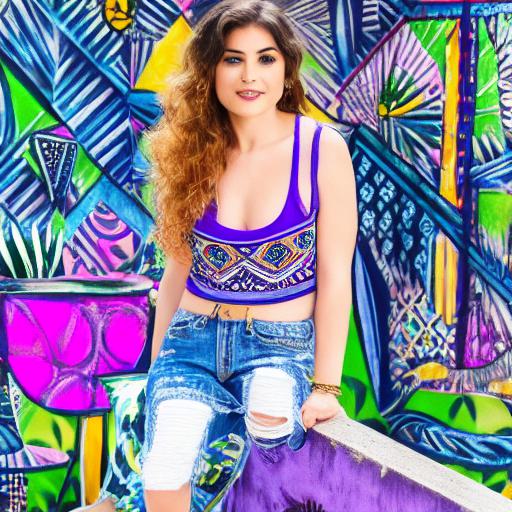}
    \end{subfigure}
    \begin{subfigure}[b]{0.14\columnwidth}
        \includegraphics[width=1.0\linewidth]{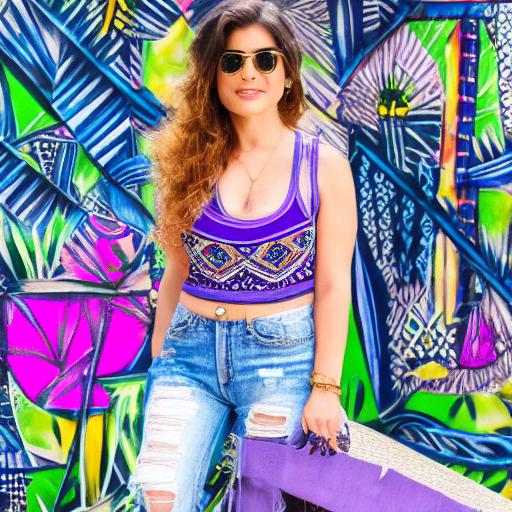}
    \end{subfigure}
    \begin{subfigure}[b]{0.14\columnwidth}
        \includegraphics[width=1.0\linewidth]{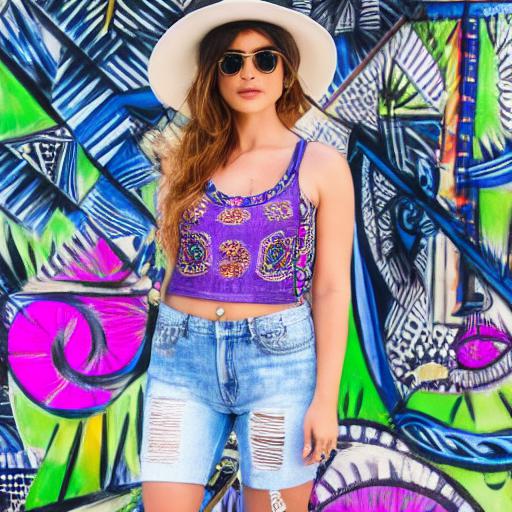}
    \end{subfigure}
    \begin{subfigure}[b]{0.14\columnwidth}
        \centering
        \fbox{\includegraphics[width=1\textwidth]{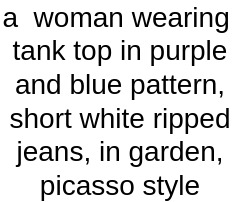}}
    \end{subfigure}        
\end{subfigure}
\begin{subfigure}[b]{1\linewidth}
    \begin{turn}{90}
        \begin{minipage}{0.15\textwidth}
            \centering
            {\fontsize{8}{0}\selectfont IP-Adapter}
            
        \end{minipage}
    \end{turn}    
    \begin{subfigure}[b]{0.14\columnwidth}
        \includegraphics[width=1.0\linewidth]{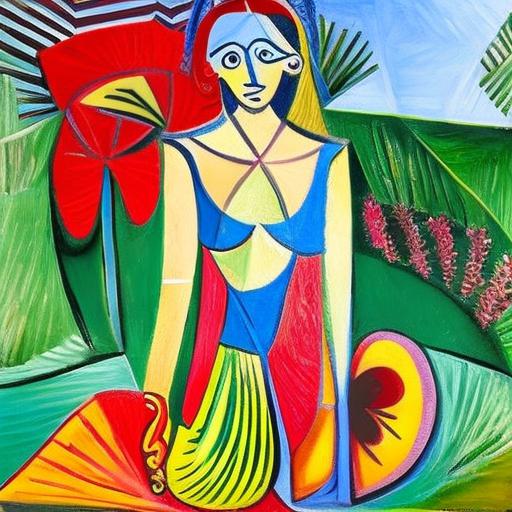}
    \end{subfigure}
    \begin{subfigure}[b]{0.14\columnwidth}
        \includegraphics[width=1.0\linewidth]{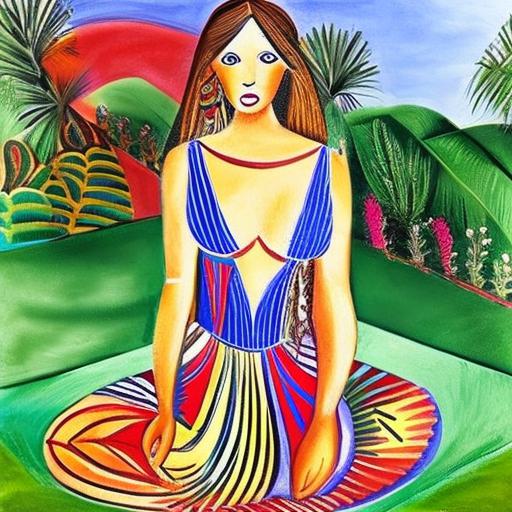}
    \end{subfigure}
    \begin{subfigure}[b]{0.14\columnwidth}
        \includegraphics[width=1.0\linewidth]{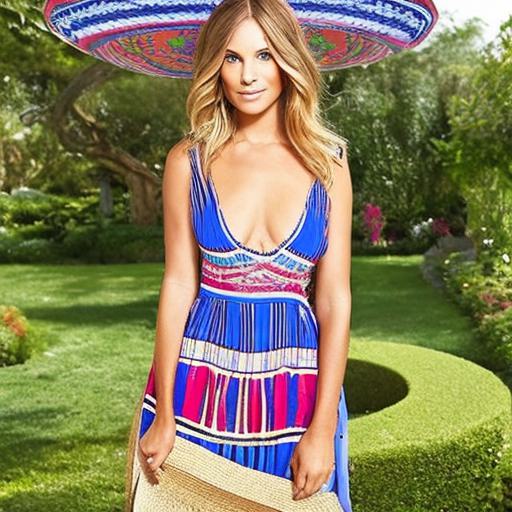}
    \end{subfigure}
    \begin{subfigure}[b]{0.14\columnwidth}
        \includegraphics[width=1.0\linewidth]{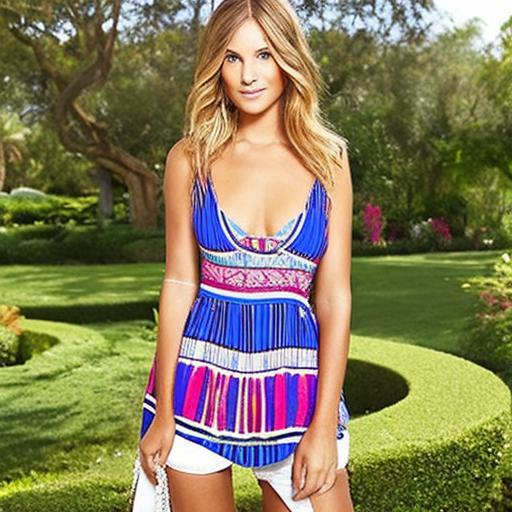}
    \end{subfigure}
    \begin{subfigure}[b]{0.14\columnwidth}
        \includegraphics[width=1.0\linewidth]{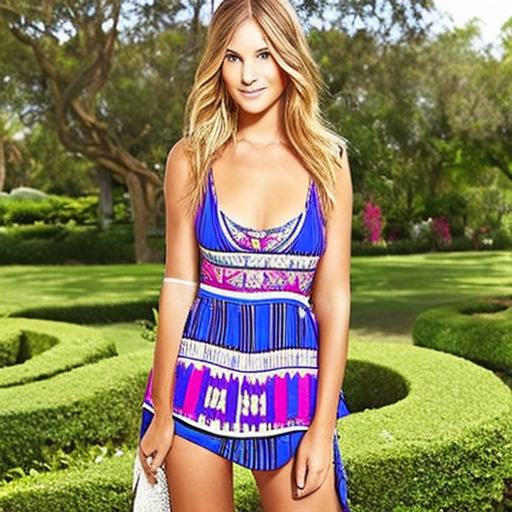}
    \end{subfigure}
    \begin{subfigure}[b]{0.14\columnwidth}
        \includegraphics[width=1.0\linewidth]{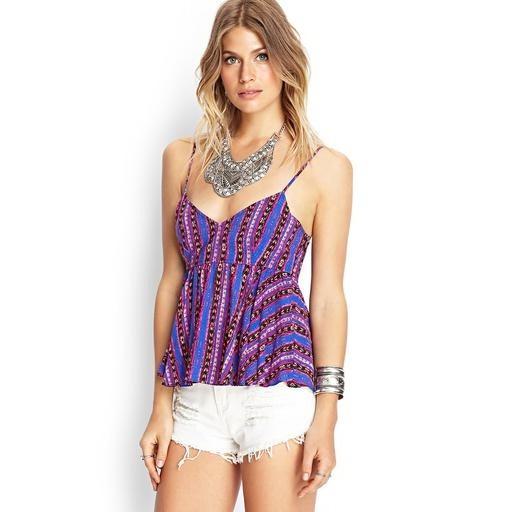}
    \end{subfigure}    
\end{subfigure}
\begin{subfigure}[b]{1\linewidth}
    \begin{turn}{90}
        \begin{minipage}{0.15\textwidth}
            \centering
            Ours
        \end{minipage}
    \end{turn}    
    \begin{subfigure}[b]{0.14\columnwidth}
        \includegraphics[width=1.0\linewidth]{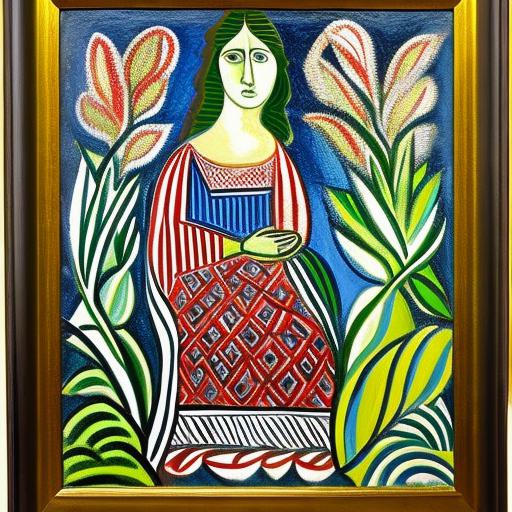}
        \caption{20\%}
        \label{fig:mc:20}
    \end{subfigure}
    \begin{subfigure}[b]{0.14\columnwidth}
        \includegraphics[width=1.0\linewidth]{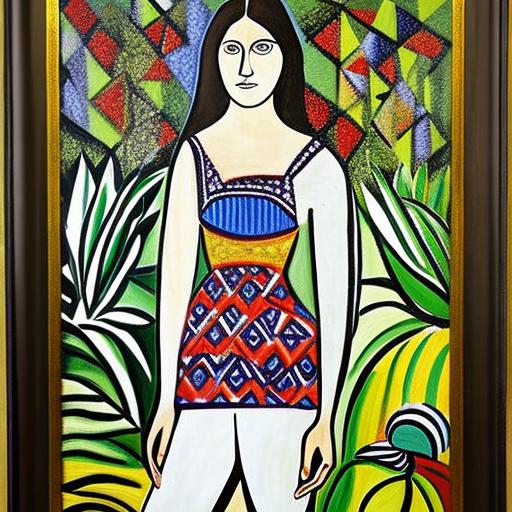}
        \caption{40\%}
        \label{fig:mc:40}
    \end{subfigure}
    \begin{subfigure}[b]{0.14\columnwidth}
        \includegraphics[width=1.0\linewidth]{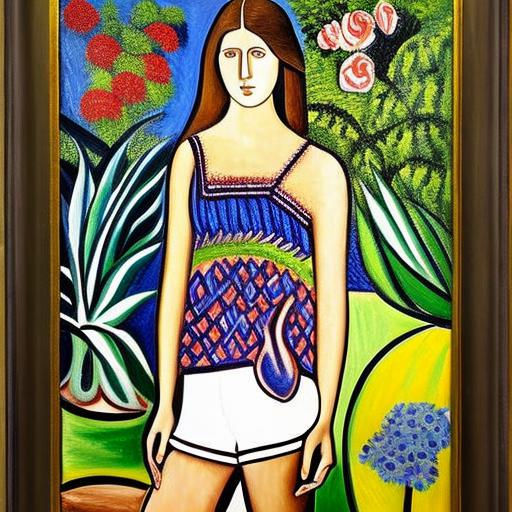}
        \caption{50\%}
        \label{fig:mc:50}
    \end{subfigure}
    \begin{subfigure}[b]{0.14\columnwidth}
        \includegraphics[width=1.0\linewidth]{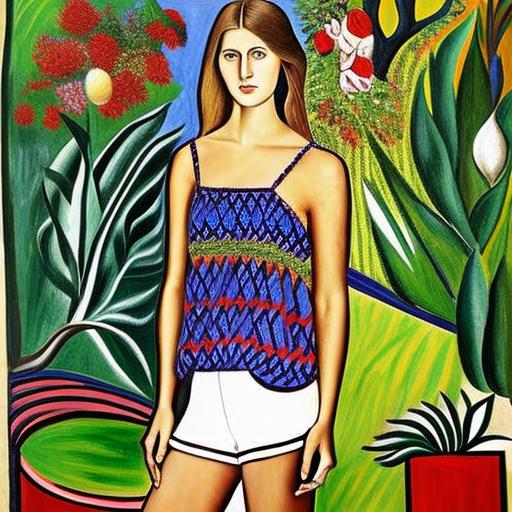}
        \caption{60\%}
        \label{fig:mc:60}
    \end{subfigure}
    \begin{subfigure}[b]{0.14\columnwidth}
        \includegraphics[width=1.0\linewidth]{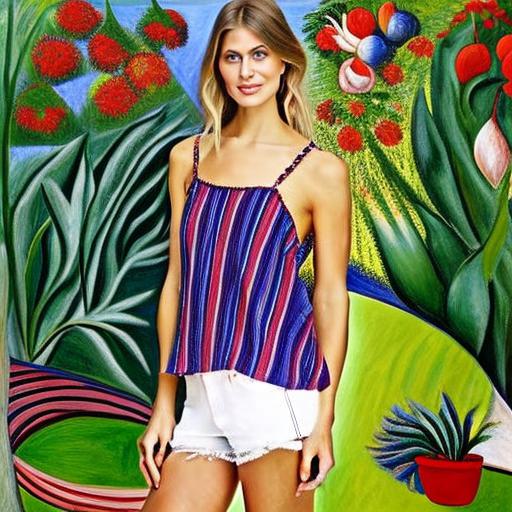}
        \caption{80\%}
        \label{fig:mc:80}
    \end{subfigure}
    \begin{subfigure}[b]{0.14\columnwidth}
        \centering
        \includegraphics[width=1.0\linewidth]{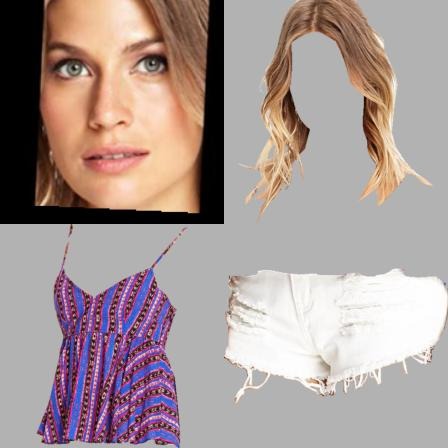}
        \caption{Reference}
        \label{fig:mc:ref}
    \end{subfigure}    
\end{subfigure}

\caption{Effect of control strength (\%). Compared to ControlNet and IP Adapter, our method can escape mode collapse faster, generating a harmonious image style while maintaining good visual control.}
\label{fig:mc_strength}
\end{figure}

In this section, we examine the prevalence of mode collapse and its impact on existing spatial and visual adapter models compared to our proposed model. In Figure \ref{fig:mc_strength}, conditioned on the same human pose, we generate images of \textit{Picasso style} at different control strengths. ControlNet does not have visual input. Therefore, we use text to describe the person's appearance and background, which include conflicting word \textit{``ripped jeans''} to invoke mode collapse. In this example, mode collapse happens to both ControlNet~\cite{controlnet}, IP-Adapter~\cite{ip-adapter}, and our proposed ViscoNet at a control strength of 80\%, as observed with the realistic person and background in Figure \ref{fig:mc:80}. However, our method has quickly escaped mode collapse at a control strength of 60\% (Figure \ref{fig:mc:60}), as at this point, the visual conditioning is still effective, maintaining the overall clothing styles and colors. We can also observe the harmonized transition towards the desired image style as reduced control strength tips the balance towards text prompt depicting \textit{``Picasso''}(Figure \ref{fig:mc:20}). Both reference methods only managed to escape mode collapse at around 40\% (Figure \ref{fig:mc:40}), which has considerably weakened pose or visual control for ControlNet and IP-Adapter, respectively.

We confirm our qualitative observation with quantitative results. Like \cite{ip-adapter}, we measure the effectiveness in generating the correct image styles by employing the CLIP similarity score between the text prompt and the generated image. A high CLIP score indicates a low or absence of mode collapse. We measure control effectiveness by measuring pose accuracy using the Object Keypoint Similarity (OKS) standard in MSCOCO challenge\cite{mscoco}. We will also introduce new metrics to measure and interpret mode collapse better. In this experiment, we selected 5 image styles - Picasso, Van Gogh, oil painting, Ukiyoe, and stained glass that are more likely to conflict with modern clothing. We generated 20 samples at each control strength (over 5000 images). The results are plotted in Figure \ref{fig:graph_strength}, and we include the entire table in the appendix. 

At 100\% control strength, IP-Adapter lost most of its text capability, including changing image style (Figure \ref{fig:intro:collapse}e), resulting in the lowest CLIP scores (Figure \ref{fig:graph_strength}a), indicating substantial mode collapse. The CLIP score for ControlNet remains constant in regions above 40\%, whereas our method exhibits linear improvement in the same range. Both visual adapters effectively reduce mode collapse by using weaker control strength as indicated by weaker pose accuracy (Figure \ref{fig:graph_strength}c).  Our method consistently outperforms IP-Adapter in CLIP score at every control strength. It is worth noting that the IP-Adapter maintains its pose control for control strength <40\% as they use separate adapters for pose control. Their drop of <40\% is attributed to inaccuracy in pose detectors' recognition of humans in artistic painting. Our quantitative results align with qualitative observations, establishing our method's superior interpolation capability and ability to minimize mode collapse. 

\begin{figure}
    \centering
    \includegraphics[width=1.0\linewidth]{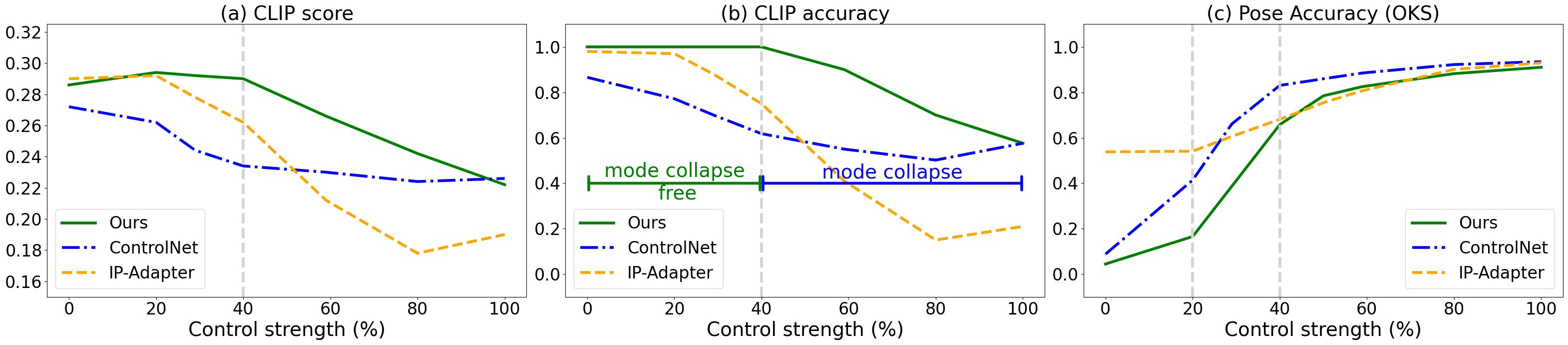}
    \caption{(a) Reducing control strength alleviates mode collapse, our method can escape mode collapse faster, retaining better pose and visual control (b) CLIP accuracy provides better interpretability of mode collapse (c) Level of control as measured by pose accuracy. Visual prompting methods have slightly weaker pose control. }
    \label{fig:graph_strength}    

\end{figure}

While CLIP scores are effective, their limitation lies in the lack of interpretability regarding the degree of mode collapse. Additionally, the absence of a standardized CLIP model within the machine learning community introduces variability, making cross-model comparisons challenging. Given these challenges, we explore alternative metrics for a more comprehensive evaluation. As mode collapse is an inherent discrete state, we employ CLIP binary classification ($CLIP_{acc}$) by comparing CLIP image embedding to two CLIP text classes - [image style],``\textit{real photo}''. More generally, two modes are compared - target mode and stuck mode.  In other words, we detect mode collapse if the image is classified as a real photo when it was supposed to be in the target image style. As shown in Figure \ref{fig:graph_strength}b, $CLIP_{acc}$ correlates well to the CLIP similarity score but provides a normalized score easier for interpretation and enhanced robustness against CLIP model variation. We define \textbf{mode collapse rate (MCR)} as : 
\begin{equation}
    \mathcal{MCR}  := 1 - CLIP_{acc}
    \label{eq:mcr}
\end{equation}

Mode collapse is a phenomenon that occurs randomly, depending on the prompts and random seeds applied. Consequently, the MCR is a batch statistic that reflects the overall method performance. In Figure \ref{fig:graph_strength}b, we achieve mode collapse free at a control strength of 40\% (MCR=0\% or $CLIP_{acc}=100\%$) while IP-Adapter reaches that state later at much-weakened control strength.

\subsection{Re-identification}
\label{sec:exp:2}

\begin{figure}[ht]
    \centering
    \begin{subfigure}[b]{1\linewidth}
        \centering
        \begin{subfigure}[b]{0.16\columnwidth}
            \includegraphics[width=1.0\linewidth]{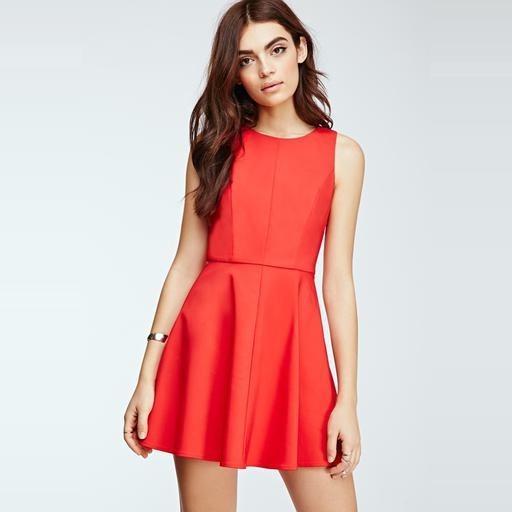}
            \caption{Reference}
        \end{subfigure}        
        \begin{subfigure}[b]{0.16\columnwidth}
            \includegraphics[width=1.0\linewidth]{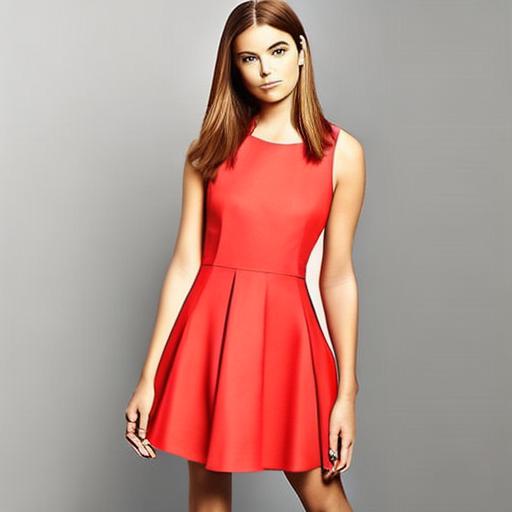}
            \caption{100\%}
        \end{subfigure}        
        \begin{subfigure}[b]{0.16\columnwidth}
            \includegraphics[width=1.0\linewidth]{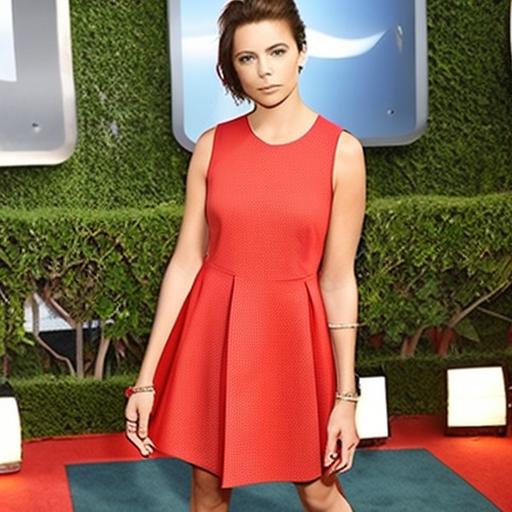}
            \caption{63\%}
        \end{subfigure}        
        \begin{subfigure}[b]{0.16\columnwidth}
            \includegraphics[width=1.0\linewidth]{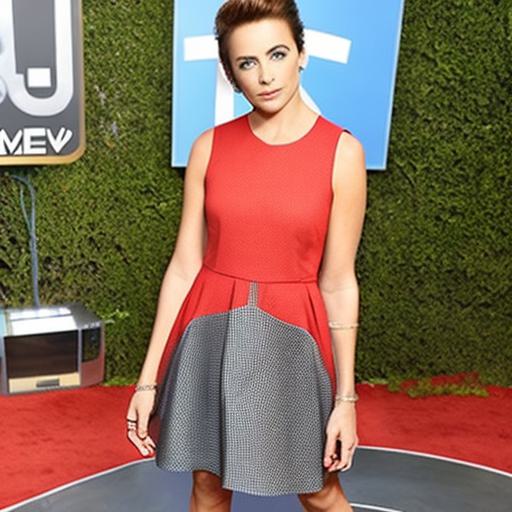}
            \caption{\textcolor{red}{\textbf{62\%}}}
        \end{subfigure}
        \begin{subfigure}[b]{0.16\columnwidth}
            \includegraphics[width=1.0\linewidth]{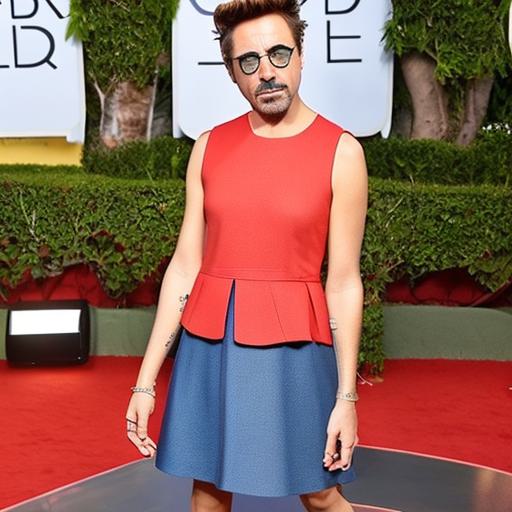}
            \caption{\textcolor{red}{\textbf{61\%}}}
        \end{subfigure}
        \begin{subfigure}[b]{0.16\columnwidth}
            \includegraphics[width=1.0\linewidth]{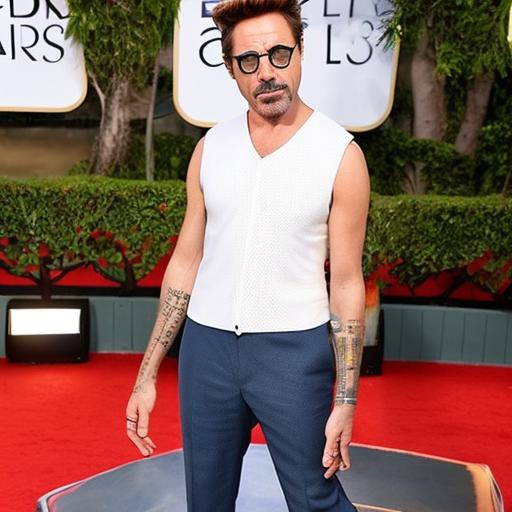}
            \caption{40\%}
        \end{subfigure}       
    \end{subfigure}
    \caption{IP-Adapter showing the transition from the reference image to text prompt ``Robert Downey Jr.'' by reducing control strength. There exists a big domain gap between (d) and (e).}
    \label{fig:bridge1:ip}

    \begin{subfigure}[b]{1\linewidth}
        \centering
        \begin{subfigure}[b]{0.16\columnwidth}
            \includegraphics[width=1.0\linewidth]{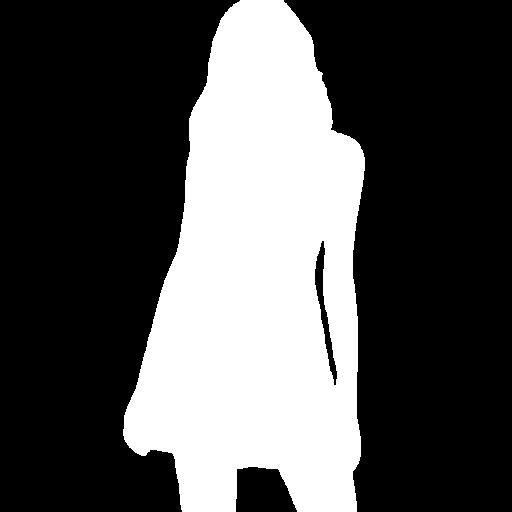}
            \caption{Mask}
        \end{subfigure}        
        \begin{subfigure}[b]{0.16\columnwidth}
            \includegraphics[width=1.0\linewidth]{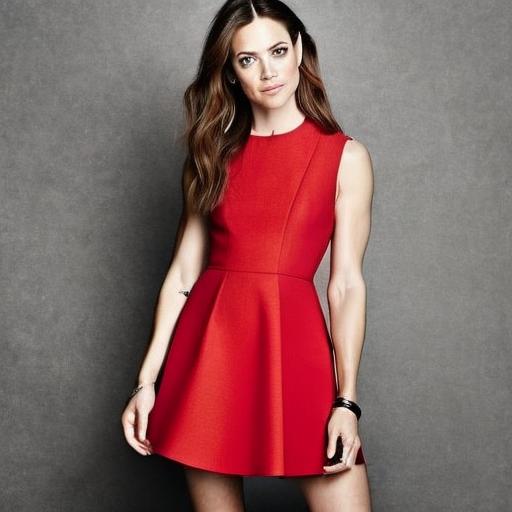}
            \caption{50\%}
        \end{subfigure}        
        \begin{subfigure}[b]{0.16\columnwidth}
            \includegraphics[width=1.0\linewidth]{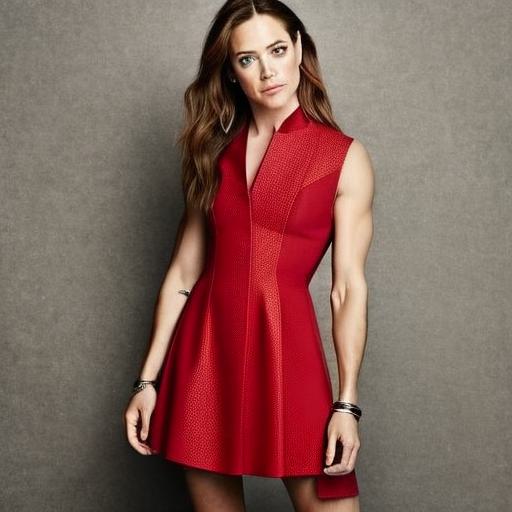}
            \caption{40\%}
        \end{subfigure}        
        \begin{subfigure}[b]{0.16\columnwidth}
            \includegraphics[width=1.0\linewidth]{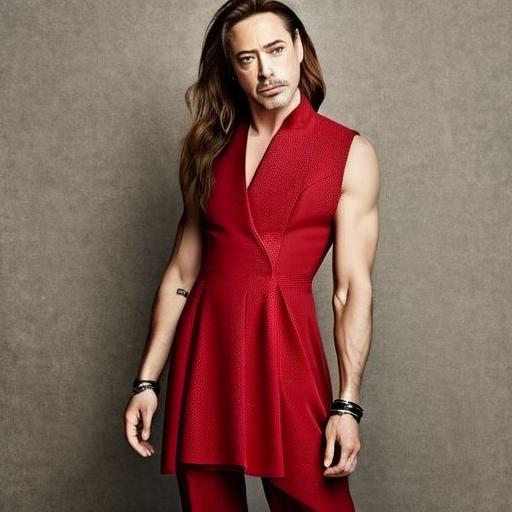}
            \caption{39\%}
        \end{subfigure}
        \begin{subfigure}[b]{0.16\columnwidth}
            \includegraphics[width=1.0\linewidth]{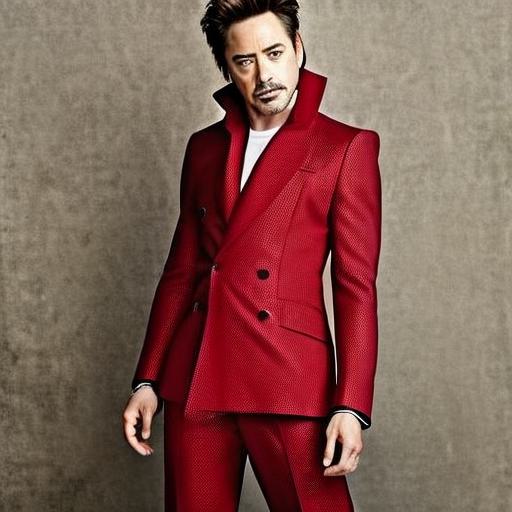}
            \caption{38\%}
        \end{subfigure}
        \begin{subfigure}[b]{0.16\columnwidth}
            \includegraphics[width=1.0\linewidth]{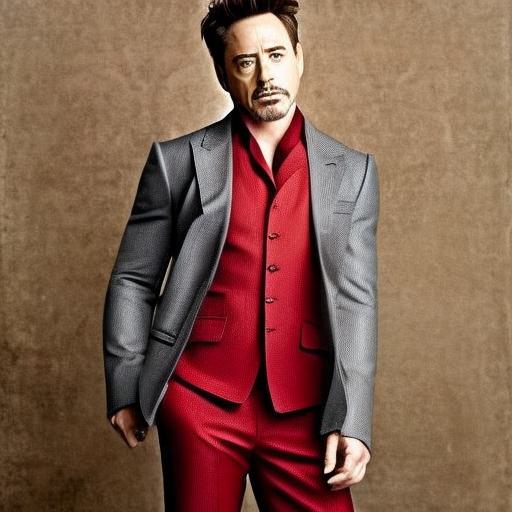}
            \caption{30\%}
        \end{subfigure}
    \end{subfigure}
    \caption{Method 1 - with head mask, smoother transition with smaller mode gap between (c) and (d).}
    \label{fig:bridge1:ours:whead}

    \begin{subfigure}[b]{1\linewidth}
        \centering
        \begin{subfigure}[b]{0.16\columnwidth}
            \includegraphics[width=1.0\linewidth]{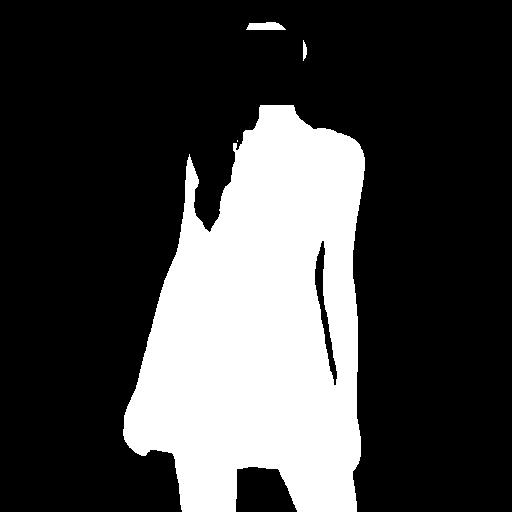}
            \caption{Mask}
        \end{subfigure}        
        \begin{subfigure}[b]{0.16\columnwidth}
            \includegraphics[width=1.0\linewidth]{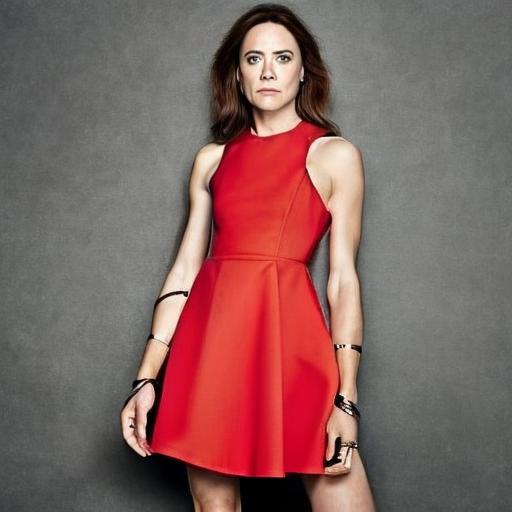}
            \caption{100\%}
        \end{subfigure}        
        \begin{subfigure}[b]{0.16\columnwidth}
            \includegraphics[width=1.0\linewidth]{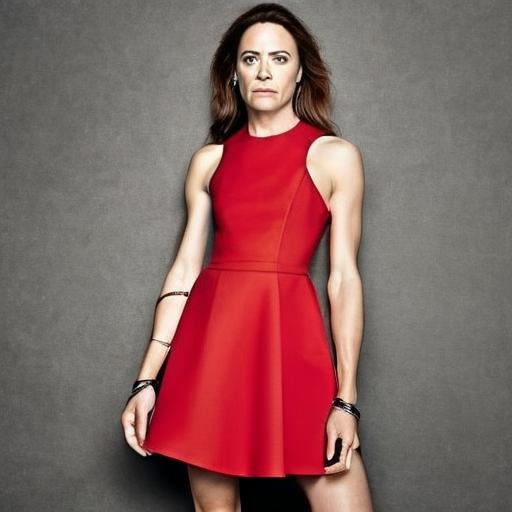}
            \caption{70\%}
        \end{subfigure}        
        \begin{subfigure}[b]{0.16\columnwidth}
            \includegraphics[width=1.0\linewidth]{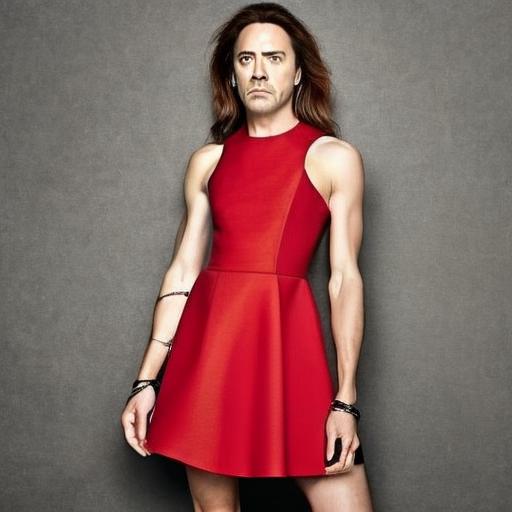}
            \caption{60\%}
        \end{subfigure}
        \begin{subfigure}[b]{0.16\columnwidth}
            \includegraphics[width=1.0\linewidth]{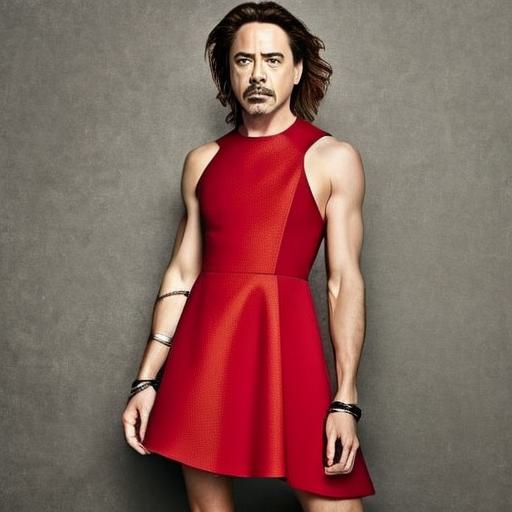}
            \caption{50\% \textbf{(best)}}
        \end{subfigure}
        \begin{subfigure}[b]{0.16\columnwidth}
            \includegraphics[width=1.0\linewidth]{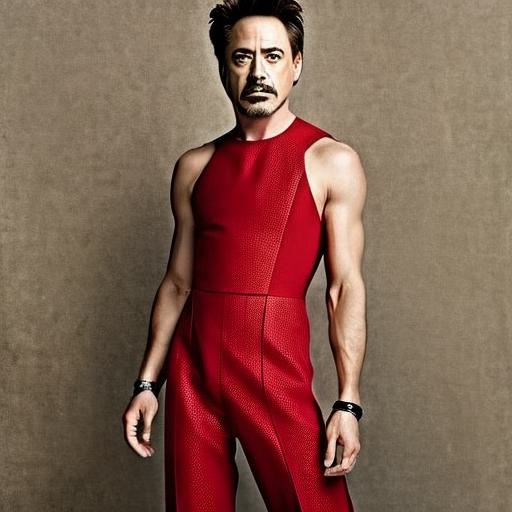}
            \caption{40\%}
        \end{subfigure}
    \end{subfigure}
    \caption{Method 2 - without head mask. Smooth transition with (e) achieving good balance to deliver the desired result. The face and hair mask are detected and removed by segmentation tool. Our method has good tolerance over the mask region and does not require it to be pixel-accurate.}
    \label{fig:bridge1:ours:wohead}
    
\end{figure}

We formulated a demanding task to scrutinize an extreme instance of domain gap and assess the qualitative efficacy of visual prompting methods in addressing such challenges. In this task, the goal is to transform the person's identity in the reference image into the person depicted in the text prompt, all while preserving the original clothing depicted in the reference image. In Figure \ref{fig:bridge1:ip}, we show that decreasing control strength in IP-Adapter morphs the face towards the target (Robert Downey Jr.) at the expense of clothing faithfulness (red dress). A small control strength change between Figure \ref{fig:bridge1:ip}d and \ref{fig:bridge1:ip}e causes a significant shift in the image, indicating a big domain gap it fails to bridge harmoniously. 

This common problem also affects our default configuration Method 1, which uses full human tasks.  It achieves good results close to the target as shown in Figure \ref{fig:bridge1:ours:whead}d. Through extensive experimentation, we discovered that the face has disproportionately influenced the entire image generation process. Consequently, it becomes imperative to substantially reduce the control strength (to around 40\% in this example) to mitigate the impact of the face, albeit at the expense of visual control. Leveraging our novel architecture, we can effectively bridge this gap by selectively excluding the face from the feature mask, as shown in Figure \ref{fig:bridge1:ours:wohead} (Method 2). In essence, this action prevents the control signal from reaching the face region of the LDM. We tried applying a similar approach to the IP-Adapter by masking off the head from the reference image in pixel space, but this proved ineffective. This underscores the efficacy of our novel architecture in harmonizing text-visual controls. This has also proved useful in escaping mode collapse in certain challenging styles in stylization tasks (see appendix).

\begin{figure}[!ht]
    \centering
    \includegraphics[width=1.0\linewidth]{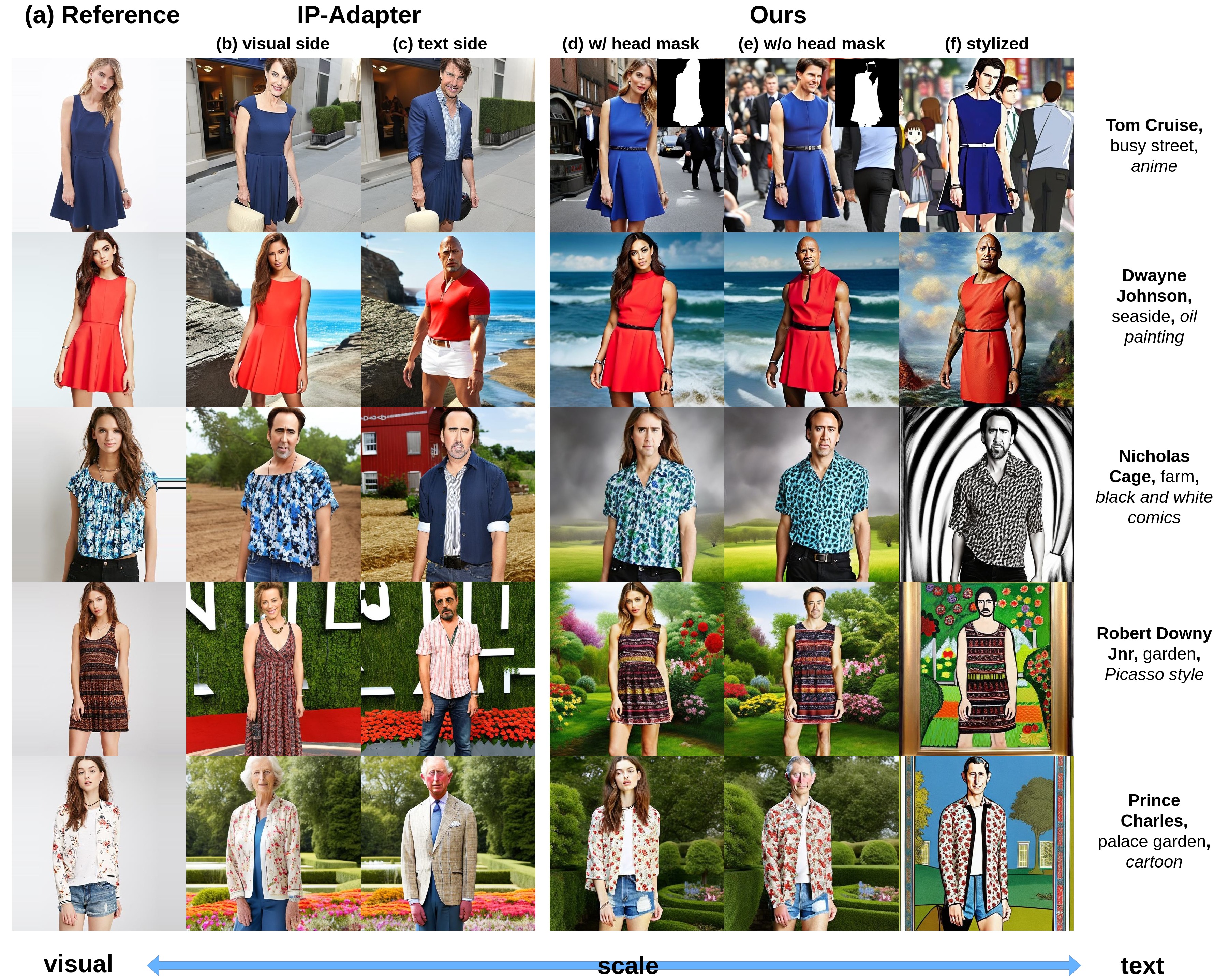}
    \caption{Challenging re-identification task to transform female in (a) reference image to male celebrities depicted in text prompt. We included an additional stylization step (not included in the result) to demonstrate our ability to bridge the domain gap.}

    \label{fig:bridge2}
\end{figure}

We generated over 5000 images from each method to perform the quantitative study; some test samples (input image and text) are shown in Figure \ref{fig:bridge2}. We include the image background and style to demonstrate our capability to maintain a constant background and bridging domain gaps across multiple dimensions to achieve stylization. We do not include them in our experiments as the objective is the foreground person identity and clothing. The experiment results are summarized in Figure \ref{fig:celeb_chart} (full table in appendix). The presence of steep change in CLIP score (Figure \ref{fig:celeb_chart}a) and MCR (Figure \ref{fig:celeb_chart}b) with our Method 1 proves the evident domain gap within the 30\%-50\% control strength range. However, removing the face from the mask in Method 2 drastically improves performance, outperforming IP-Adapter considerably. On the other hand, we measure effectiveness of visual control with image similarity score MS-SSIM \cite{mssim} (Figure \ref{fig:celeb_chart}c). Method 1 (and 2) is consistently higher than IP-Adapter in MS-SSIM, suggesting more faithful visual appearance once escaping mode collapse (Figure \ref{fig:bridge1:ip} and \ref{fig:bridge1:ours:whead}) .

\begin{figure}[!ht]
    \centering
    \includegraphics[width=1.0\linewidth]{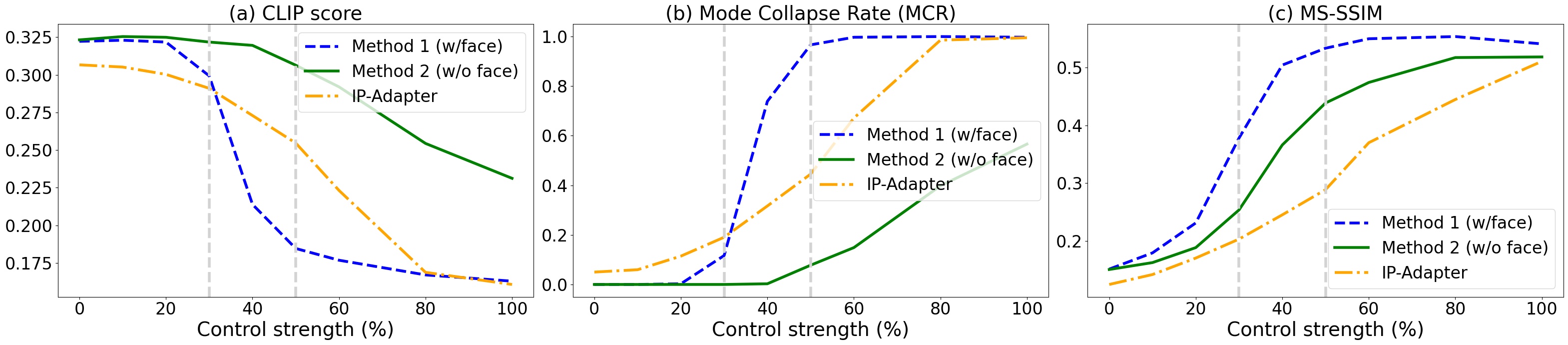}
    \caption{Quantitative result showing the effectiveness of our method to escape mode collapse in the challenging re-identification task.}
    \label{fig:celeb_chart}
\end{figure}

\subsection{Generating Diverse Human Image Styles}
\label{sec:exp:3}
\begin{figure*}[!ht]
\centering
\begin{subfigure}[b]{1\linewidth}
\centering
\begin{turn}{90}
    \begin{minipage}{0.1\textwidth}
        \centering
        ControlNet
    \end{minipage}
\end{turn}
\begin{subfigure}[b]{0.14\columnwidth}
    \includegraphics[width=1.0\linewidth]{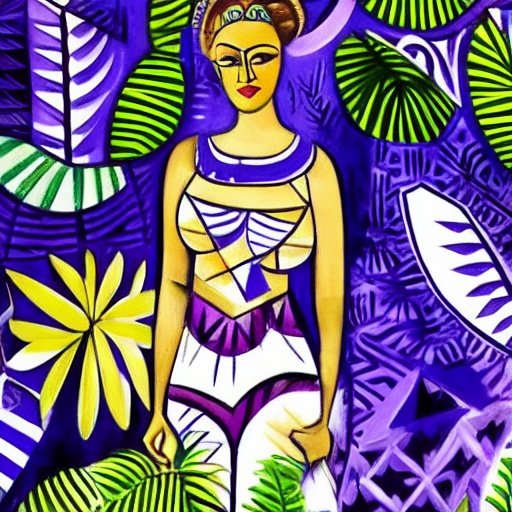}
\end{subfigure}
\begin{subfigure}[b]{0.14\columnwidth}
    \includegraphics[width=1.0\linewidth]{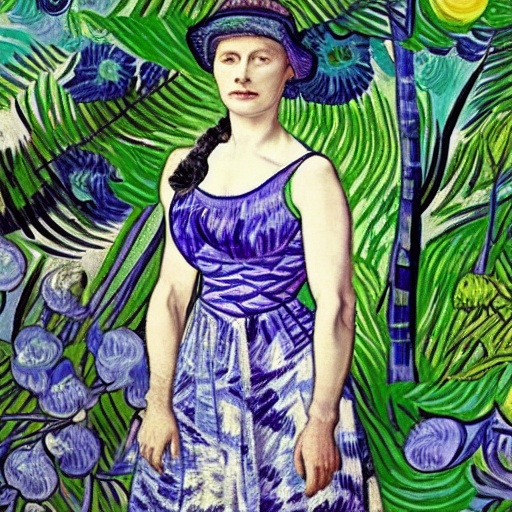}
\end{subfigure}
\begin{subfigure}[b]{0.14\columnwidth}
    \includegraphics[width=1.0\linewidth]{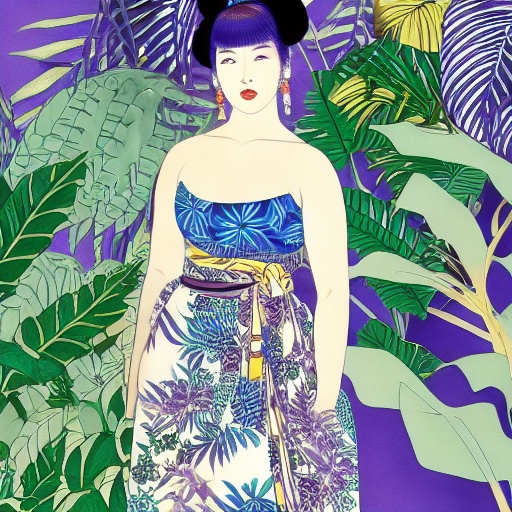}
\end{subfigure}
\begin{subfigure}[b]{0.14\columnwidth}
    \includegraphics[width=1.0\linewidth]{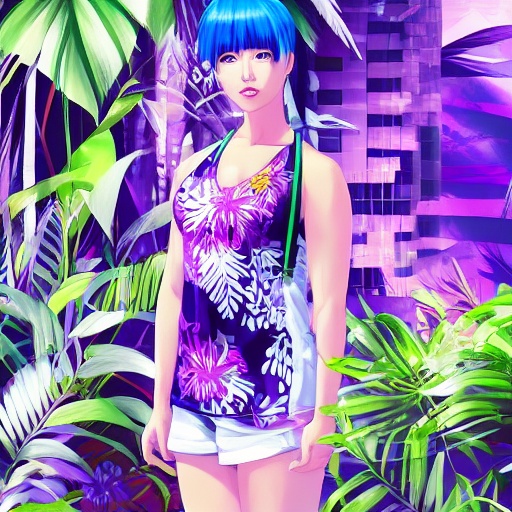}
\end{subfigure}
\begin{subfigure}[b]{0.14\columnwidth}
    \includegraphics[width=1.0\linewidth]{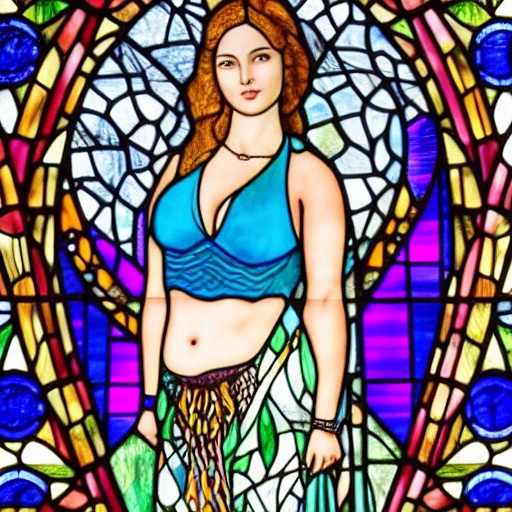}
\end{subfigure}
\begin{subfigure}[b]{0.14\columnwidth}
    \includegraphics[width=1.0\linewidth]{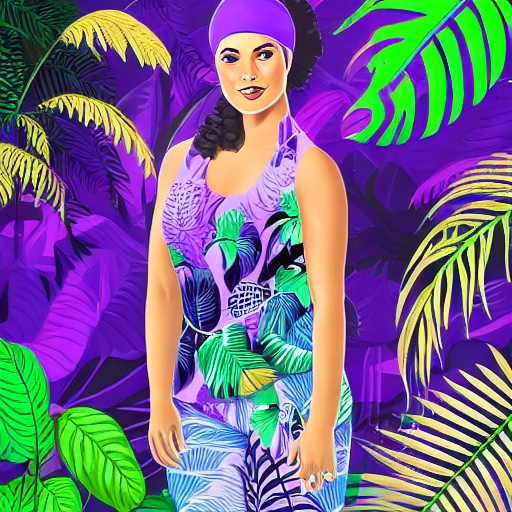}
\end{subfigure}
\end{subfigure}
\begin{subfigure}[b]{1\linewidth}
\centering
\begin{turn}{90}
    \begin{minipage}{0.1\textwidth}
        \centering
        T2I
    \end{minipage}
\end{turn}
\begin{subfigure}[b]{0.14\columnwidth}
    \includegraphics[width=1.0\linewidth]{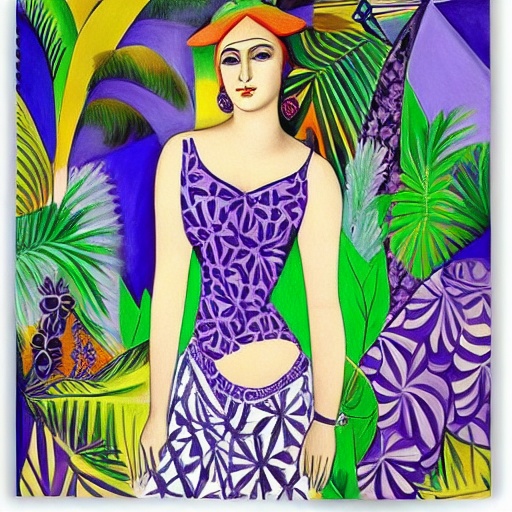}
\end{subfigure}
\begin{subfigure}[b]{0.14\columnwidth}
    \includegraphics[width=1.0\linewidth]{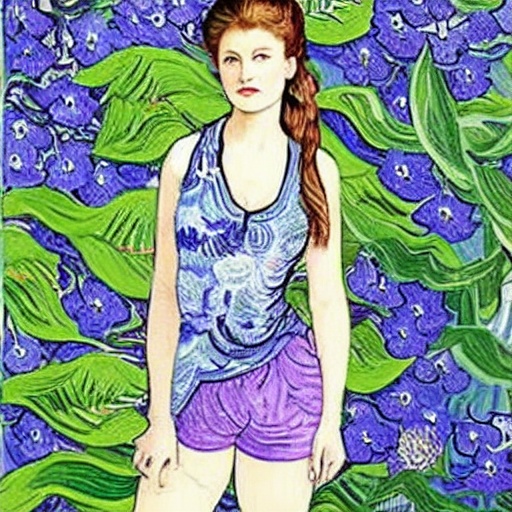}
\end{subfigure}
\begin{subfigure}[b]{0.14\columnwidth}
    \includegraphics[width=1.0\linewidth]{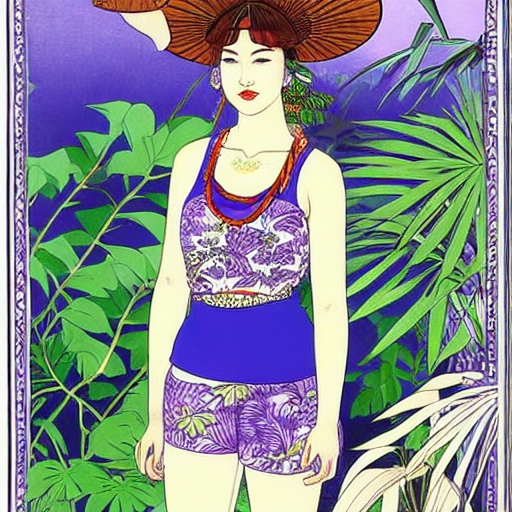}
\end{subfigure}
\begin{subfigure}[b]{0.14\columnwidth}
    \includegraphics[width=1.0\linewidth]{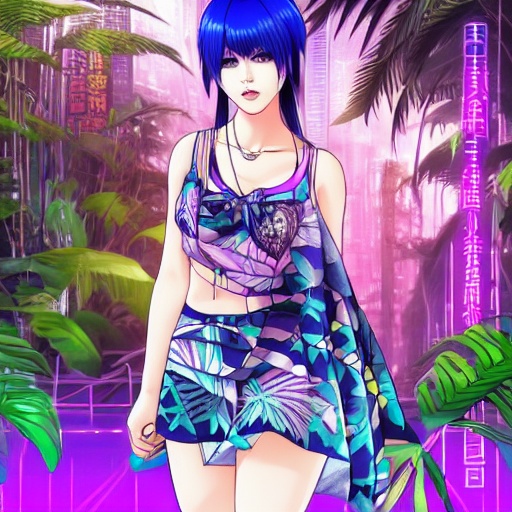}
\end{subfigure}
\begin{subfigure}[b]{0.14\columnwidth}
    \includegraphics[width=1.0\linewidth]{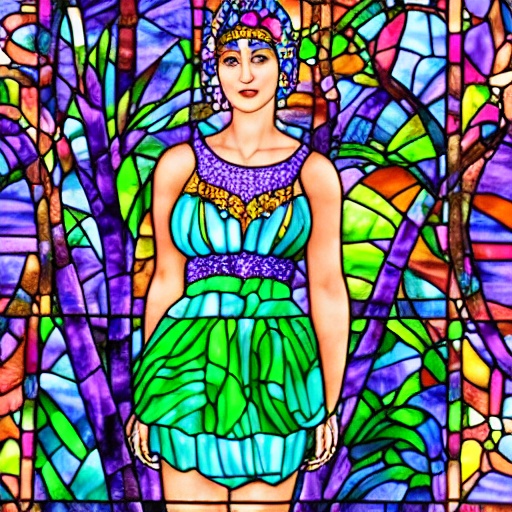}
\end{subfigure}
\begin{subfigure}[b]{0.14\columnwidth}
    \includegraphics[width=1.0\linewidth]{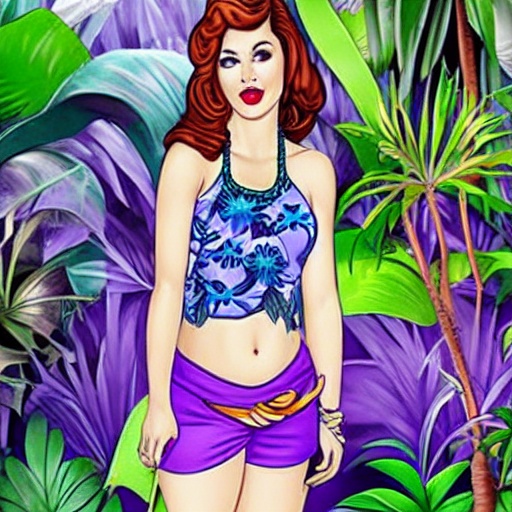}
\end{subfigure}
\end{subfigure}
\begin{subfigure}[b]{1\linewidth}
\centering
\begin{turn}{90}
    \begin{minipage}{0.1\textwidth}
        \centering
        HumanSD
    \end{minipage}
\end{turn}
\begin{subfigure}[b]{0.14\columnwidth}
    \includegraphics[width=1.0\linewidth]{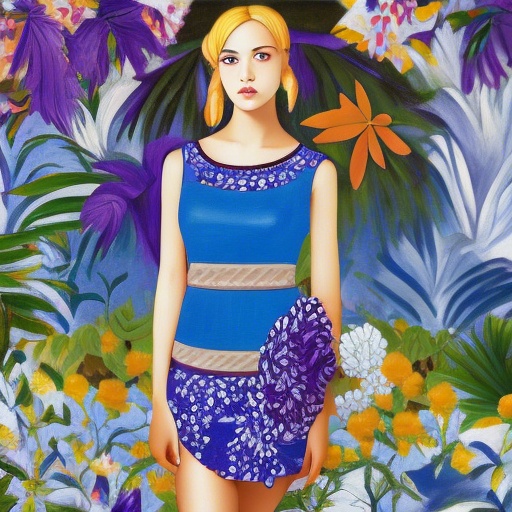}
\end{subfigure}
\begin{subfigure}[b]{0.14\columnwidth}
    \includegraphics[width=1.0\linewidth]{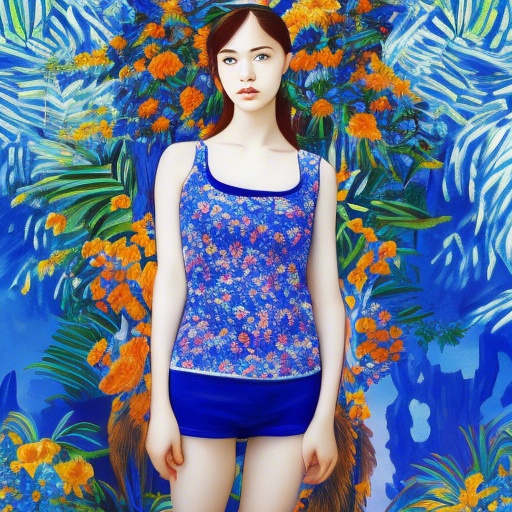}
\end{subfigure}
\begin{subfigure}[b]{0.14\columnwidth}
    \includegraphics[width=1.0\linewidth]{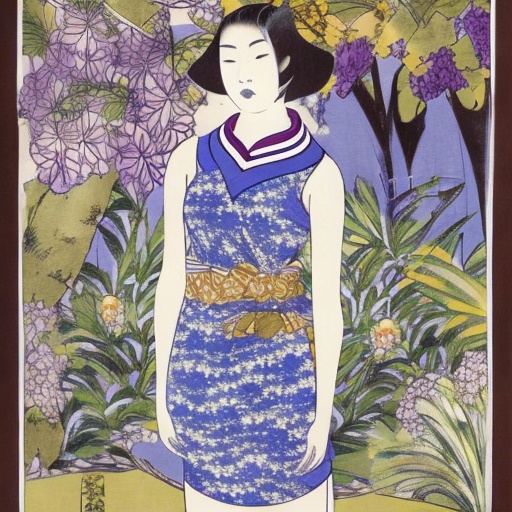}
\end{subfigure}
\begin{subfigure}[b]{0.14\columnwidth}
    \includegraphics[width=1.0\linewidth]{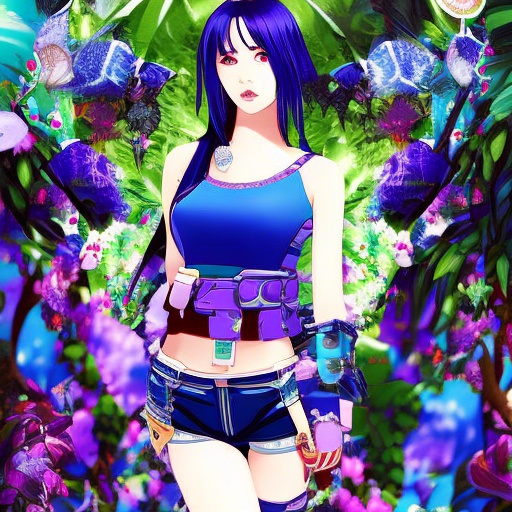}
\end{subfigure}
\begin{subfigure}[b]{0.14\columnwidth}
    \includegraphics[width=1.0\linewidth]{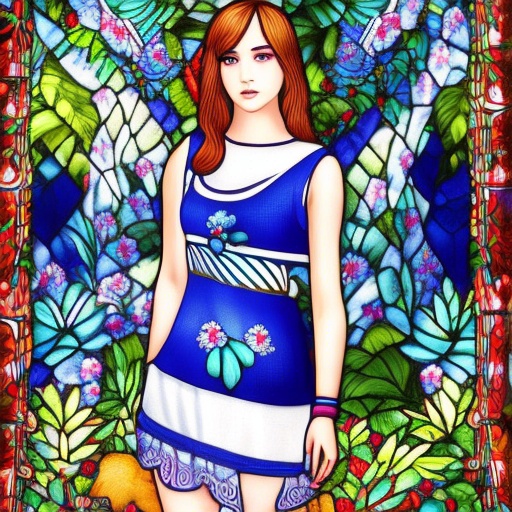}
\end{subfigure}
\begin{subfigure}[b]{0.14\columnwidth}
    \includegraphics[width=1.0\linewidth]{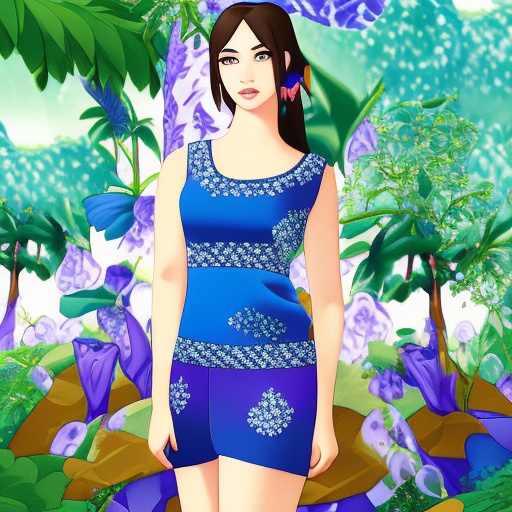}
\end{subfigure}
\end{subfigure}
\begin{subfigure}[b]{1.0\linewidth}
\centering
\begin{turn}{90}
    \begin{minipage}{0.1\textwidth}
        \centering
        Ours
    \end{minipage}
\end{turn}
\begin{subfigure}[b]{0.14\columnwidth}
    \includegraphics[width=1.0\linewidth]{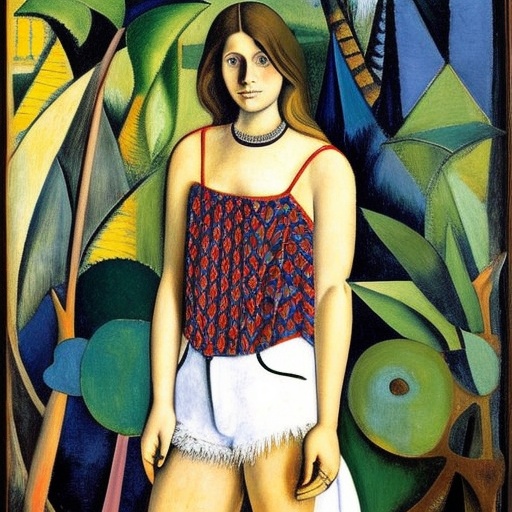}
    \caption{Picasso}
\end{subfigure}
\begin{subfigure}[b]{0.14\columnwidth}
    \includegraphics[width=1.0\linewidth]{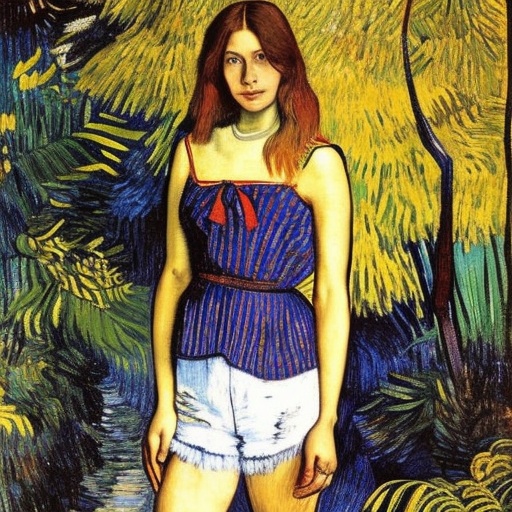}
    \caption{Van Gogh}
\end{subfigure}
\begin{subfigure}[b]{0.14\columnwidth}
    \includegraphics[width=1.0\linewidth]{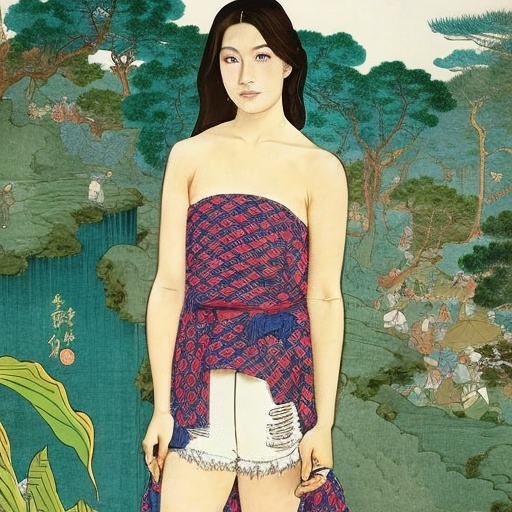}
    \caption{Ukiyoe}
\end{subfigure}
\begin{subfigure}[b]{0.14\columnwidth}
    \includegraphics[width=1.0\linewidth]{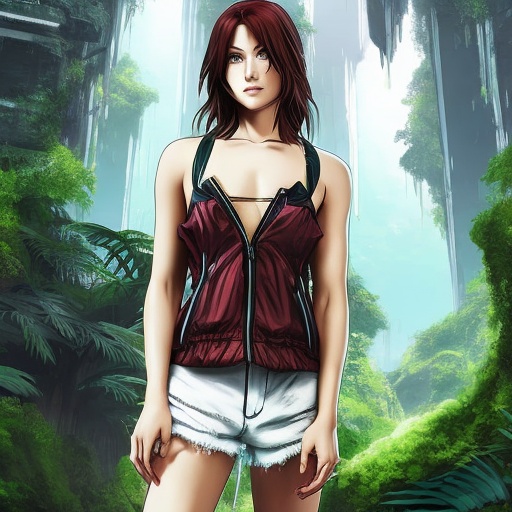}
    \caption{cyberpunk}
\end{subfigure}
\begin{subfigure}[b]{0.14\columnwidth}
    \includegraphics[width=1.0\linewidth]{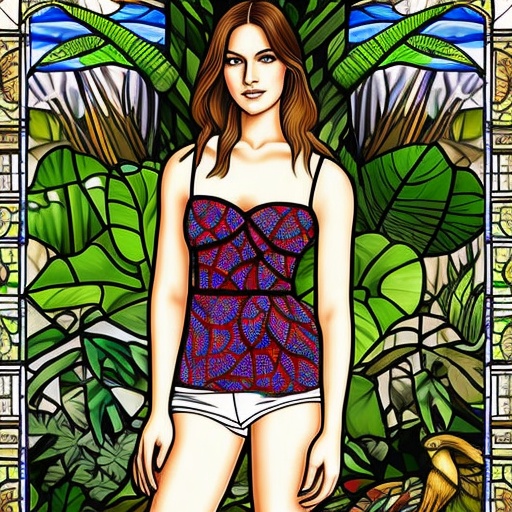}
    \caption{stained glass}
\end{subfigure}
\begin{subfigure}[b]{0.14\columnwidth}
    \includegraphics[width=1.0\linewidth]{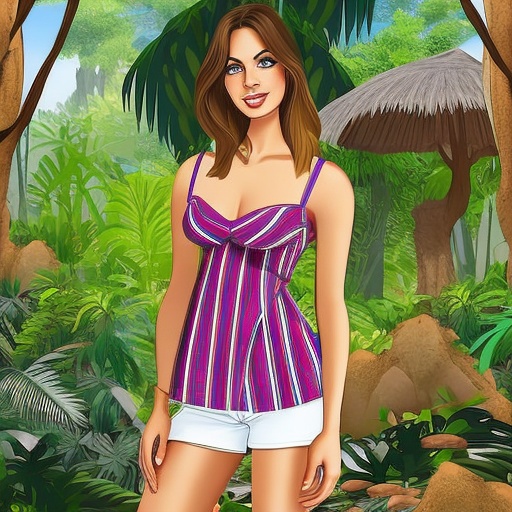}
    \caption{Disney}
    \label{fig:disney}    
\end{subfigure}
\end{subfigure}
\caption{All reference methods have the purple clothing color spread into the forest background, while our method avoids this problem and can generate a vibrant and diverse background.}
\label{fig:color_spread}
\end{figure*}

We performed large-scale human evaluation comparing specialist SOTA pose-guided HIG models HumandSD \cite{huggingface_llm}, ControlNet \cite{controlnet}, and T2I-Adapter \cite{t2i_adapter}. In this experiment, we generate 1400 images evenly across seven image styles and the models (Figure \ref{fig:color_spread}). We use text prompts to describe clothing for reference methods and visual prompts for our process. In each test sample, 221 human evaluators were randomly shown a sample from each model and asked to pick one that best matches the text prompt. The majority, 55\% of 700 responses (full table included in appendix), prefer our samples, proving overall superiority in image quality and visual control.

\section{Ablations}

\begin{figure}[!htb]
    \centering
    \begin{subfigure}[b]{0.14\columnwidth}
        \includegraphics[width=1.0\linewidth]{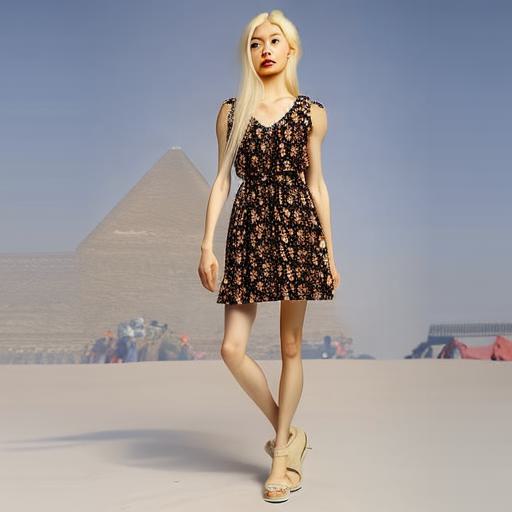}
        \caption{}
        \label{fig:forget1:a}
    \end{subfigure}
    \begin{subfigure}[b]{0.14\columnwidth}
        \includegraphics[width=1.0\linewidth]{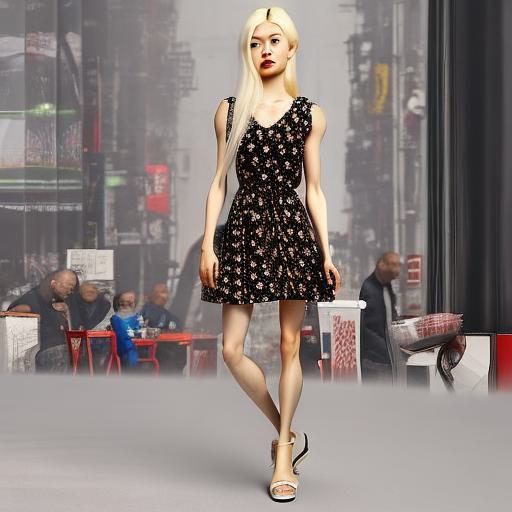}    
        \caption{}
    \end{subfigure}
    \begin{subfigure}[b]{0.14\columnwidth}
        \includegraphics[width=1.0\linewidth]{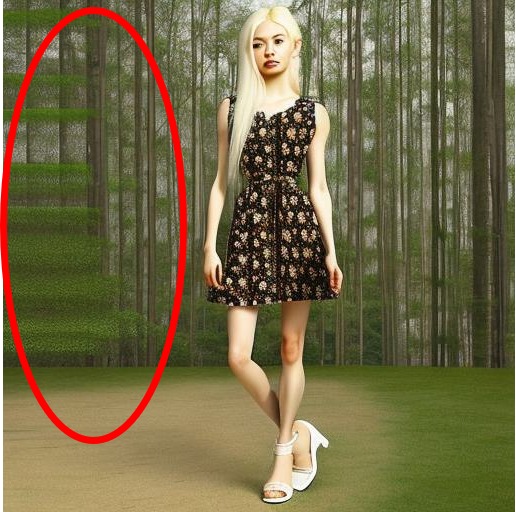}
        \caption{}
        \label{fig:forget1:c}
    \end{subfigure}
    \begin{subfigure}[b]{0.1\columnwidth}
        \includegraphics[width=1.0\linewidth]{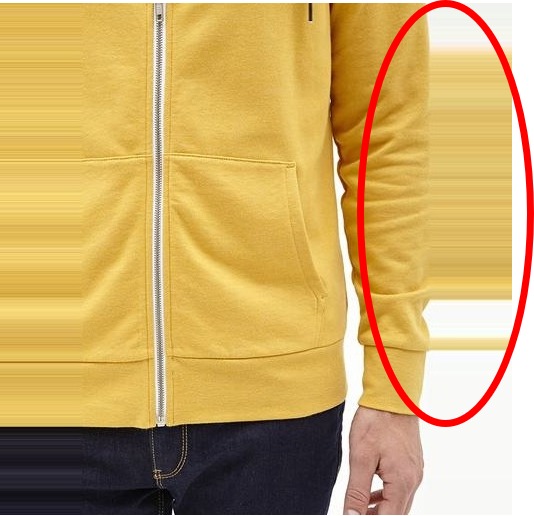}
        \caption{}
        \label{fig:forget1:artifact}
    \end{subfigure}    
    \begin{subfigure}[b]{0.14\columnwidth}
        \includegraphics[width=1.0\linewidth]{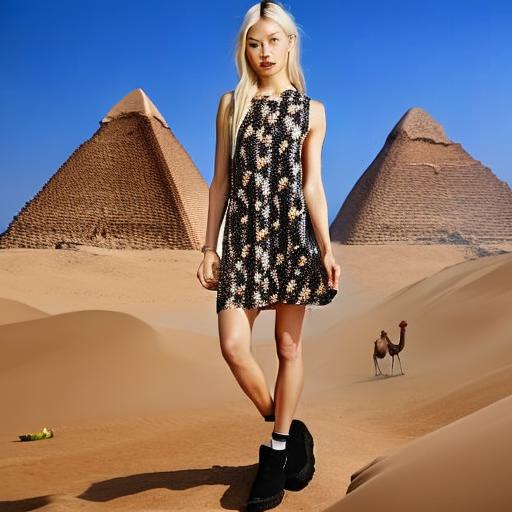}    
        \caption{}
        \label{fig:forget1:e}
    \end{subfigure}
    \begin{subfigure}[b]{0.14\columnwidth}
        \includegraphics[width=1.0\linewidth]{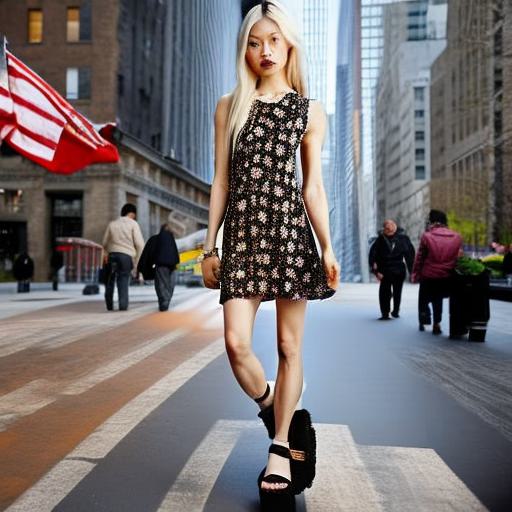}        
        \caption{}
    \end{subfigure}    
    \begin{subfigure}[b]{0.14\columnwidth}
        \includegraphics[width=1.0\linewidth]{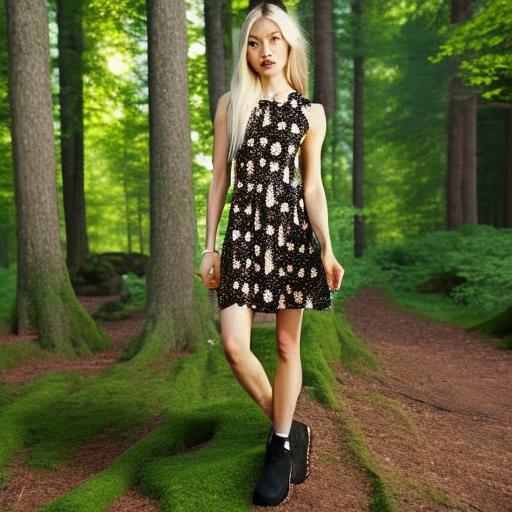}
        \caption{}
        \label{fig:forget1:g}
    \end{subfigure}            
    \caption{Without feature masking in training :(a)-(c) pale, dull background (d) padding artifact from dataset.
            Using feature masking in training: (e)-(g) vibrant colors and rich background. }
    \label{fig:forget1}
\end{figure}
\textbf{Necessity of Feature Mask in Training.}
\label{sec:ablation:mask}
Catastrophic forgetting can be demonstrated using the DeepFashion dataset; in our initial experiments, we applied feature masking to the training loss function but excluded it from the control signals. However, applying the feature mask post-training is ineffective, as shown by the pale and dull background in Figure \ref{fig:forget1:a} -\ref{fig:forget1:c}. In particular, the artifact (circled in red) in Figure \ref{fig:forget1:c} gives a clear indication of leakage of background originating from padding artifact uniquely caused by our dataset pre-processing error as shown in Figure \ref{fig:forget1:artifact}. Evidently, our method of applying feature masking in training produces a vibrant background (Figure \ref{fig:forget1:e}-\ref{fig:forget1:g}), demonstrating that our method is effective in avoiding catastrophic forgetting.

\begin{figure}[!htb]
    \centering
    \includegraphics[width=1\linewidth]{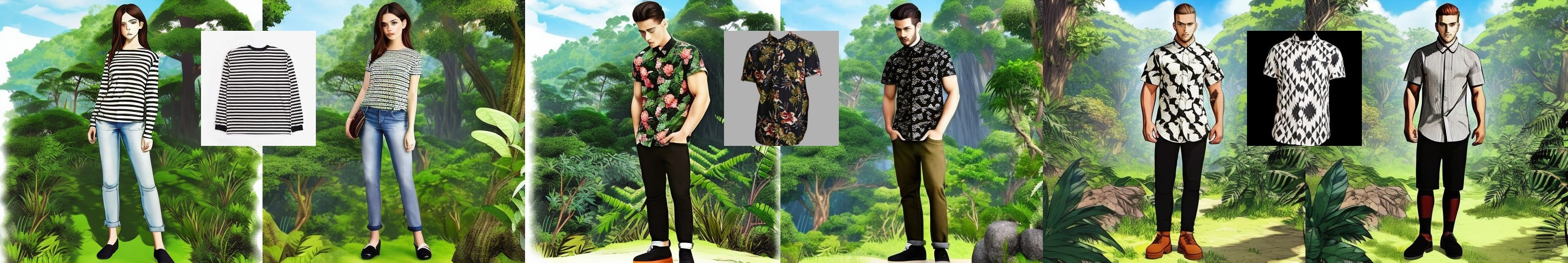}
    \caption{Local CLIP image embedding used in our method can capture fine texture details. (left) local embedding (mid) visual prompt (right) global embedding.}
    \label{fig:supp:clip}
\end{figure}

\textbf{CLIP Local Image Embedding Captures Fine Texture.}
We experimented with two image embedding methods for visual conditioning - global and local CLIP image embedding. Figure \ref{fig:supp:clip} shows that local CLIP embedding used in our method is better at capturing fine texture details. 

\section{Limitations}
\label{sec:limitation}
Like IP-Adapter, the clothing color in generated images is shaped by the inherent randomness in the initialized latent variables of the LDM backbone. While visual prompting proves effective with our method, attaining consistent and faithful image reconstruction necessitates careful selection of random seeds. On the other hand, by design choice, our model's visual prompting method learns only the foreground people and leaves the background generation to the LDM backbone. Conversely, pose transfer requires perfect reconstruction of the image background solely from the reference image. Consequently, we refrained from conducting a large-scale evaluation in virtual try-on and pose transfer tasks. Nevertheless, through careful random seed selection, we can still generate high-quality virtual try-on, pose transfer, and face swap images, as included in the appendix.
\section{Conclusions}

We present ViscoNet, a pioneering approach that seamlessly integrates visual control into a spatial adapter. Our method, characterized by a single branch handling both pose and visual control stands out for its lightweight design and significantly smaller footprint when compared to existing two-adapter solutions. Through a comprehensive blend of qualitative and quantitative assessments, we demonstrate the remarkable efficacy of ViscoNet in seamlessly bridging and harmonizing text and visual prompts. This unique capability not only mitigates mode collapse but also empowers the model to excel across diverse tasks, positioning it as one of the most versatile human image generation models available.

Furthermore, our feature masking technique significantly contributes to our model's strength by preserving the generative power of the backbone image model. Remarkably, this is achieved despite training exclusively on a human image dataset that is orders of magnitude smaller than the datasets used by reference methods. This underscores the efficiency and generalization prowess of ViscoNet in handling image generation tasks with limited training data.

\clearpage  

%
%
\bibliographystyle{splncs04}
\bibliography{main}

\title{\textbf{Appendix}:\\ \tcolor{ViscoNet}: Bridging and Harmonizing \tcolor{Vis}ual and Textual Conditioning for \tcolor{Co}ntrol\tcolor{Net}}
\titlerunning{ViscoNet}
\maketitle
\appendix
\setcounter{figure}{14}

\noindent This appendix is split into 3 sections:
\begin{itemize}
    \item \textbf{Section \ref{sec:appendix:re_id}} provides a further quantitative and qualitative comparison with IP-Adatper in the re-identification task (Section 4.2). 
    \item \textbf{Section \ref{sec:supp:qualitative}} showcases more image examples produced by our methods in a variety of tasks, including re-identification, pose re-target, fashion virtual try-on, and stylization. 
    \item \textbf{Section \ref{sec:supp:quantitative}} includes the quantitative results from the main paper with further analysis (Section 4.1, 4.3). 
\end{itemize}

\section{Re-identification: Comparing IP-Adapter}
\label{sec:appendix:re_id}
In the Section 4.2 experiment, we utilized 7 male celebrities in the text prompt, 6 reference clothing items, and 9 control strengths to generate 10 samples per control strength, resulting in a total of 7560 images. The quantitative result corresponds to Figure 11 and is listed in Table \ref{table:celeb}. We will elaborate on the quantitative findings in conjunction with the qualitative results.
\subsection{Control Strength Analysis}
\begin{table}
\centering
    \begin{adjustbox}{width=0.8\columnwidth}
    \begin{tabular}{l |ccccccccc}
    \toprule
    Strength & 0.0 & 0.1 & 0.2 & 0.3 & 0.4 & 0.5 & 0.6 & 0.8 & 1.0 \\
    \midrule
    \multicolumn{1}{c|}{} & \multicolumn{9}{c}{\underline{CLIP score}} \\
    IP-Adapter & 0.3066 & 0.3052 & 0.3003 & 0.2910 & 0.2729 & 0.2546 & 0.2231 & 0.1687 & 0.1606\\
    Method 1 (ours) & 0.3223 & 0.3230 & 0.3218 & 0.2993 & 0.2139 & 0.1846 & 0.1768 & 0.1670 & 0.1628\\
    Method 2 (ours) & \textbf{0.3232} & \textbf{0.3254} & \textbf{0.3250} & \textbf{0.3218} & \textbf{0.3196} & \textbf{0.3064} & \textbf{0.2920} & \textbf{0.2544} & \textbf{0.2312} \\

    \midrule
    \multicolumn{1}{c|}{} & \multicolumn{9}{c}{\underline{Mode Collapse Rate (MCR)}} \\
    IP-Adapter & 0.05 & 0.06 & 0.11 & 0.19 & 0.32 & 0.45 & 0.67 & 0.99 & 1.00\\
    Method 1 (ours) & \textbf{0.00} & \textbf{0.00} &  \textbf{0.00} & 0.12 & 0.74 & 0.97 & 1.00 & 1.00 & 1.00\\
    Method 2 (ours) & \textbf{0.00} & \textbf{0.00} & \textbf{0.00} & \textbf{0.00} & \textbf{0.00} & \textbf{0.08} & \textbf{0.15} & \textbf{0.40} & \textbf{0.57} \\
    \midrule
    \multicolumn{1}{c|}{} & \multicolumn{9}{c}{\underline{MS-SSIM}} \\
    IP-Adapter & 0.1248 & 0.1420 & 0.1705 & 0.2033 & 0.2452 & 0.2888 & 0.3700 & 0.4447 & 0.5107\\
    Method 1 (ours) & \textbf{0.1514} & \textbf{0.1795} & \textbf{0.2312} & \textbf{0.3781} & \textbf{0.5043} & \textbf{0.5335} & \textbf{0.5498} & \textbf{0.5536} & \textbf{0.5409} \\    
    Method 2 (ours) & 0.1507 & 0.1627 & 0.1886 & 0.2538 & 0.3661 & 0.4387 & 0.4742 & 0.5173 & 0.5183\\
    \bottomrule
    \end{tabular}        
    \end{adjustbox}
    \vspace{3mm}
    \caption{Reduced control strength results in higher CLIP scores and accuracy, translating to less mode collapse.}
    \vspace{3mm}
    \label{table:celeb}
\end{table}

\begin{figure}[!htb]
    \begin{subfigure}[b]{1\linewidth}
        \centering
        \includegraphics[width=1\linewidth]{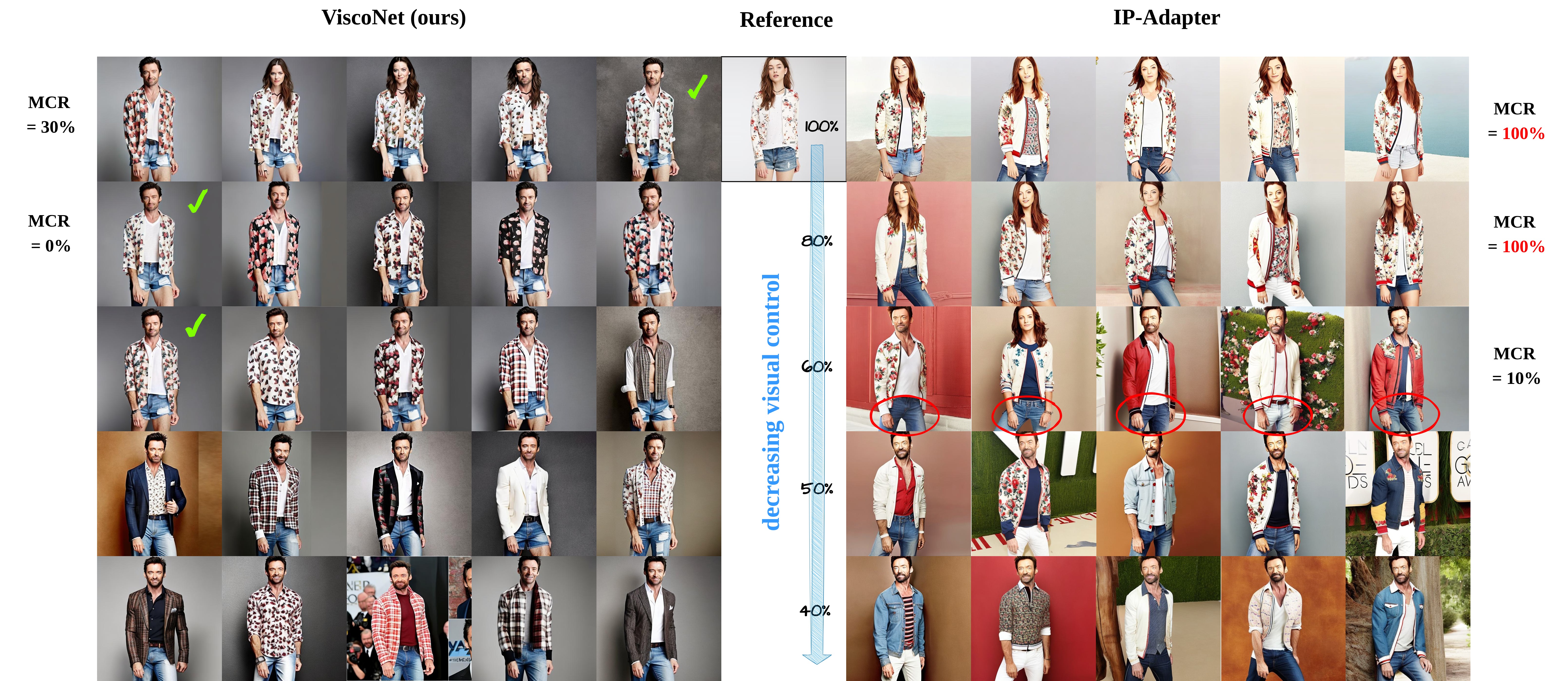}
        \caption{IP-Adapter suffer 100\% mode collapse at control strength over 80\% and unable to generate the target person \textit{Hugh Jackman}. Its visual conditioning power is much weaker when it finally escapes mode collapse at lower control strength, and \textcolor{red}{unable to generate the short pant}, and correct clothing style and color. In contrast, we are robust against mode collapse and avoid much of the problems above suffered by IP-Adapter, and able to generate desired results \textcolor{green}{51} at 100\% control strength, preserving faithfulness of both the person identity and clothing appearance.}
        \label{fig:jackman2}
    \end{subfigure}
    \begin{subfigure}[b]{1\linewidth}
        \centering
        \includegraphics[width=1\linewidth]{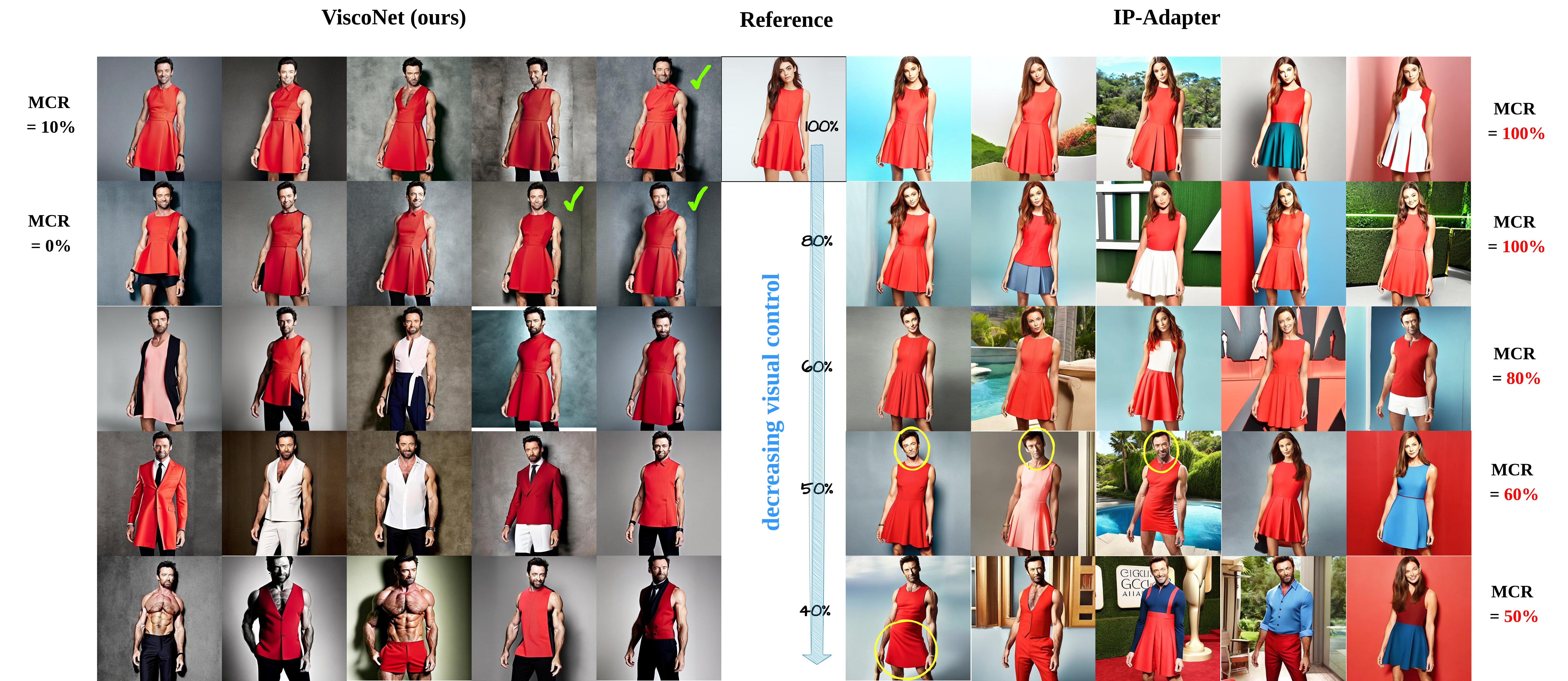}
        \caption{The conflict between the feminine reference image and Hugh Jackman's masculine image creates more conflict and hence mode collapse as suffered by IP-Adatper. IP-Adatper struggles to generate correct faces and pleated dress patterns (circled in yellow) at weaker control strength. This does not affect our method. }
        \label{fig:jackman4}
    \end{subfigure}
    \caption{Comparing the effect of control strength on re-identification task. IP-Adapter suffers much more severe mode collapse and struggles to create perfect image balancing the reference image and text prompt of \textit{Hugh Jackman}.}
    \label{fig:jackman24}
\end{figure}

In Figure \ref{fig:jackman24}, we show experiment samples of text prompt \textit{Hugh Jackman} and reference image 2, where we randomly sample 50\% of the samples for various control strengths from both our and IP-Adapter. With 100\% strength, although IP-Adapter can reconstruct the reference image, it suffers 100\% 

Overall, our method is effectively mode collapse free at 60\% while IP-Adapter still \ref{table:celeb} has 67\% MCR. As shown in Figure \ref{fig:jackman24}, although high control strength introduces some mode collapse to our method. However, we can still generate high-quality images, preserving visual conditioning and a person's identity.


\subsection{Further Quantitative Comparison}
\begin{figure}
\centering
\begin{subfigure}[b]{0.9\linewidth}
    \centering
    \includegraphics[width=1\linewidth]{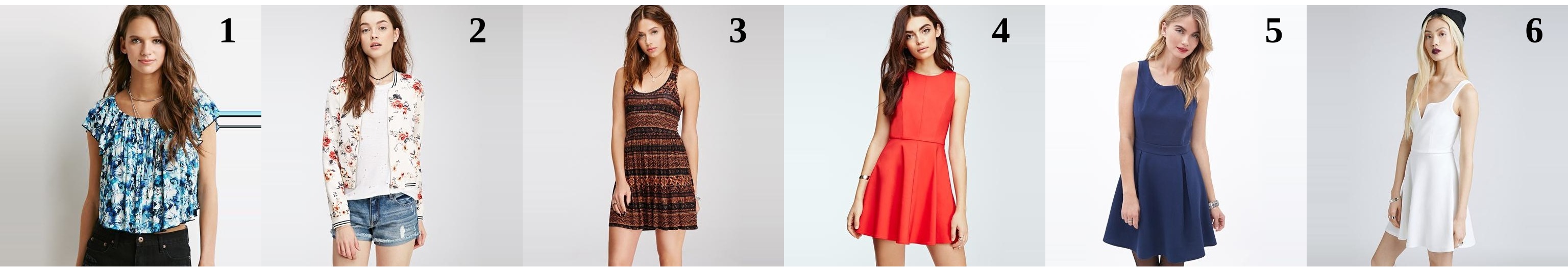}
    \caption{Visual reference taken from the unseen test dataset.}
\end{subfigure}
\begin{subfigure}[b]{0.9\linewidth}
    \centering
    \includegraphics[width=1\linewidth]{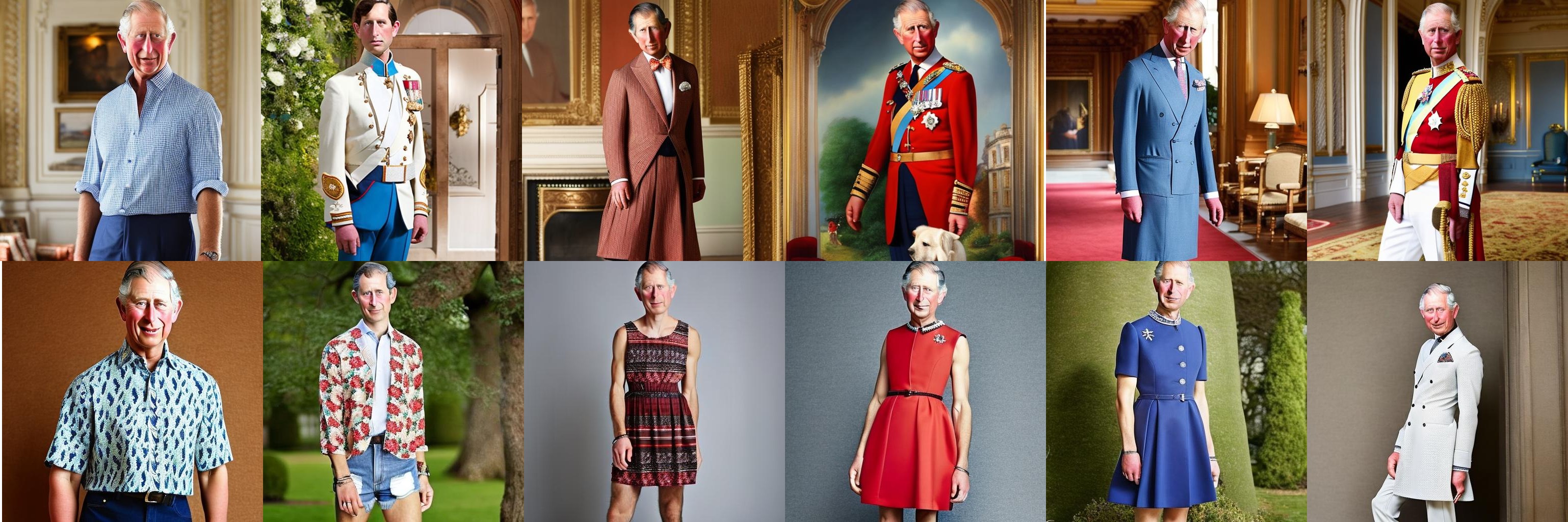}
    \caption{Unlike other movie stars with more diverse costumes, Prince Charles' limited clothing range presents the toughest challenge. (Top) IP-Adatper cannot produce any image of Prince Charles wearing the reference clothing. (Bottom) despite the extreme data gap, our method can produce reasonable images. }
    \label{fig:appendix2:charles}    
\end{subfigure}
\caption{Most challenging example in re-identification task - Prince Charles.}
\label{fig:appendix:reidentification:1}
\end{figure}

\begin{figure}
\centering
\begin{subfigure}[b]{0.9\linewidth}
    \centering
    \includegraphics[width=1\linewidth]{images/eccv/appendix/ref.jpg}
    \caption{Visual reference taken from the unseen test dataset.}
\end{subfigure}
\begin{subfigure}[b]{0.9\linewidth}
    \centering
    \includegraphics[width=1\linewidth]{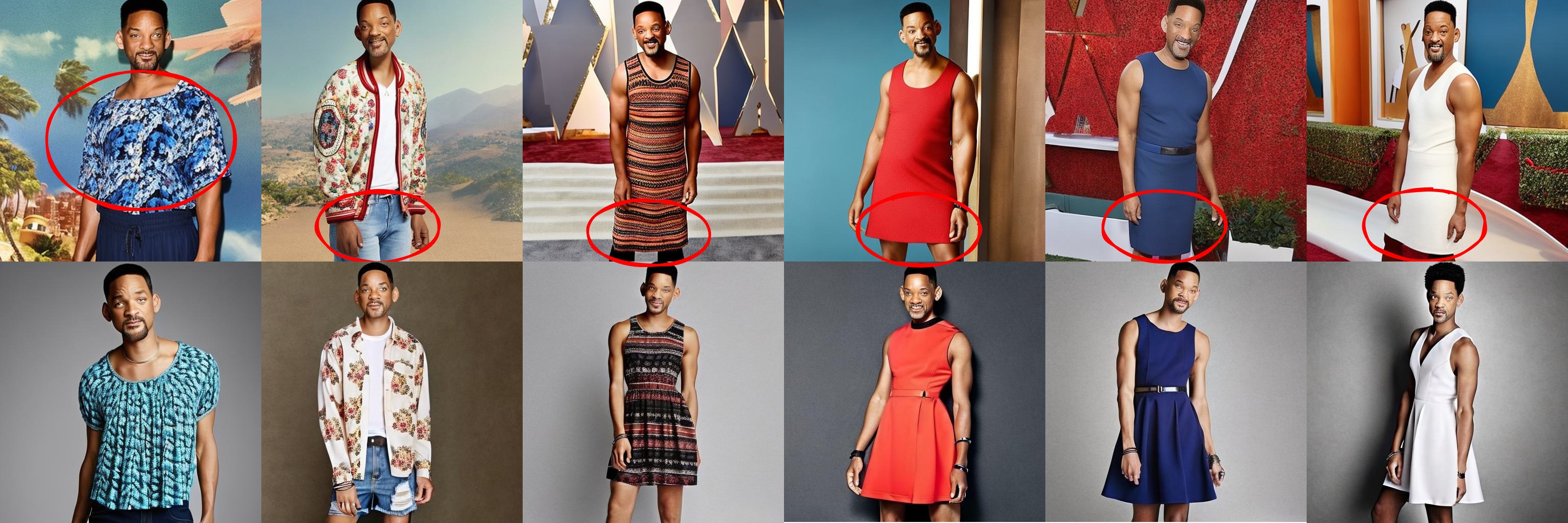}
    \caption{Will Smith: (top) IP-Adaptor showing incorrect clothing color, length, or style (no pleated dress pattern). (bottom)  Ours}
    \label{fig:appendix2:smith}    
\end{subfigure}
\begin{subfigure}[b]{0.9\linewidth}
    \centering
    \includegraphics[width=1\linewidth]{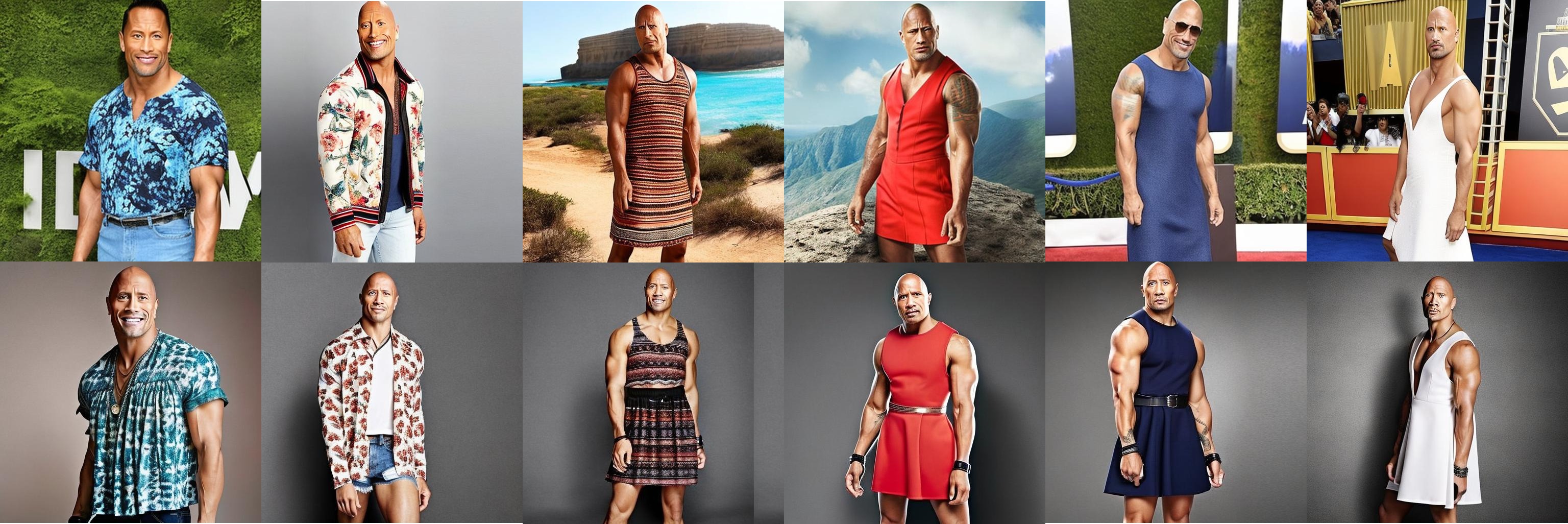}
    \caption{Dwayne Johnson: (top) IP-Adaptor  (bottom) Ours}
    \label{fig:appendix2:rock}    
\end{subfigure}
\begin{subfigure}[b]{0.88\linewidth}
    \centering
    \includegraphics[width=1\linewidth]{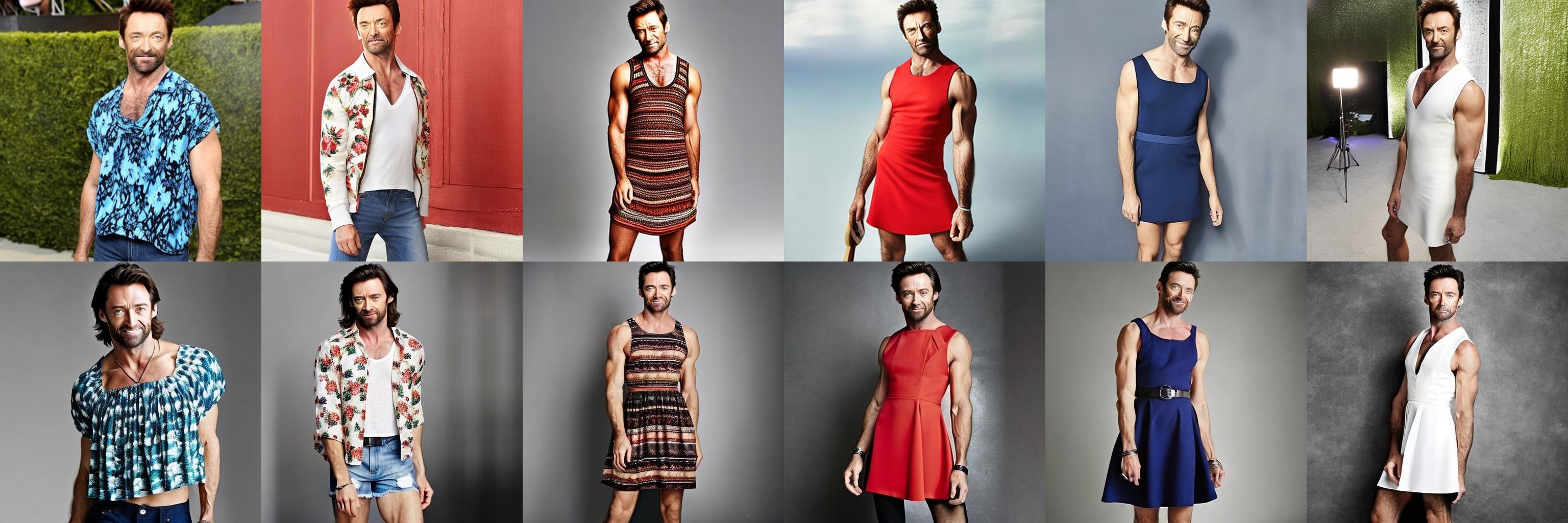}
    \caption{Hugh Jackman: (top) IP-Adaptor  (bottom) Ours.}
\end{subfigure}
\caption{Re-identification comparison with IP-Adapter.}
\label{fig:appendix:reidentification:2}
\end{figure}

\begin{figure}
\centering
\begin{subfigure}[b]{0.9\linewidth}
    \centering
    \includegraphics[width=1\linewidth]{images/eccv/appendix/ref.jpg}
    \caption{Visual reference taken from the unseen test dataset.}
    \label{fig:appendix2:ref}    
\end{subfigure}
\begin{subfigure}[b]{0.9\linewidth}
    \centering
    \includegraphics[width=1\linewidth]{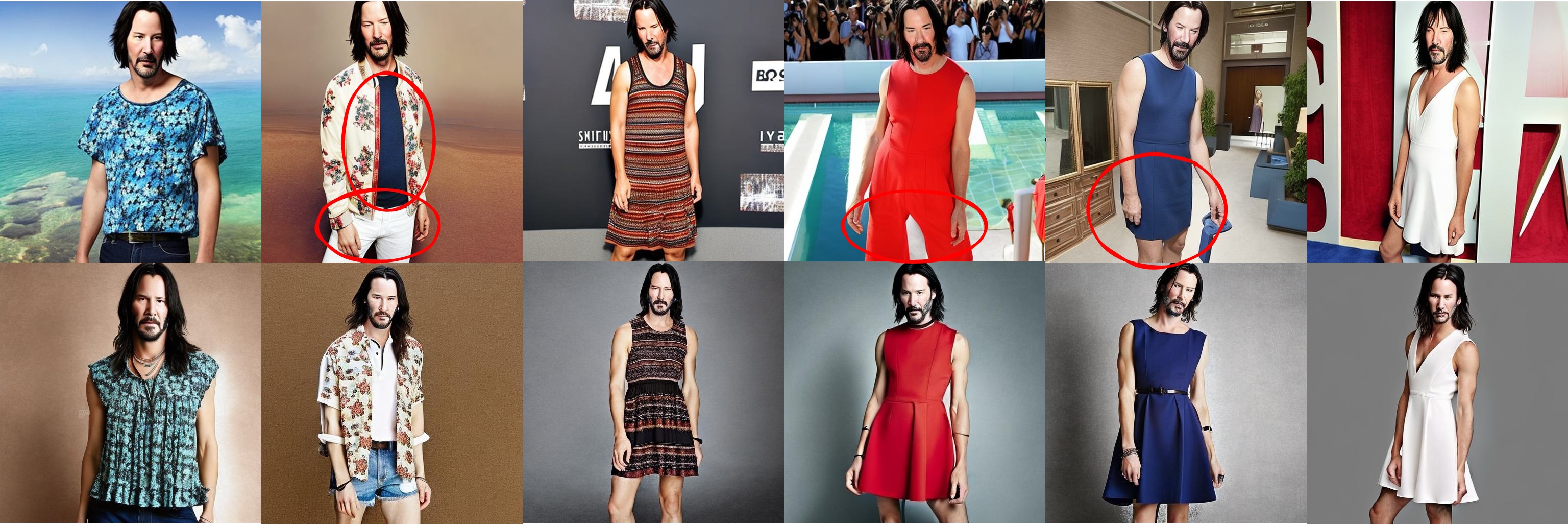}
    \caption{Keanu Reeves: (top) IP-Adaptor  (bottom) Ours.}
    \label{fig:appendix2:keanu}    
\end{subfigure}
\begin{subfigure}[b]{0.9\linewidth}
    \centering
    \includegraphics[width=1\linewidth]{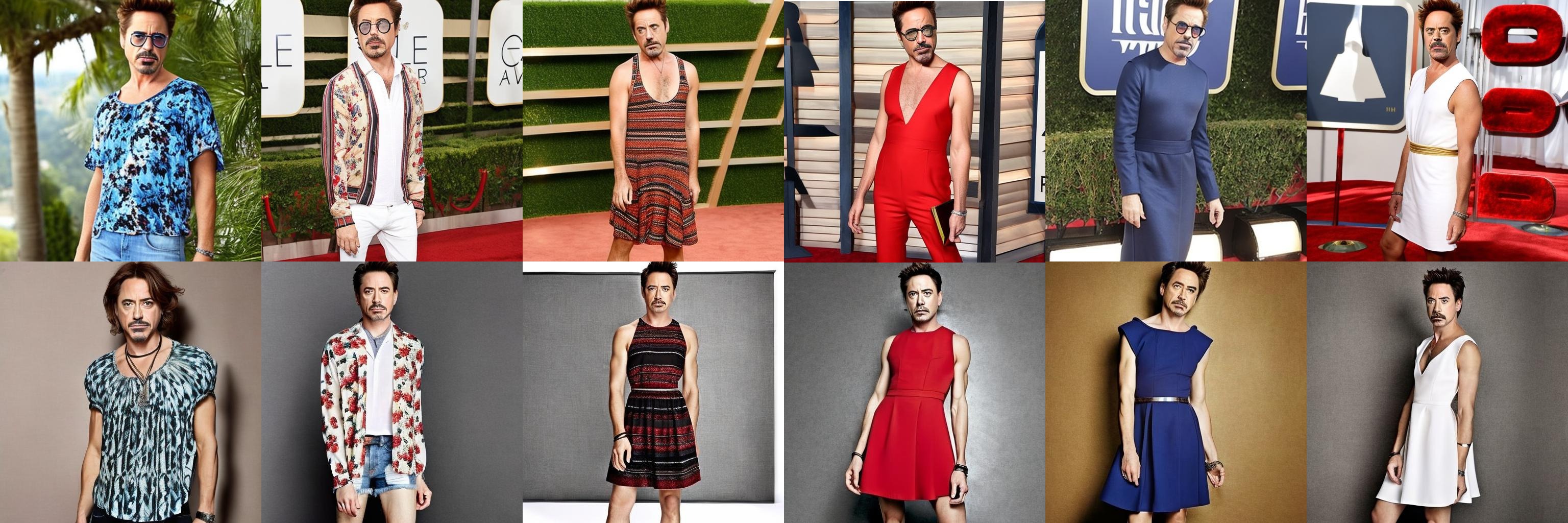}
    \caption{Robert Downey Jr.: (top) IP-Adaptor  (bottom) Ours}
    \label{fig:appendix2:downey}    
\end{subfigure}
\begin{subfigure}[b]{0.88\linewidth}
    \centering
    \includegraphics[width=1\linewidth]{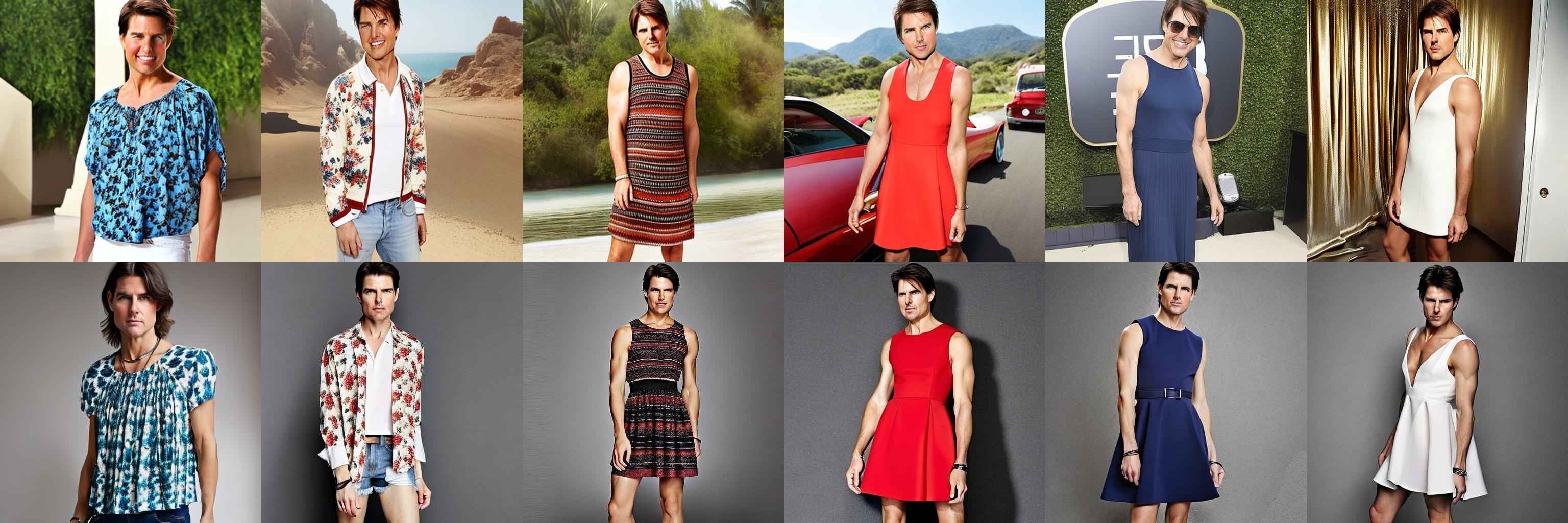}
    \caption{Tom Cruise: (top) IP-Adaptor  (bottom) Ours}
    \label{fig:appendix2:cruise}    
\end{subfigure}
\caption{Re-identification comparison with IP-Adapter.}
\label{fig:appendix:reidentification:3}
\end{figure}

 We further explore qualitative results in this section. Unlike Figure 7-10, where we slide along the control strength on the same random seed to demonstrate latent space discontinuity, we extend Section \ref{sec:appendix:re_id} to present the best samples across all control strengths from both methods for direct comparison, as shown in Figure \ref{fig:appendix:reidentification:1}-\ref{fig:appendix:reidentification:3}.

Among all the celebrities mentioned in the text prompt, \textit{Prince Charles}\footnote{Stable Diffusion was trained on dated data before Prince Charles ascended to be king, so we adhere to his old title in the experiment.} - known for having a limited wardrobe of formal attires in public images - presents the greatest challenge to the generalization capability of the models. IP-Adapter encounters difficulties and fails to generate any image of Prince Charles in casual or feminine clothing, as depicted in the reference image (Figure \ref{fig:appendix:reidentification:1}). In contrast, our method achieves reasonable success despite the monumental challenge. Figure \ref{fig:appendix:reidentification:2} - \ref{fig:appendix:reidentification:3} shows samples from the rest of the text prompts used in the experiment. Overall, IP-Adapter needs to have much-lowered control strength to escape mode collapse, resulting in loss of fidelity in clothing to the reference images, including the incorrect length of pants or dress, wrong color and pattern, i.e., loss of the pleated dress pattern, it previously able to generate (Figure \ref{fig:jackman4}).

\begin{figure}
    \centering
    \includegraphics[width=1\linewidth]{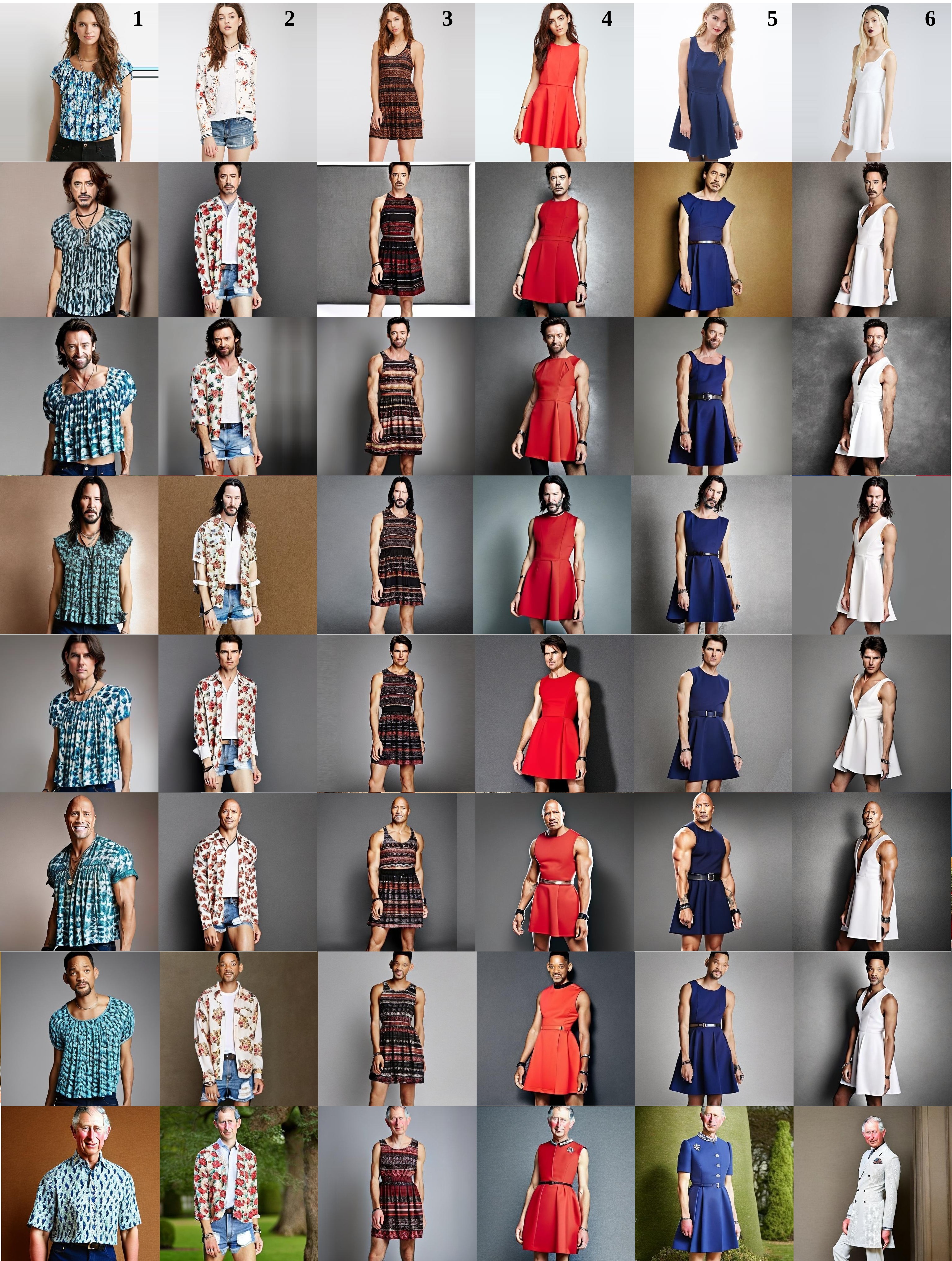}
    \caption{Putting our images together shows the consistency of our method in delivering celebrity re-identification.}
    \label{fig:appendix:consistentl}
\end{figure}

\clearpage

\section{Versatile Human Image Generation Task}
\label{sec:supp:qualitative}
\subsection{Re-identification (visual prompt)}
\begin{figure*}[!ht]
\centering
\centering
    \begin{subfigure}[]{0.8\linewidth}    
    \centering
        \begin{subfigure}[b]{0.31\linewidth}
            \includegraphics[width=1.0\linewidth]{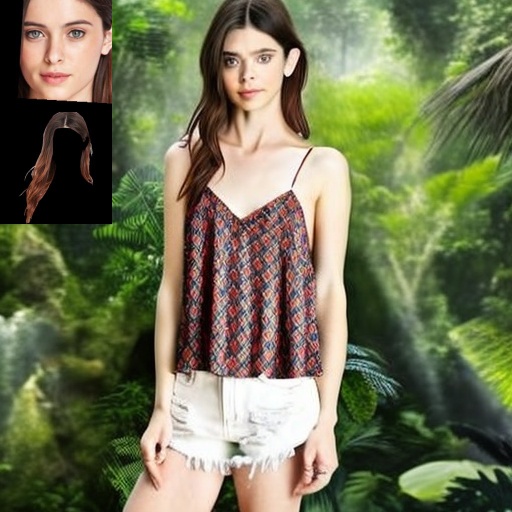}
        \end{subfigure}
        \begin{subfigure}[b]{0.31\linewidth}
            \includegraphics[width=1.0\linewidth]{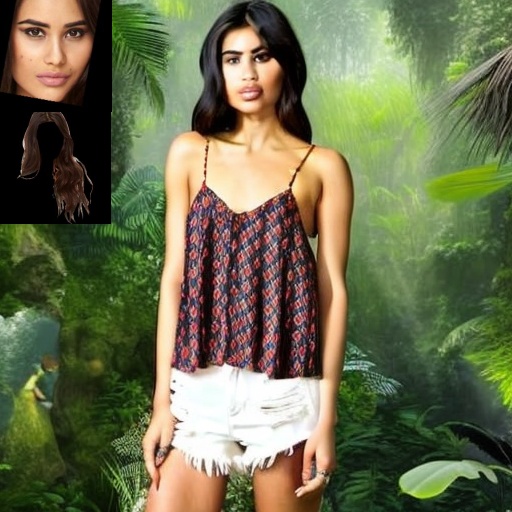}
        \end{subfigure}
        \begin{subfigure}[b]{0.31\linewidth}
            \includegraphics[width=1.0\linewidth]{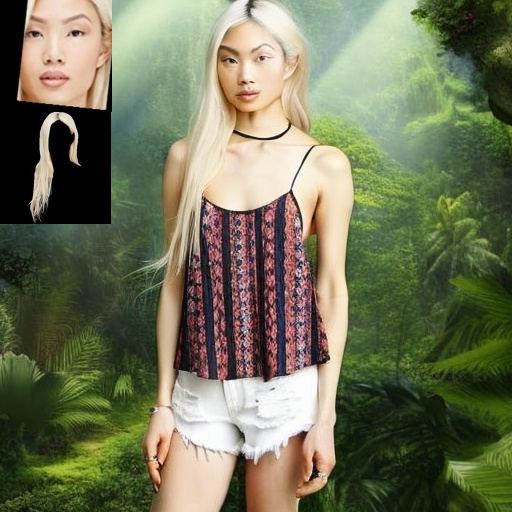}
        \end{subfigure}
    \end{subfigure}    
    \begin{subfigure}[]{0.8\linewidth}
    \centering
        \begin{subfigure}[b]{0.31\linewidth}
            \includegraphics[width=1.0\linewidth]{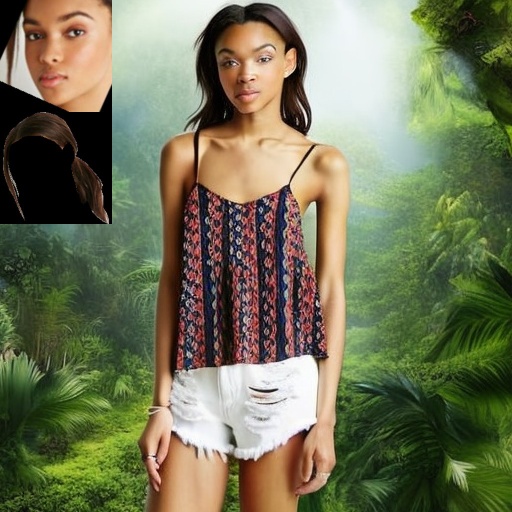}
        \end{subfigure}    
        \begin{subfigure}[b]{0.31\linewidth}
            \includegraphics[width=1.0\linewidth]{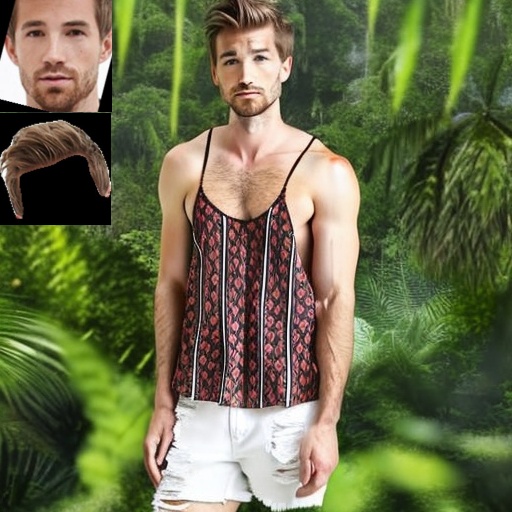}
        \end{subfigure}  
        \begin{subfigure}[b]{0.31\linewidth}
            \includegraphics[width=1.0\linewidth]{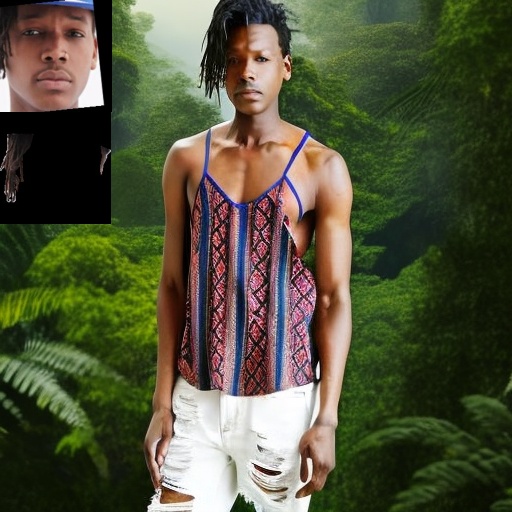}
        \end{subfigure}          
    \end{subfigure}
    \caption{Re-identification with a visual prompt.}
    \label{fig:supp:deepfakes:1}
\end{figure*}

Figure \ref{fig:supp:deepfakes:1} shows by conditioning on face and hair images, our method generates realistic people with diverse skin tones and body shapes correctly matching the faces despite the DeepFashion dataset consisting of more than 90\% of female images, predominately fair-skinned women. 

\subsection{Stylization}
\begin{figure*}[!ht]
\centering
\begin{subfigure}[b]{1.0\linewidth}
    \centering
    \begin{subfigure}[b]{0.24\linewidth}
        \includegraphics[width=1.0\linewidth]{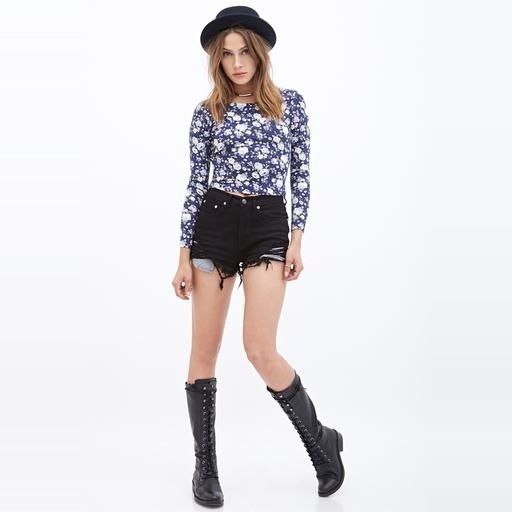}
        \caption{Reference}
    \end{subfigure}
    \begin{subfigure}[b]{0.24\linewidth}
        \includegraphics[width=1.0\linewidth]{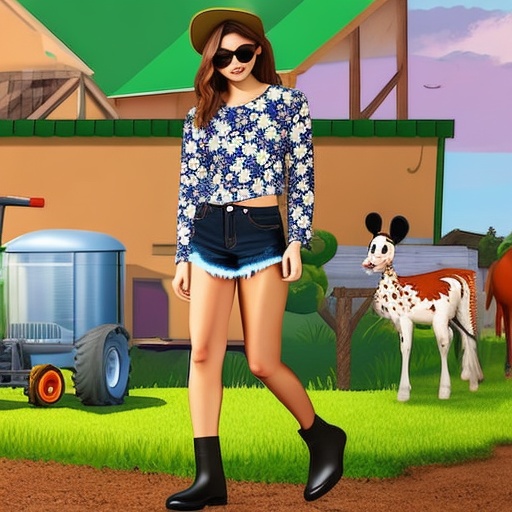}
        \caption{Cartoon}
    \end{subfigure}
    \begin{subfigure}[b]{0.24\linewidth}
        \includegraphics[width=1.0\linewidth]{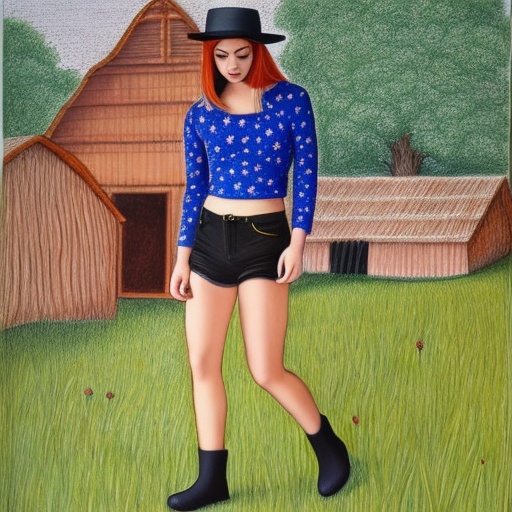}
        \caption{Color sketch}
    \end{subfigure}
    \begin{subfigure}[b]{0.24\linewidth}
        \includegraphics[width=1.0\linewidth]{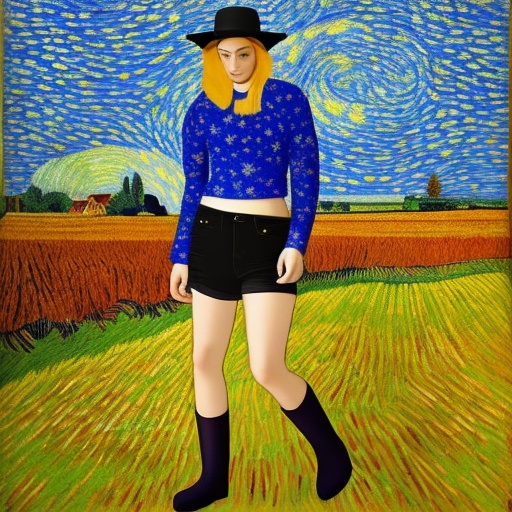}
        \caption{Van Gogh}    
    \end{subfigure}    
\end{subfigure} 
\begin{subfigure}[b]{1.0\linewidth}
    \centering    
    \begin{subfigure}[b]{0.24\linewidth}
        \includegraphics[width=1.0\linewidth]{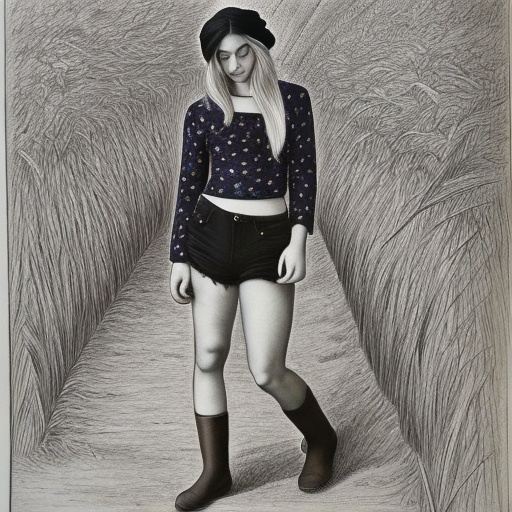}
        \caption{Pencil sketch}
    \end{subfigure}      
    \begin{subfigure}[b]{0.24\linewidth}
        \includegraphics[width=1.0\linewidth]{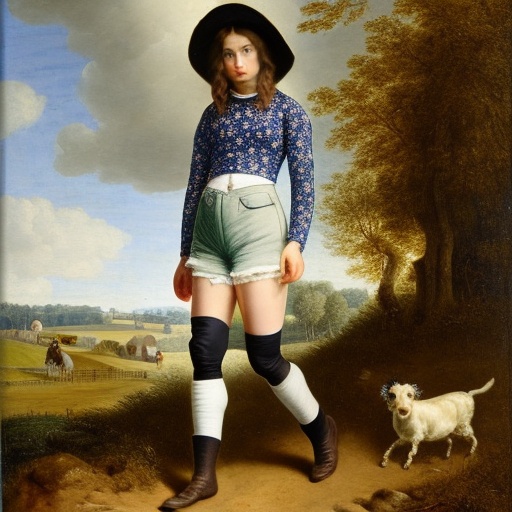}
        \caption{Portrait de Messieurs}
    \end{subfigure}    
    \begin{subfigure}[b]{0.24\linewidth}
        \includegraphics[width=1.0\linewidth]{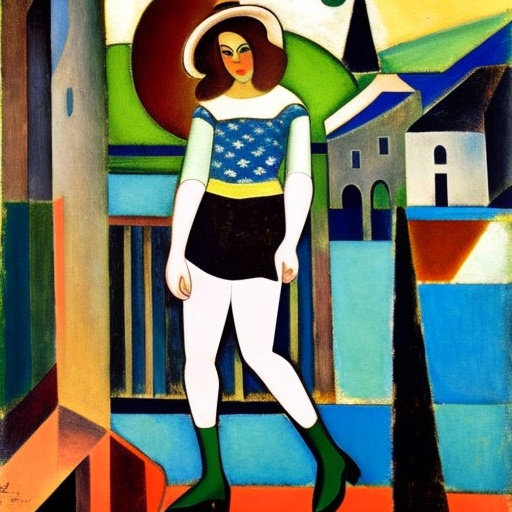}
        \caption{Cubism Art}
    \end{subfigure}         
    \begin{subfigure}[b]{0.24\linewidth}
        \includegraphics[width=1.0\linewidth]{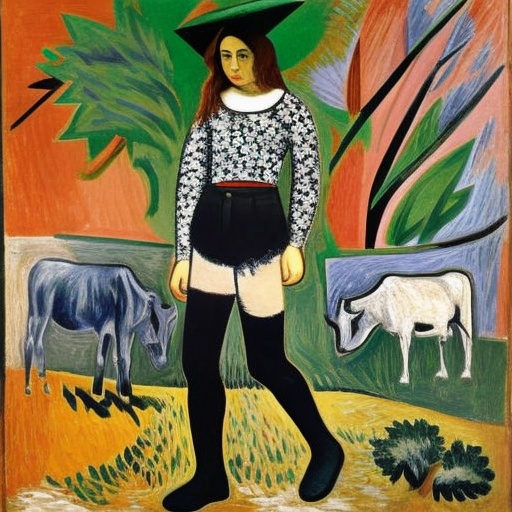}
        \caption{Picasso}
    \end{subfigure}    
\end{subfigure} 
\begin{subfigure}[b]{1.0\linewidth}
    \centering
    \begin{subfigure}[b]{0.24\linewidth}
        \includegraphics[width=1.0\linewidth]{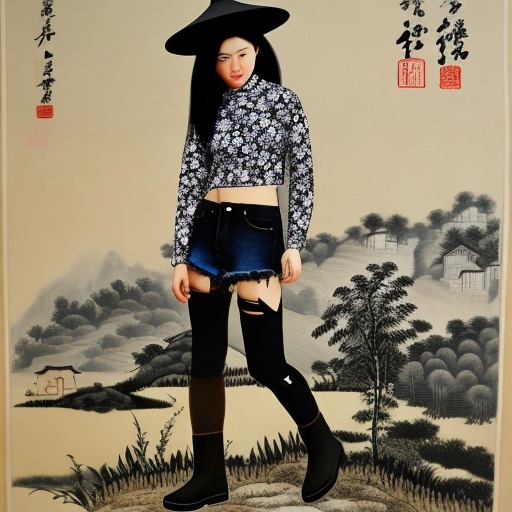}
        \caption{Chinese ink painting}
    \end{subfigure}           
    \begin{subfigure}[b]{0.24\linewidth}
        \includegraphics[width=1.0\linewidth]{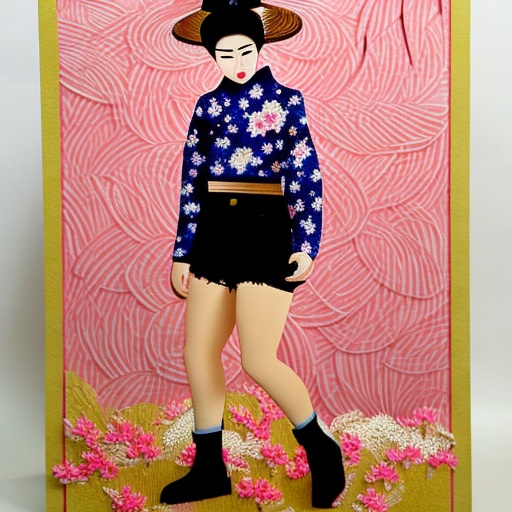}
        \caption{Japanese paper art}
    \end{subfigure}      
    \begin{subfigure}[b]{0.24\linewidth}
        \includegraphics[width=1.0\linewidth]{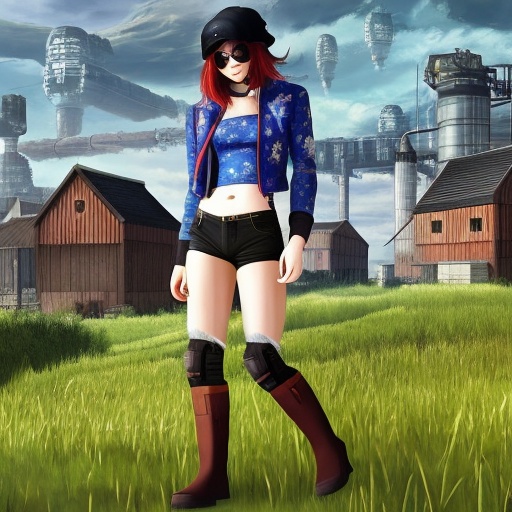}
        \caption{cyberpunk anime}
    \end{subfigure}    
    \begin{subfigure}[b]{0.24\linewidth}
        \includegraphics[width=1.0\linewidth]{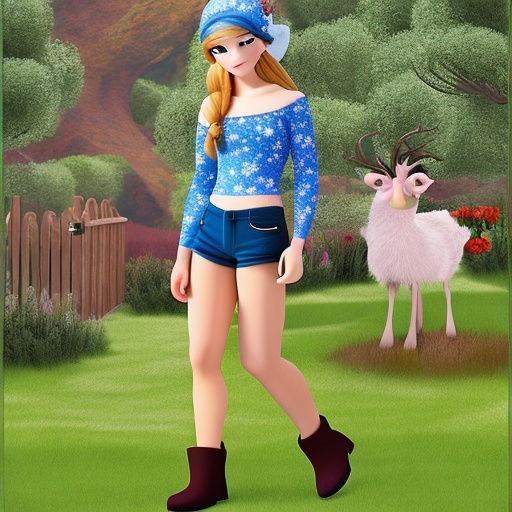}
        \caption{Disney's Frozen}
        \label{fig:style1:disney}
    \end{subfigure}    
\end{subfigure} 
\begin{subfigure}[b]{1.0\linewidth}
    \centering 
    \begin{subfigure}[b]{0.24\linewidth}
        \includegraphics[width=1.0\linewidth]{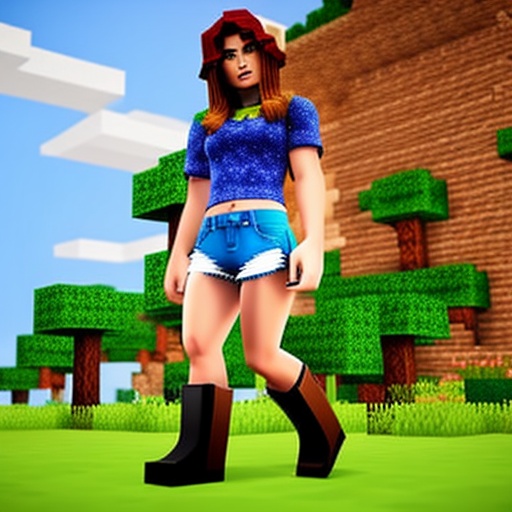}
        \caption{Minecraft}
    \end{subfigure}       
    \begin{subfigure}[b]{0.24\linewidth}
        \includegraphics[width=1.0\linewidth]{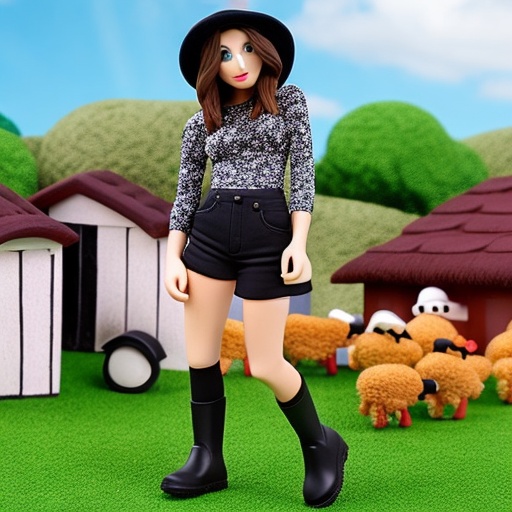}
        \caption{Shaun The Sheep}
    \end{subfigure}                 
    \begin{subfigure}[b]{0.24\linewidth}
        \includegraphics[width=1.0\linewidth]{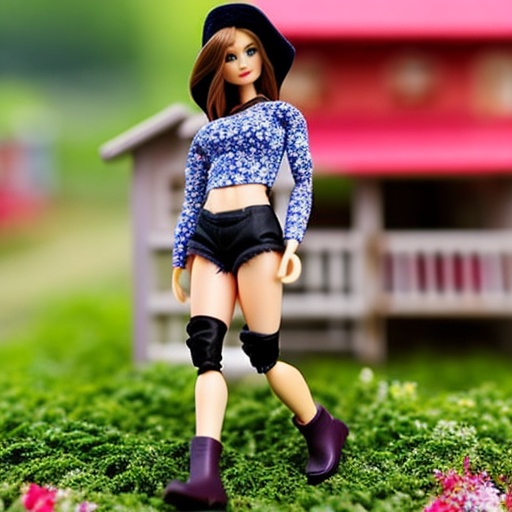}
        \caption{Barbie doll}
    \end{subfigure}
    \begin{subfigure}[b]{0.24\linewidth}
        \includegraphics[width=1.0\linewidth]{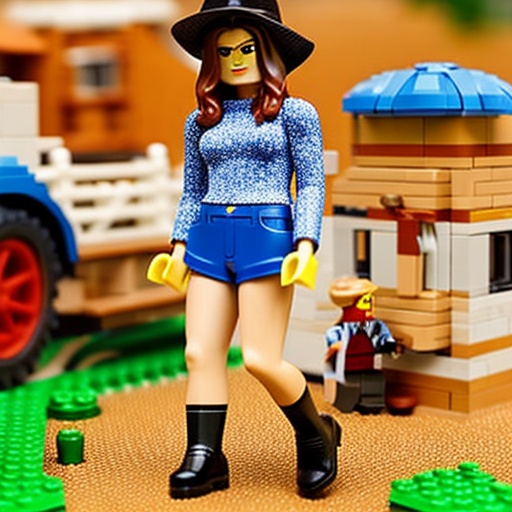}
        \caption{Lego}
    \end{subfigure}        
\end{subfigure} 
\caption{Stylization. Text prompt: \textit{``a woman, in farm.''}}
\label{fig:supp:style_1}
\end{figure*}

\begin{figure*}[!ht]
\centering
\begin{subfigure}[b]{1.0\linewidth}
    \centering
    \begin{subfigure}[b]{0.24\linewidth}
        \includegraphics[width=1.0\linewidth]{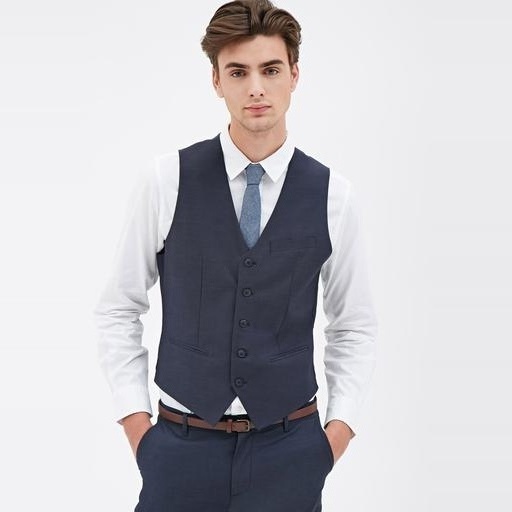}
        \caption{Reference}
    \end{subfigure}
    \begin{subfigure}[b]{0.24\linewidth}
        \includegraphics[width=1.0\linewidth]{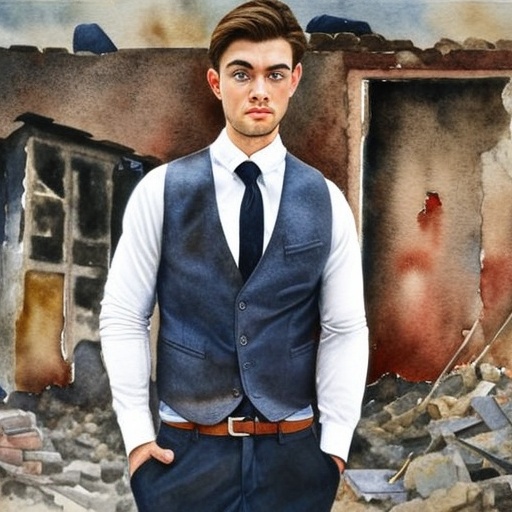}
        \caption{Watercolor}
    \end{subfigure}
    \begin{subfigure}[b]{0.24\linewidth}
        \includegraphics[width=1.0\linewidth]{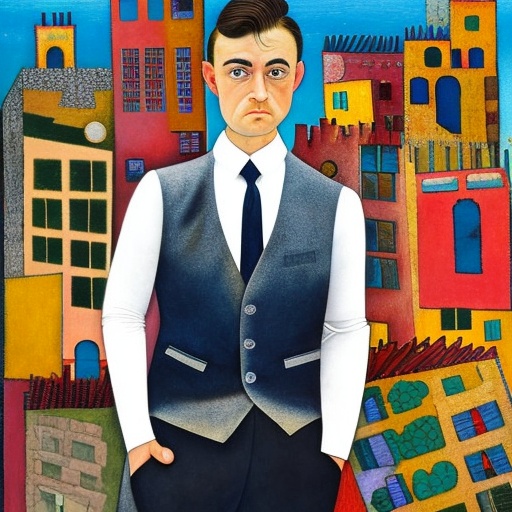}
        \caption{Expressionism}
    \end{subfigure}
    \begin{subfigure}[b]{0.24\linewidth}
        \includegraphics[width=1.0\linewidth]{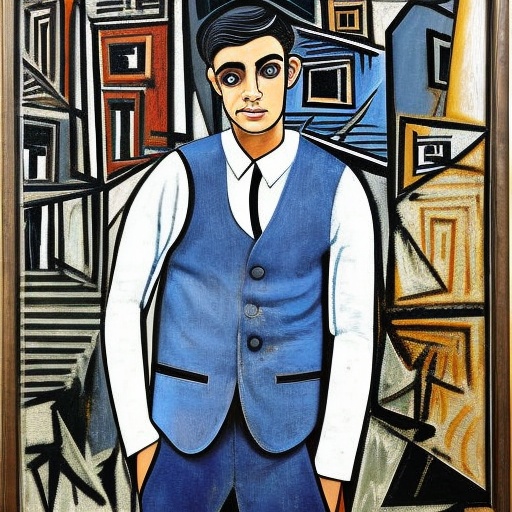}
        \caption{Picasso}    
    \end{subfigure}        
  
\end{subfigure} 
\begin{subfigure}[b]{1.0\linewidth}
    \centering   
    \begin{subfigure}[b]{0.24\linewidth}
        \includegraphics[width=1.0\linewidth]{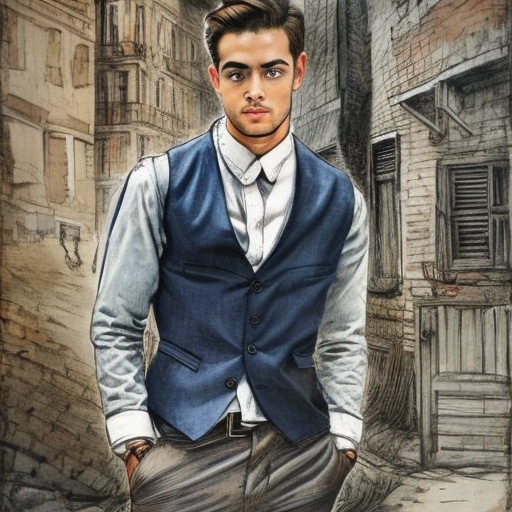}
        \caption{Sketch}
    \end{subfigure}        
    \begin{subfigure}[b]{0.24\linewidth}
        \includegraphics[width=1.0\linewidth]{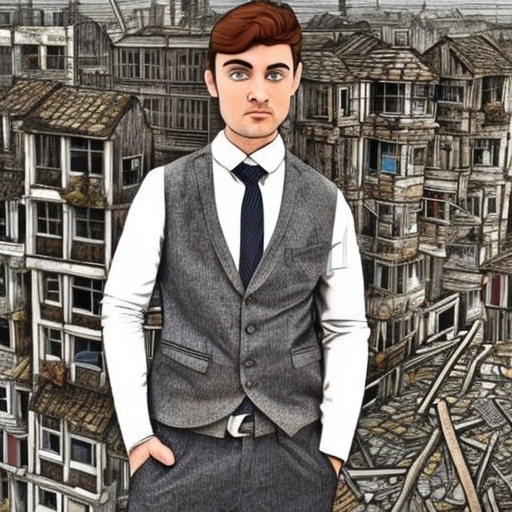}
        \caption{Children illustration}
    \end{subfigure}         
    \begin{subfigure}[b]{0.24\linewidth}
        \includegraphics[width=1.0\linewidth]{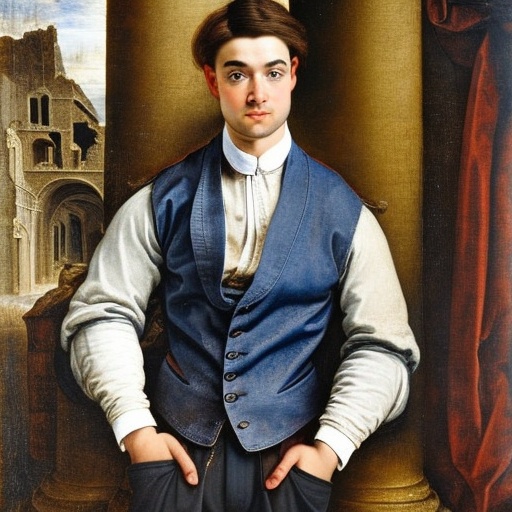}
        \caption{Renaissance Art}
    \end{subfigure}         
    \begin{subfigure}[b]{0.24\linewidth}
        \includegraphics[width=1.0\linewidth]{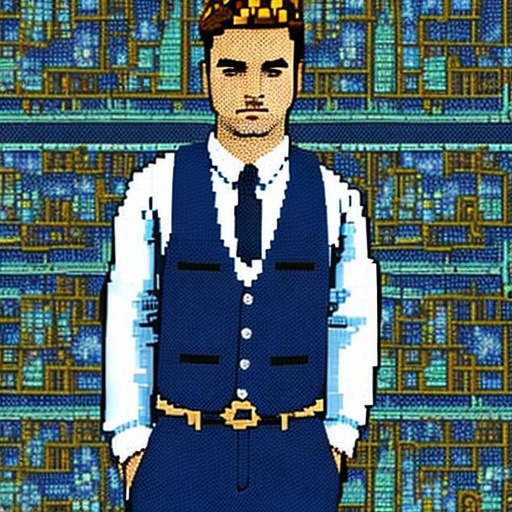}
        \caption{8-bit computer graphics}
    \end{subfigure}            
\end{subfigure} 
\begin{subfigure}[b]{1.0\linewidth}
    \centering    
    \begin{subfigure}[b]{0.24\linewidth}
        \includegraphics[width=1.0\linewidth]{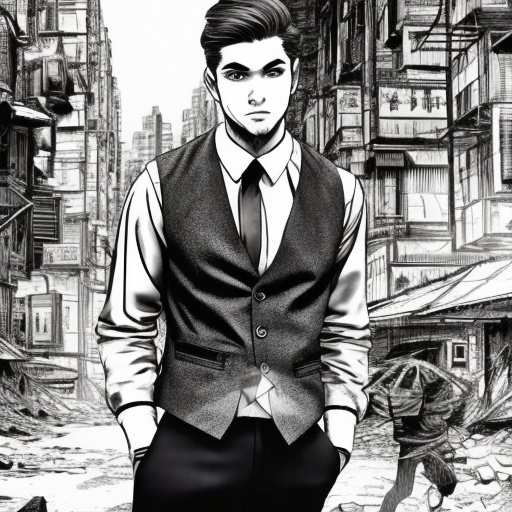}
        \caption{Black \& White Manga}
    \end{subfigure}         
    \begin{subfigure}[b]{0.24\linewidth}
        \includegraphics[width=1.0\linewidth]{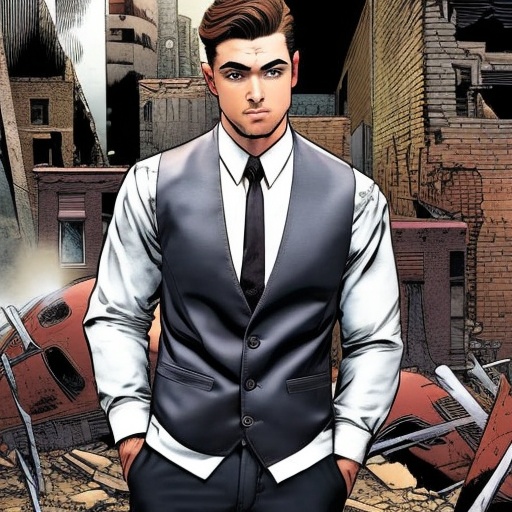}
        \caption{Marvel's comics}
    \end{subfigure}      
    \begin{subfigure}[b]{0.24\linewidth}
        \includegraphics[width=1.0\linewidth]{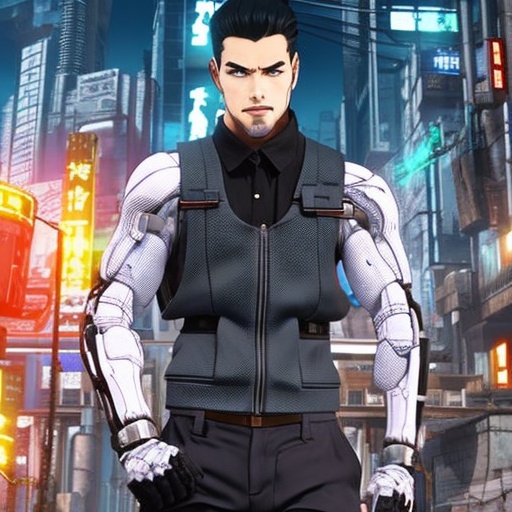}
        \caption{Cyborg, anime}
    \end{subfigure}        
    \begin{subfigure}[b]{0.24\linewidth}
        \includegraphics[width=1.0\linewidth]{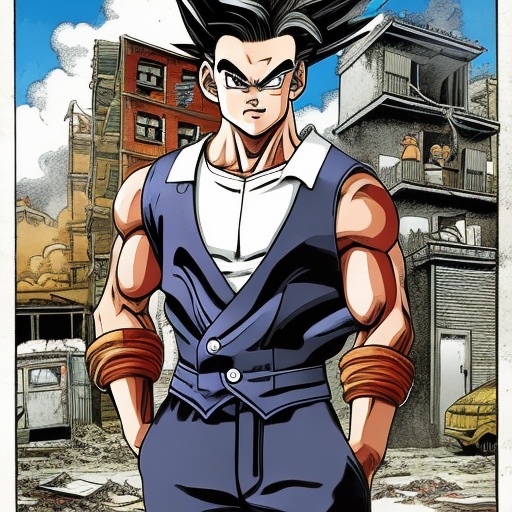}
        \caption{Dragonball}
        \label{fig:style12:dragonball}
    \end{subfigure}      
\end{subfigure}
\begin{subfigure}[b]{1.0\linewidth}
    \centering    
    \begin{subfigure}[b]{0.24\linewidth}
        \includegraphics[width=1.0\linewidth]{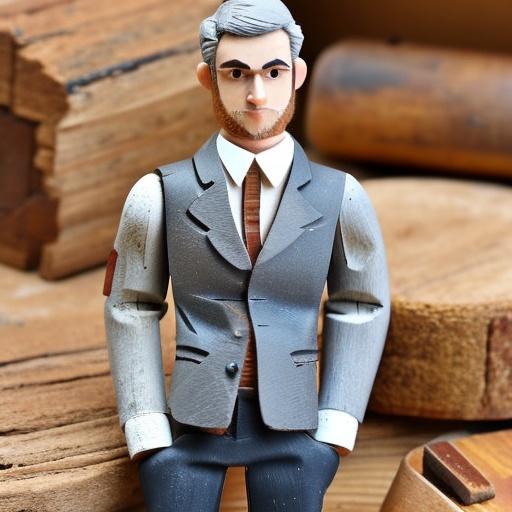}
        \caption{Wooden toy}
    \end{subfigure}
    \begin{subfigure}[b]{0.24\linewidth}
        \includegraphics[width=1.0\linewidth]{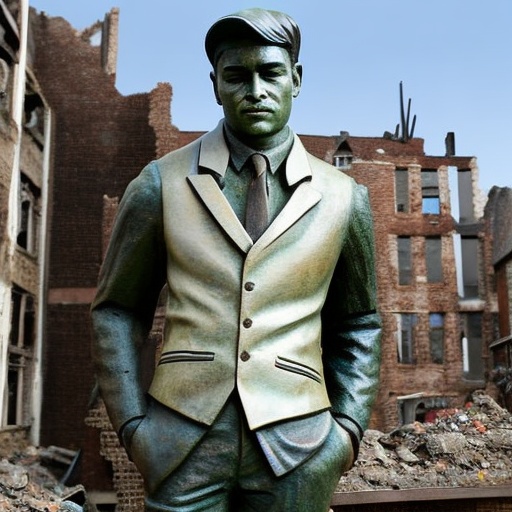}
        \caption{Stature}
    \end{subfigure}
    \begin{subfigure}[b]{0.24\linewidth}
        \includegraphics[width=1.0\linewidth]{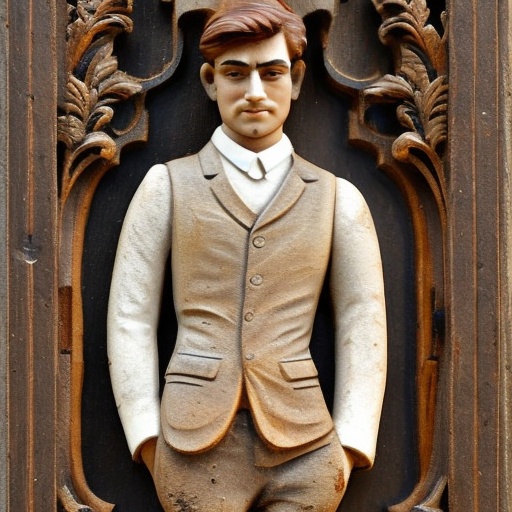}
        \caption{Wood Carving}
    \end{subfigure}          
    \begin{subfigure}[b]{0.24\linewidth}
        \includegraphics[width=1.0\linewidth]{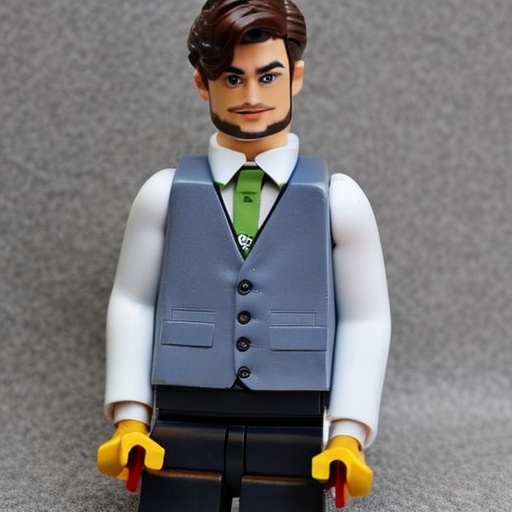}
        \caption{Lego}
    \end{subfigure}          
\end{subfigure} 
\caption{Stylization. Text prompt: \textit{``a man, in a derelict city.''}}
\label{fig:supp:style_2}

\end{figure*}

Figure \ref{fig:supp:style_1} and Figure \ref{fig:supp:style_2} show that our visual conditioning is effective across many image domains in creating a desired person's appearance, including various painting styles and also 3D objects such as statues, sculptures, toys, and 3D graphics. Some image domains have distinctive characteristics with considerable divergence from real photos, such as cartoons with disproportionate bigger heads, which can lead to a higher mode collapse rate. We circumvent this by removing the face mask to create results such as in Figure \ref{fig:style1:disney} and Figure \ref{fig:style12:dragonball}.

\clearpage
\subsection{Pose Re-target}
\begin{figure*}[!htb]
\centering
\begin{subfigure}[]{0.9\linewidth}
    \centering
    \begin{subfigure}[]{1.0\linewidth}
        \begin{subfigure}[b]{0.19\linewidth}
            \textcolor{white}{\rule{2cm}{2cm}}
        \end{subfigure}
        \begin{subfigure}[b]{0.19\linewidth}
            \includegraphics[width=1.0\linewidth]{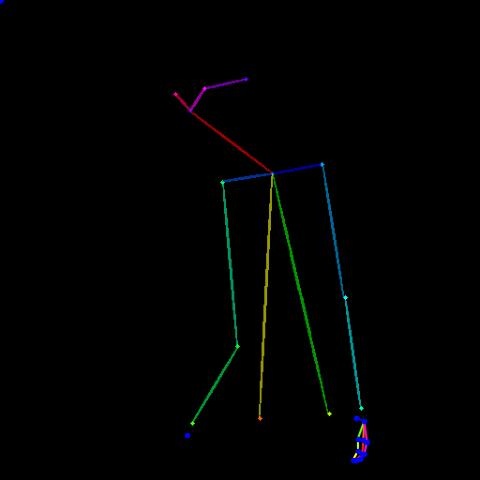}
        \end{subfigure}
        \begin{subfigure}[b]{0.19\linewidth}
            \includegraphics[width=1.0\linewidth]{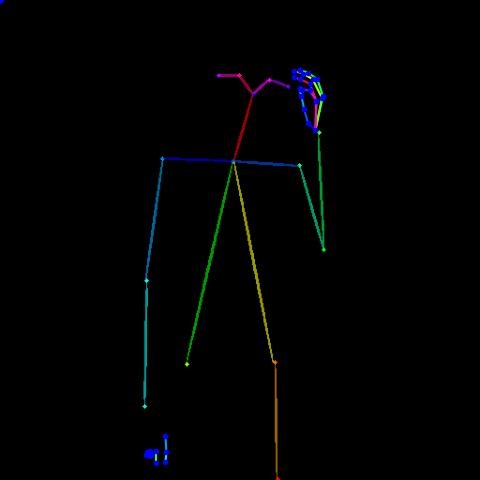}
        \end{subfigure}
        \begin{subfigure}[b]{0.19\linewidth}
            \includegraphics[width=1.0\linewidth]{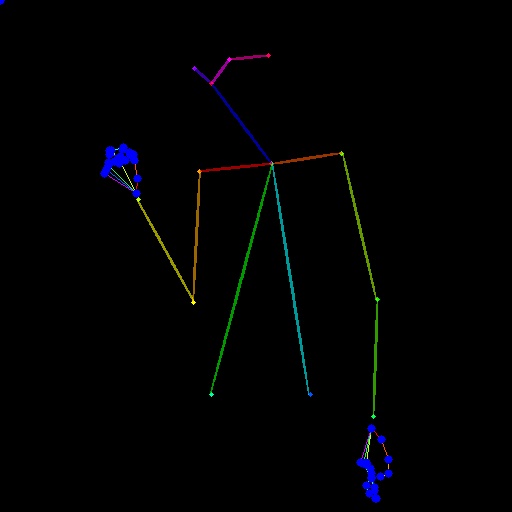}
        \end{subfigure}    
        \begin{subfigure}[b]{0.19\linewidth}
            \includegraphics[width=1.0\linewidth]{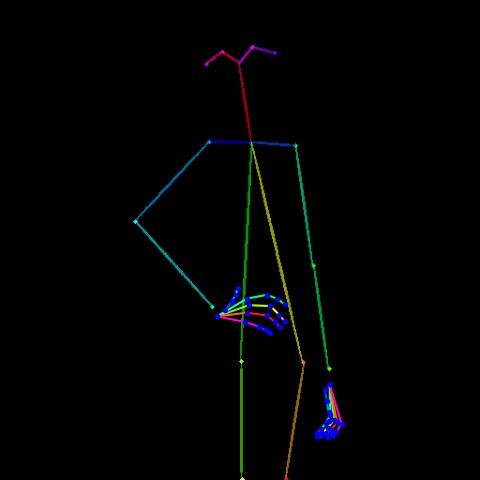}
        \end{subfigure}  
    \end{subfigure} 
\end{subfigure}
\begin{subfigure}[]{0.9\linewidth}
    \centering
    \begin{subfigure}[]{1.0\linewidth}
        \begin{subfigure}[b]{0.19\linewidth}
            \includegraphics[width=1.0\linewidth]{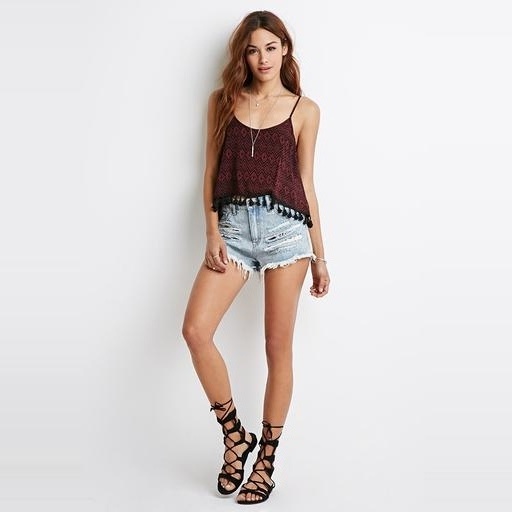}
        \end{subfigure}
        \begin{subfigure}[b]{0.19\linewidth}
            \includegraphics[width=1.0\linewidth]{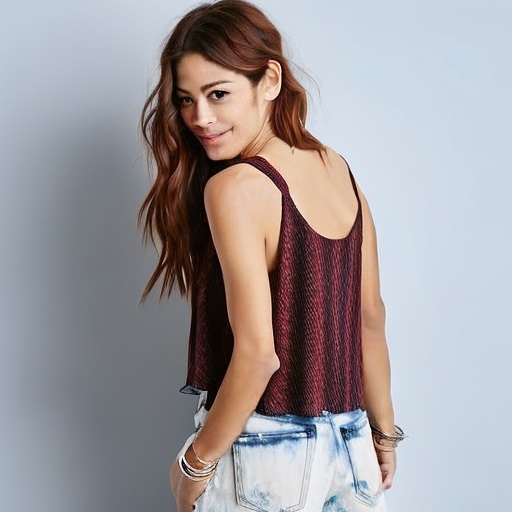}
        \end{subfigure}
        \begin{subfigure}[b]{0.19\linewidth}
            \includegraphics[width=1.0\linewidth]{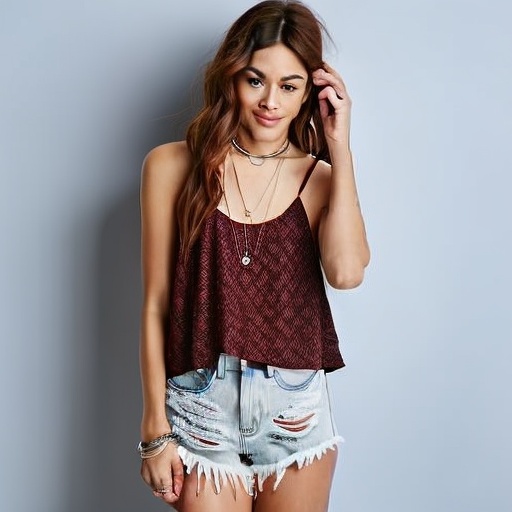}
        \end{subfigure}
        \begin{subfigure}[b]{0.19\linewidth}
            \includegraphics[width=1.0\linewidth]{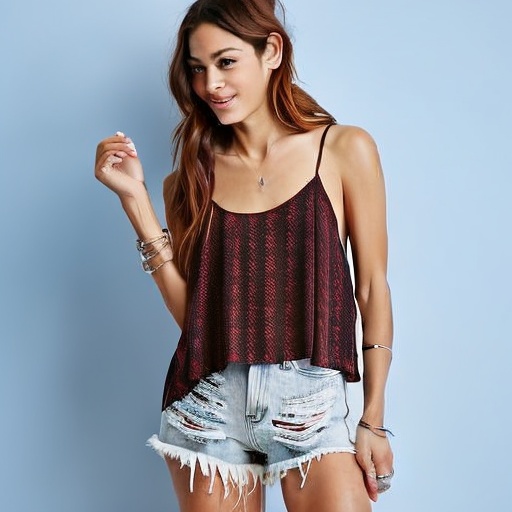}
        \end{subfigure}    
        \begin{subfigure}[b]{0.19\linewidth}
            \includegraphics[width=1.0\linewidth]{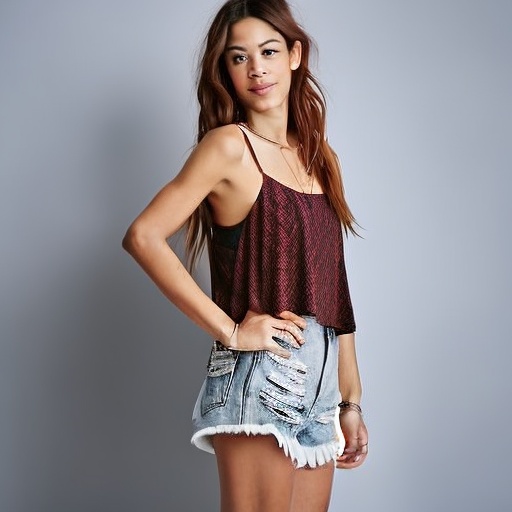}
        \end{subfigure}  
    \end{subfigure} 
\end{subfigure}
\begin{subfigure}[]{0.9\linewidth}
    \centering
    \begin{subfigure}[]{1.0\linewidth}
        \begin{subfigure}[b]{0.19\linewidth}
            \includegraphics[width=1.0\linewidth]{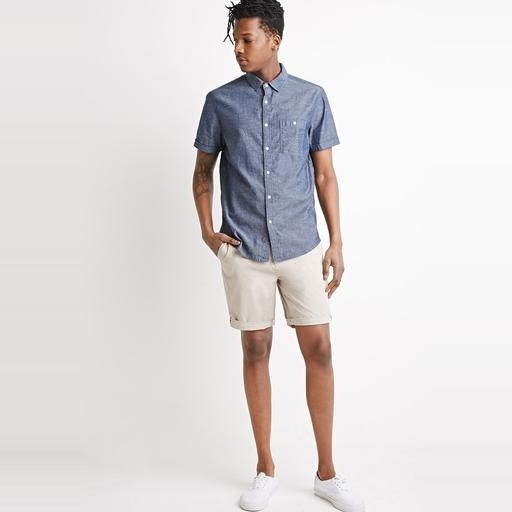}
        \end{subfigure}
        \begin{subfigure}[b]{0.19\linewidth}
            \includegraphics[width=1.0\linewidth]{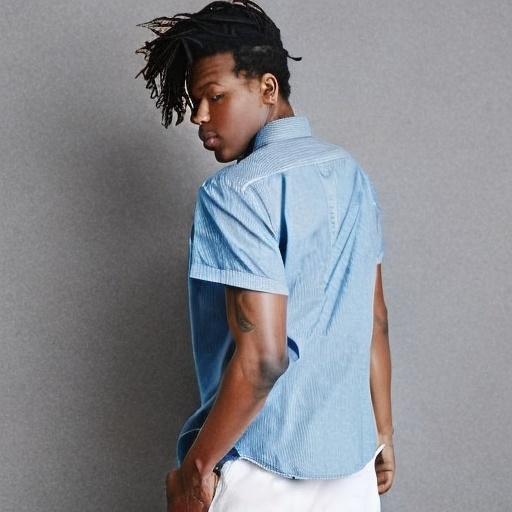}
        \end{subfigure}
        \begin{subfigure}[b]{0.19\linewidth}
            \includegraphics[width=1.0\linewidth]{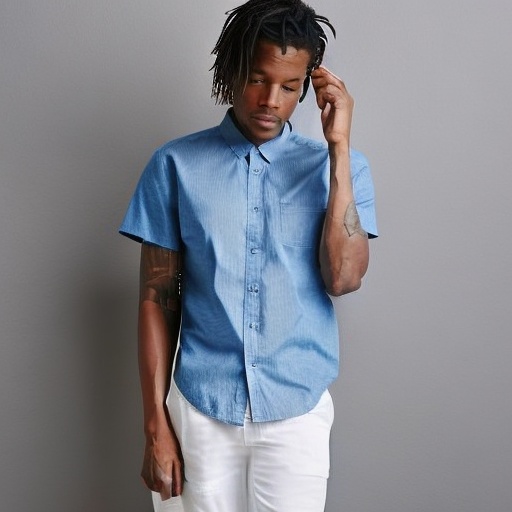}
        \end{subfigure}
        \begin{subfigure}[b]{0.19\linewidth}
            \includegraphics[width=1.0\linewidth]{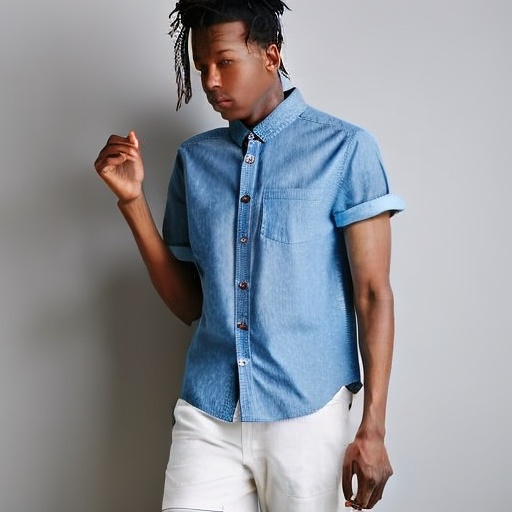}
        \end{subfigure}    
        \begin{subfigure}[b]{0.19\linewidth}
            \includegraphics[width=1.0\linewidth]{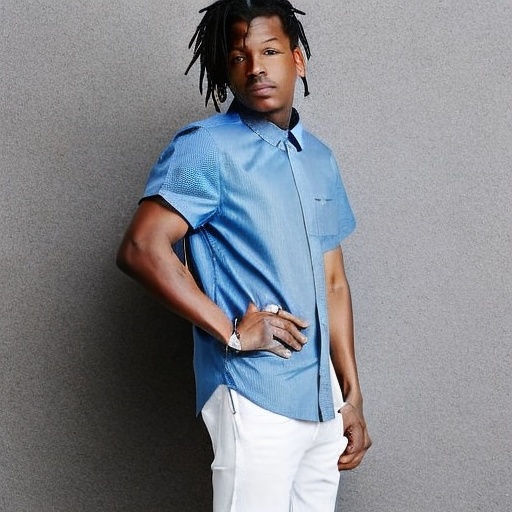}
        \end{subfigure}  
    \end{subfigure} 
\end{subfigure}
\begin{subfigure}[]{0.9\linewidth}
    \centering
    \begin{subfigure}[]{1.0\linewidth}
        \begin{subfigure}[b]{0.19\linewidth}
            \includegraphics[width=1.0\linewidth]{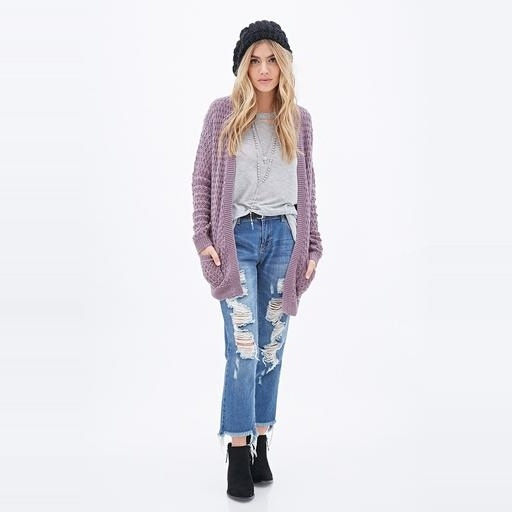}
        \end{subfigure}
        \begin{subfigure}[b]{0.19\linewidth}
            \includegraphics[width=1.0\linewidth]{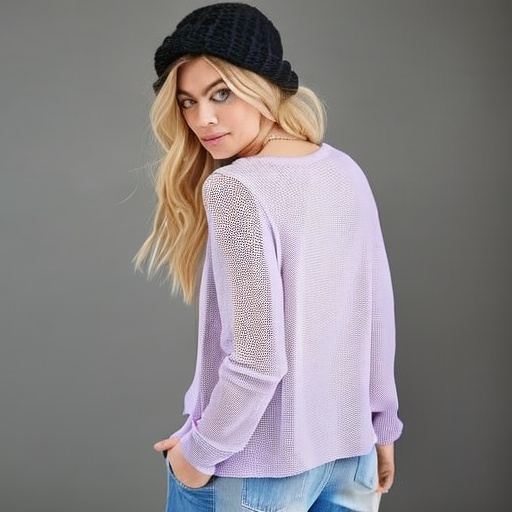}
        \end{subfigure}
        \begin{subfigure}[b]{0.19\linewidth}
            \includegraphics[width=1.0\linewidth]{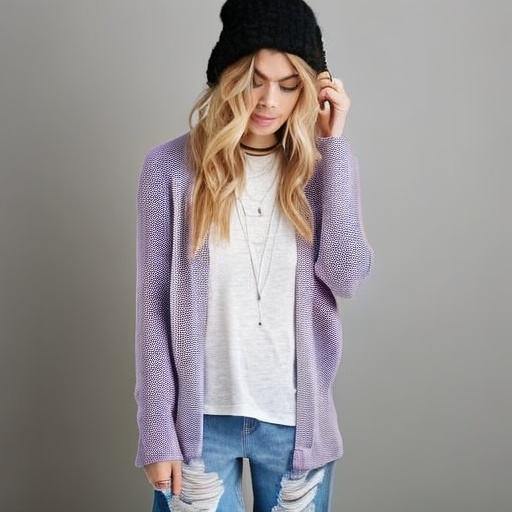}
        \end{subfigure}
        \begin{subfigure}[b]{0.19\linewidth}
            \includegraphics[width=1.0\linewidth]{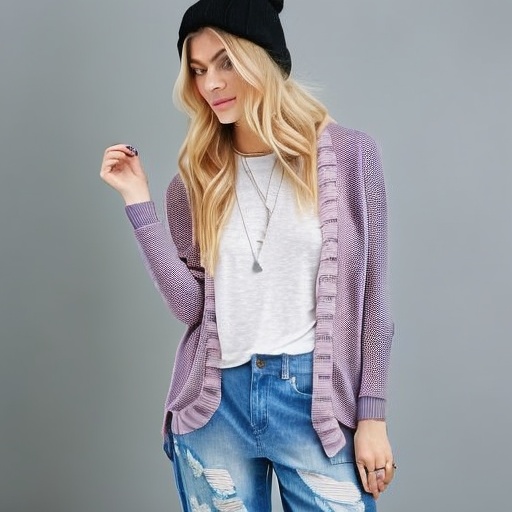}
        \end{subfigure}    
        \begin{subfigure}[b]{0.19\linewidth}
            \includegraphics[width=1.0\linewidth]{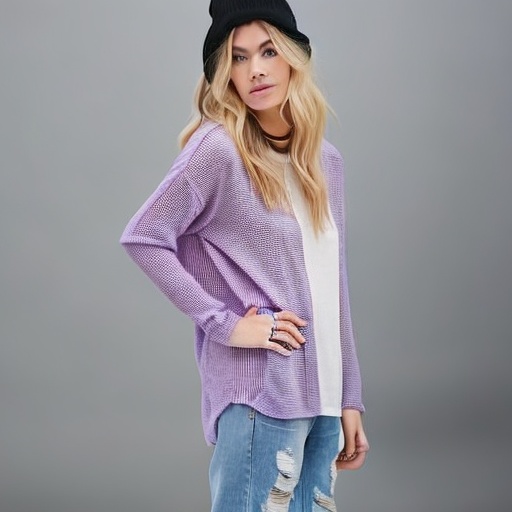}
        \end{subfigure}  
    \end{subfigure} 
\end{subfigure}
\begin{subfigure}[]{0.9\linewidth}
    \centering
    \begin{subfigure}[]{1.0\linewidth}
        \begin{subfigure}[b]{0.19\linewidth}
            \includegraphics[width=1.0\linewidth]{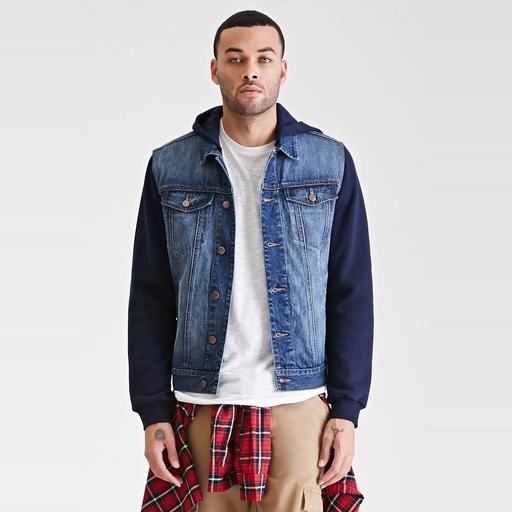}
            \caption{Reference}
        \end{subfigure}
        \begin{subfigure}[b]{0.19\linewidth}
            \includegraphics[width=1.0\linewidth]{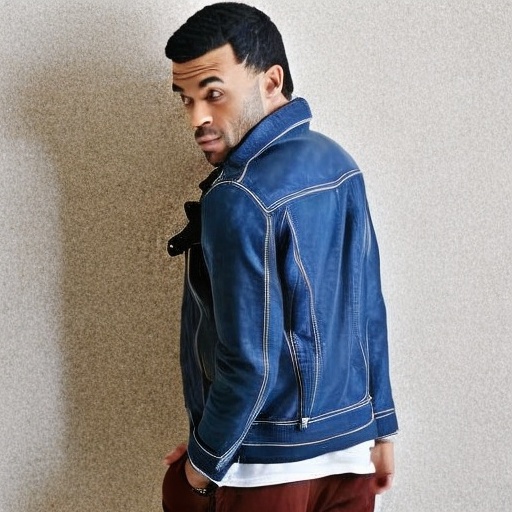}
            \caption{}
        \end{subfigure}
        \begin{subfigure}[b]{0.19\linewidth}
            \includegraphics[width=1.0\linewidth]{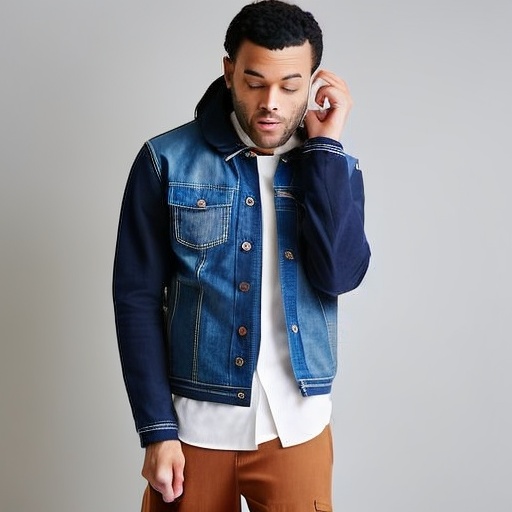}
            \caption{}
        \end{subfigure}
        \begin{subfigure}[b]{0.19\linewidth}
            \includegraphics[width=1.0\linewidth]{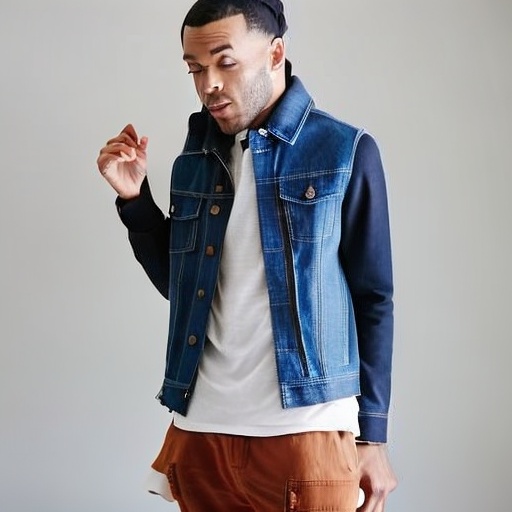}
            \caption{}
        \end{subfigure}    
        \begin{subfigure}[b]{0.19\linewidth}
            \includegraphics[width=1.0\linewidth]{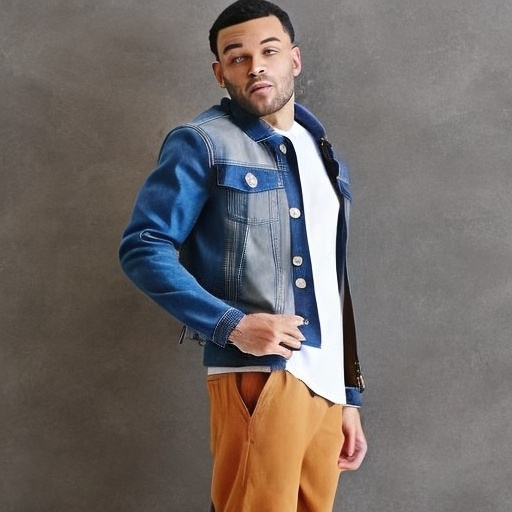}
            \caption{}
        \end{subfigure}  
    \end{subfigure} 
\end{subfigure}
\caption{Pose Transfer from (a) reference person to new poses in (b)-(e)}
\label{fig:supp:pose_transfer}

\end{figure*}

\subsection{Virtual Try-on}
\begin{figure*}[!ht]
\centering
\begin{subfigure}[b]{1.0\linewidth}
    \begin{subfigure}[b]{0.19\linewidth}
        \includegraphics[width=1.0\linewidth]{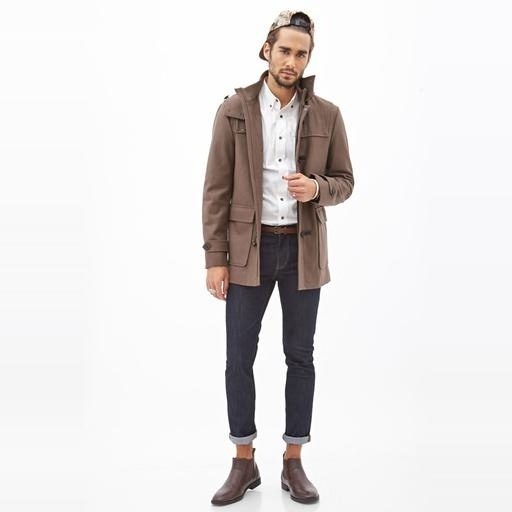}
    \end{subfigure}
    \begin{subfigure}[b]{0.19\linewidth}
        \includegraphics[width=1.0\linewidth]{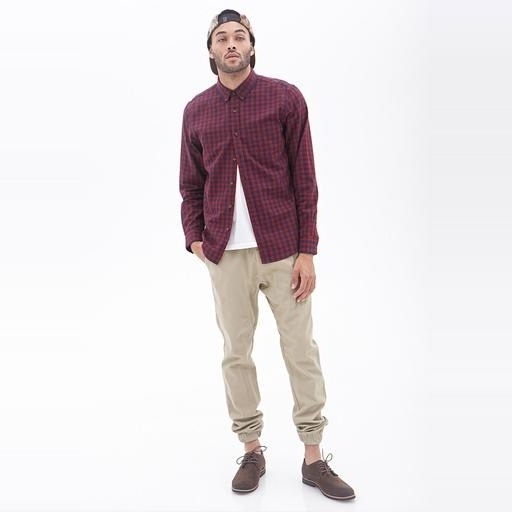}
    \end{subfigure}
    \begin{subfigure}[b]{0.19\linewidth}
        \includegraphics[width=1.0\linewidth]{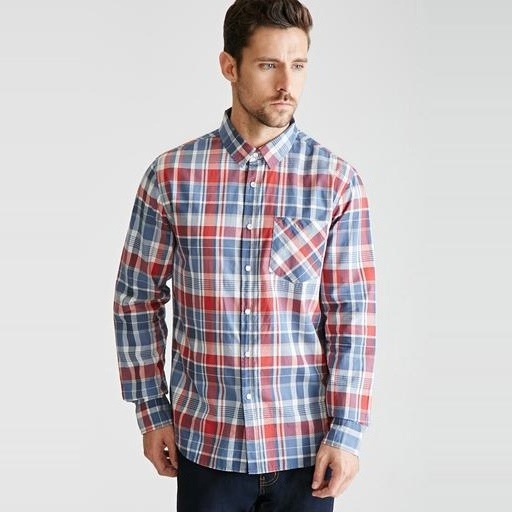}
    \end{subfigure}
    \begin{subfigure}[b]{0.19\linewidth}
        \includegraphics[width=1.0\linewidth]{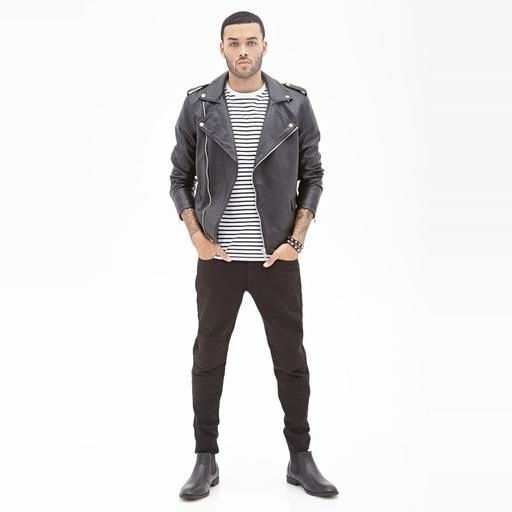}
    \end{subfigure}    
\end{subfigure}    
\begin{subfigure}[b]{1.0\linewidth}
    \begin{subfigure}[b]{0.19\linewidth}
        \includegraphics[width=1.0\linewidth]{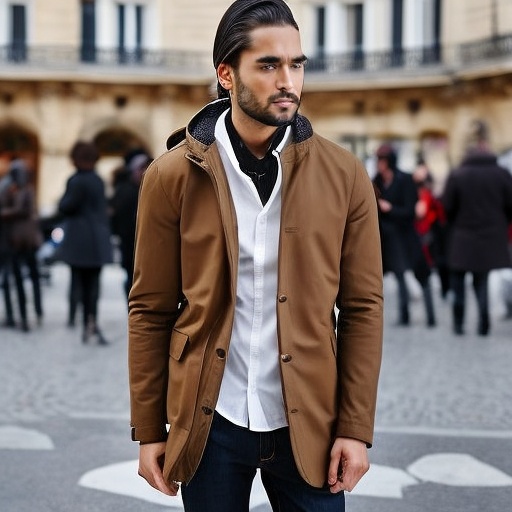}
        \caption{}
    \end{subfigure}
    \begin{subfigure}[b]{0.19\linewidth}
        \includegraphics[width=1.0\linewidth]{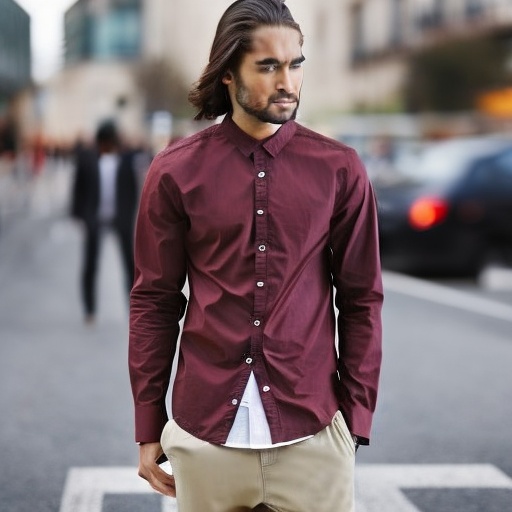}
        \caption{}
    \end{subfigure}
    \begin{subfigure}[b]{0.19\linewidth}
        \includegraphics[width=1.0\linewidth]{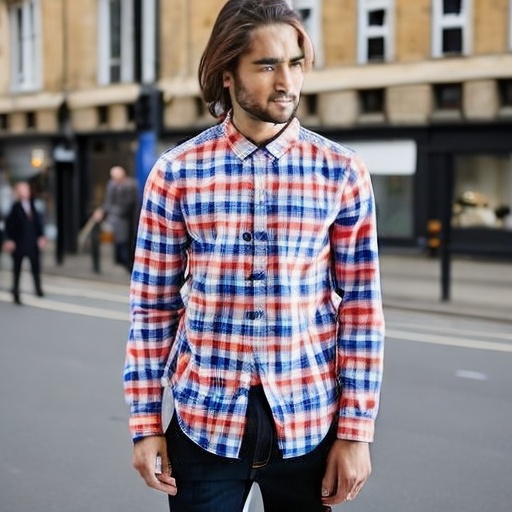}
        \caption{}
    \end{subfigure}
    \begin{subfigure}[b]{0.19\linewidth}
        \includegraphics[width=1.0\linewidth]{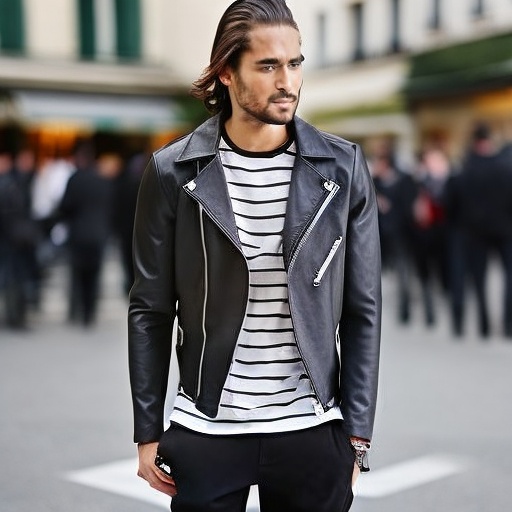}
        \caption{}
    \end{subfigure}
    \begin{subfigure}[b]{0.19\linewidth}
        \includegraphics[width=1.0\linewidth]{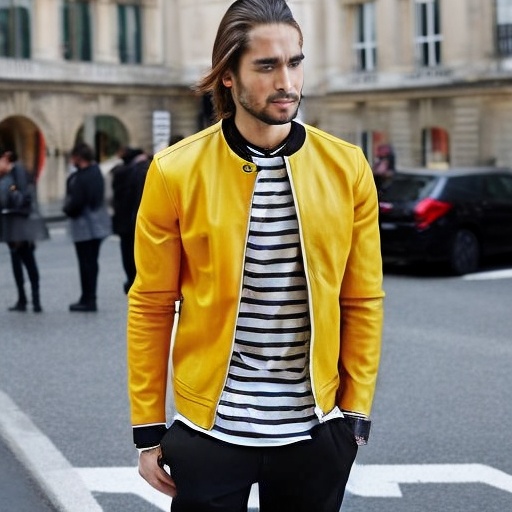}
        \caption{\textit{``yellow  jacket''}}
    \end{subfigure}    
\end{subfigure} 
\caption{High-resolution virtual try-on with real-world background. (Top) reference fashion for visual conditioning. (Bottom): virtual try-on results.}
\label{fig:vtron:1}
\end{figure*}

\begin{figure*}[!ht]
\centering
\begin{subfigure}[b]{1.0\linewidth}
    \centering
    \begin{subfigure}[b]{0.23\linewidth}
        \centering
        \includegraphics[width=0.5\linewidth]{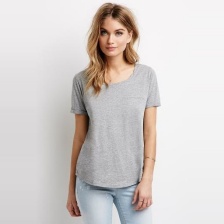}
    \end{subfigure}
    \begin{subfigure}[b]{0.23\linewidth}
        \centering
        \includegraphics[width=0.5\linewidth]{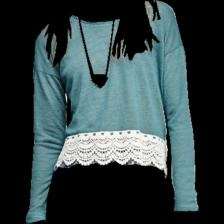}
    \end{subfigure}
    \begin{subfigure}[b]{0.23\linewidth}
        \centering
        \includegraphics[width=0.5\linewidth]{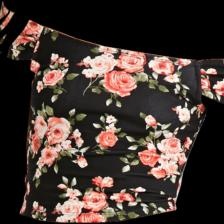}
    \end{subfigure}    
    \begin{subfigure}[b]{0.23\linewidth}
        \centering
        \includegraphics[width=0.5\linewidth]{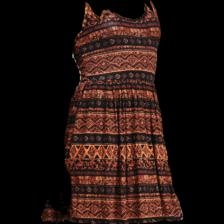}
    \end{subfigure}      
\end{subfigure}    
\vspace{5mm}
\begin{subfigure}[b]{1.0\linewidth}
    \centering
    \begin{subfigure}[b]{0.23\linewidth}
        \includegraphics[width=1.0\linewidth]{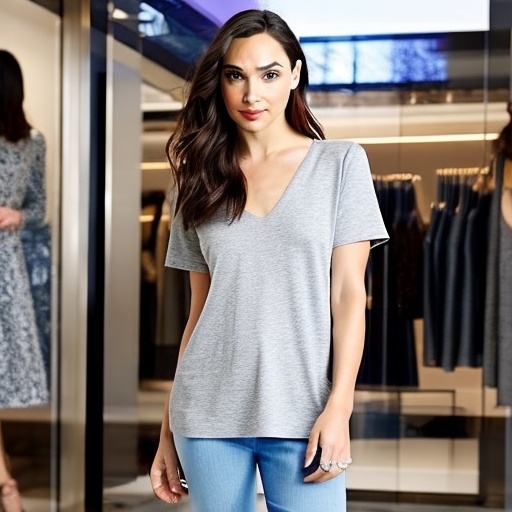}
    \end{subfigure}
    \begin{subfigure}[b]{0.23\linewidth}
        \includegraphics[width=1.0\linewidth]{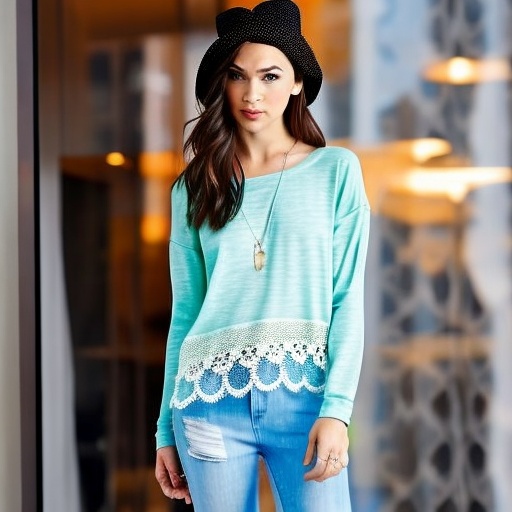}
    \end{subfigure}
    \begin{subfigure}[b]{0.23\linewidth}
        \includegraphics[width=1.0\linewidth]{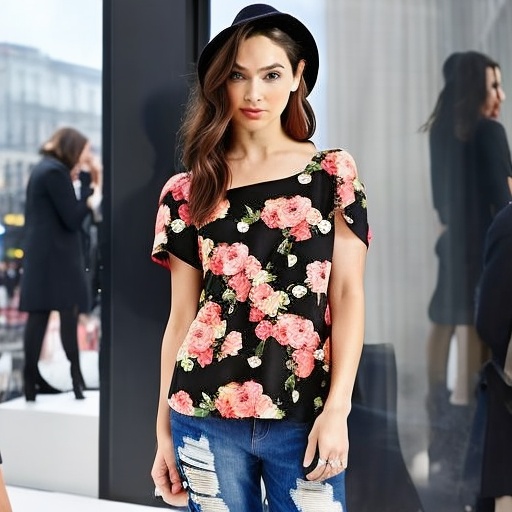}
    \end{subfigure}    
    \begin{subfigure}[b]{0.23\linewidth}
        \includegraphics[width=1.0\linewidth]{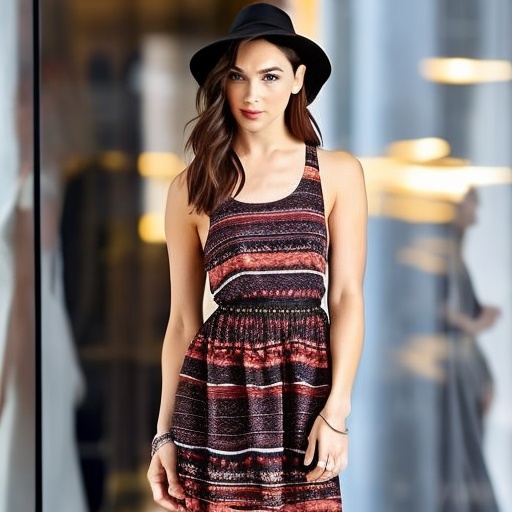}
    \end{subfigure}
\end{subfigure}
\begin{subfigure}[b]{1.0\linewidth}
    \centering
    \begin{subfigure}[b]{0.23\linewidth}
        \centering
        \includegraphics[width=0.5\linewidth]{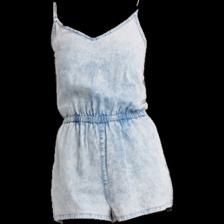}
    \end{subfigure}    
    \begin{subfigure}[b]{0.23\linewidth}
        \centering
        \includegraphics[width=0.5\linewidth]{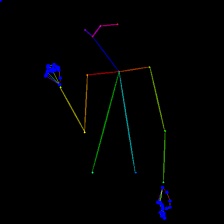}
    \end{subfigure} 
    \begin{subfigure}[b]{0.23\linewidth}
        \centering
        \includegraphics[width=0.5\linewidth]{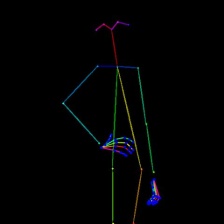}
    \end{subfigure}
    \begin{subfigure}[b]{0.23\linewidth}
        \centering
        \includegraphics[width=0.5\linewidth]{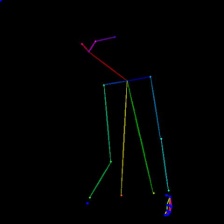}
    \end{subfigure}    
\end{subfigure}    
\begin{subfigure}[b]{1.0\linewidth}
    \centering
    \begin{subfigure}[b]{0.23\linewidth}
        \includegraphics[width=1.0\linewidth]{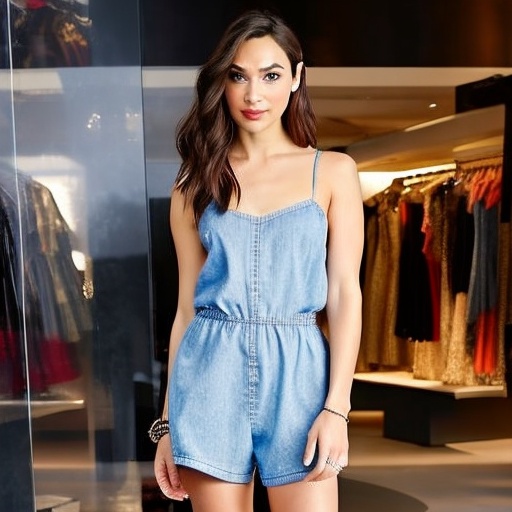}
    \end{subfigure}
    \begin{subfigure}[b]{0.23\linewidth}
        \includegraphics[width=1.0\linewidth]{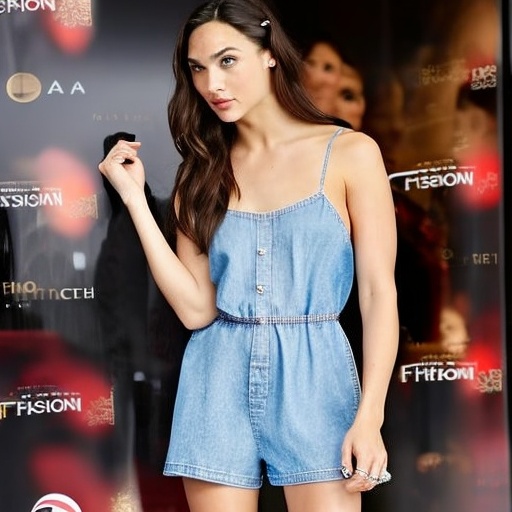}
    \end{subfigure}    
    \begin{subfigure}[b]{0.23\linewidth}
        \includegraphics[width=1.0\linewidth]{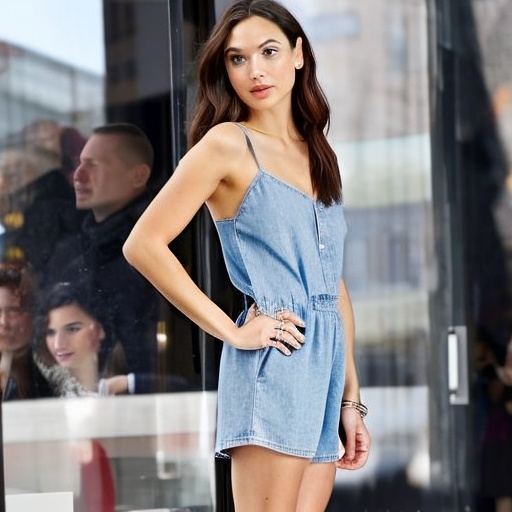}
    \end{subfigure}  
    \begin{subfigure}[b]{0.23\linewidth}
        \includegraphics[width=1.0\linewidth]{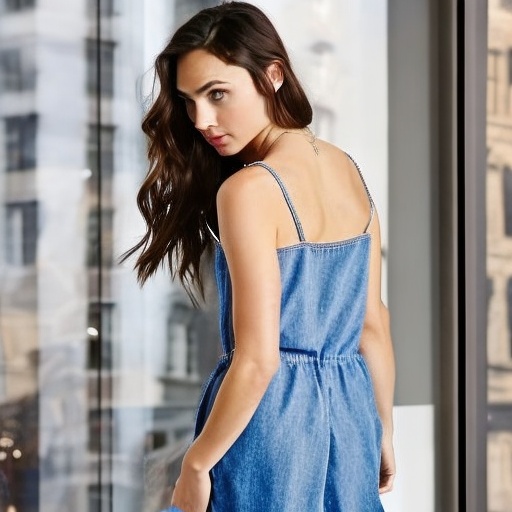}
    \end{subfigure}      
\end{subfigure} 
\caption{Combining re-identification, virtual try-on, and pose re-target, we showcase examples of posing fashion with celebrity avatars.}
\label{fig:vtron:gadot}
\end{figure*}

Figure \ref{fig:vtron:1} demonstrates how we perform fashion virtual try-on using visual and text prompts. Figure \ref{fig:vtron:gadot} illustrates the culmination of our methods, showcasing the seamless integration of re-identification, virtual try-on, and pose re-target.

\clearpage

\section{Quantitative Result}
\label{sec:supp:quantitative}
\subsection{Section 4.1: Mode Collapse Quantitative Result}

Table \ref{table:strength} shows the quantitative results corresponding to Figure 6 in Section 4.1 - Mode Collapse and Control Strength. Our method produces a higher CLIP score than the baseline at various control strengths, indicating less mode collapse. This is more evident in CLIP accuracy; at control strength 0.5, we achieve 100\% (or 0\% MCR) while baselines have only 46\% and 63\% ControlNet and IP-Adapter, respectively.  

\begin{table}
\centering
    \begin{adjustbox}{width=0.8\columnwidth}
    \begin{tabular}{l |cccccccc}
    \toprule
    Strength & 0.0  & 0.2 & 0.3 & 0.4 & 0.5 & 0.6 & 0.8 & 1.0 \\
    \midrule
    \multicolumn{1}{c|}{} & \multicolumn{8}{c}{\underline{CLIP score}} \\
    ControlNet & 0.2720 & 0.2620 & 0.2440 & 0.2340 & 0.2300 & 0.2300 & \textbf{0.2240} & \textbf{0.2260} \\
    IP-Adapter & \textbf{0.2900} & 0.2920 & 0.2780 & 0.2620 & 0.2360 & 0.2120 & 0.1780 & 0.1900\\ 
    ViscoNet(Ours) & 0.2860 & \textbf{0.2940} & \textbf{0.2920} & \textbf{0.2900} & \textbf{0.2800} & \textbf{0.2660} & \textbf{0.2420} & 0.2220 \\
    \midrule
    \multicolumn{1}{c|}{} & \multicolumn{8}{c}{\underline{CLIP accuracy}} \\
    ControlNet & 0.8660 & 0.7720 & 0.7000 & 0.6180 & 0.4620 & 0.5500 & 0.5020 & \textbf{0.5760} \\
    IP-Adapter & 0.9800 & 0.9700 & 0.8800 & 0.7500 & 0.6300 & 0.4100 & 0.1500 & 0.2100\\        
    ViscoNet(Ours) &  \textbf{1.0000} & \textbf{1.0000} & \textbf{1.0000} & \textbf{1.0000} & \textbf{1.0000} & \textbf{0.9000} & \textbf{0.7000} & \textbf{0.5760} \\
    \midrule
    \multicolumn{1}{c|}{} & \multicolumn{8}{c}{\underline{Pose accuracy (OKS)}} \\
    
    ControlNet & 0.0880 & 0.4139 & \textbf{0.6610} & \textbf{0.8305} & \textbf{0.8596} & \textbf{0.8852} & \textbf{0.9223} & \textbf{0.9348}\\
    IP-Adapter & \textbf{0.5379} & \textbf{0.5412} & 0.6060 & 0.6813 & 0.7546 & 0.8074 & 0.9010 & 0.9298\\
    ViscoNet(Ours) & 0.0446 & 0.1654 & 0.3869 & 0.6580 & 0.7845 & 0.8253 & 0.8824 & 0.9102 \\
    \bottomrule
    \end{tabular}        
    \end{adjustbox}
    \vspace{3mm}
    \caption{Reduced control strength results in higher CLIP scores and accuracy, translating to less mode collapse.}
    \vspace{3mm}
    \label{table:strength}
\end{table}

\begin{figure}
    \centering
    \includegraphics[width=1\linewidth]{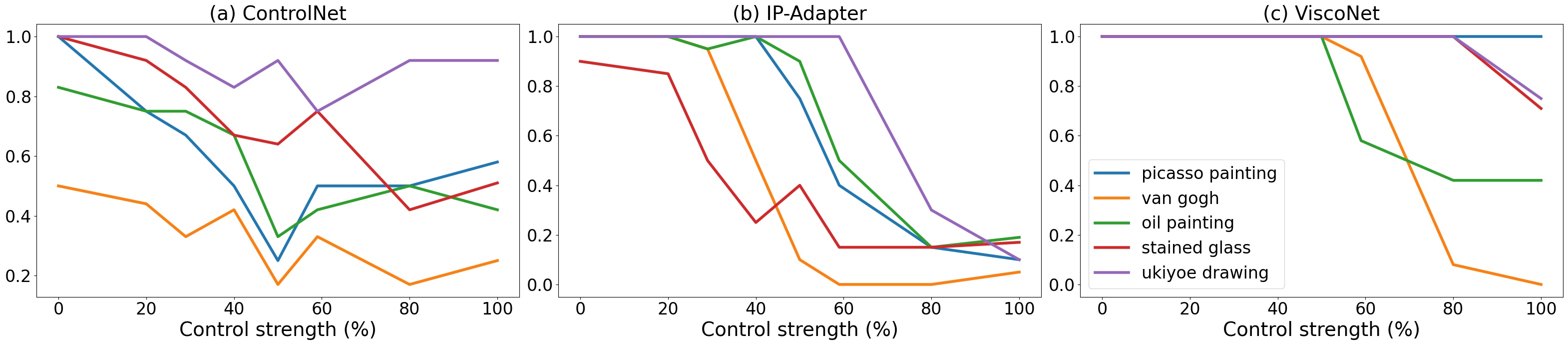}
    \caption{CLIP accuracy - comparing different image styles.}
    \label{fig:appendix:mc_styles}
\end{figure}

Figure \ref{fig:appendix:mc_styles} shows the breakdown of CLIP accuracy across the image styles in Table \ref{table:strength}. Based on the same Stable Diffusion model, all models have shown the highest mode collapse rate in Van Gogh's painting style, while Ukiyoe is the least affected. 

\clearpage
\subsection{Section 4.3: Human Evaluation Result}

\begin{figure}[!ht]
    \centering
    \includegraphics[width=0.8\linewidth]{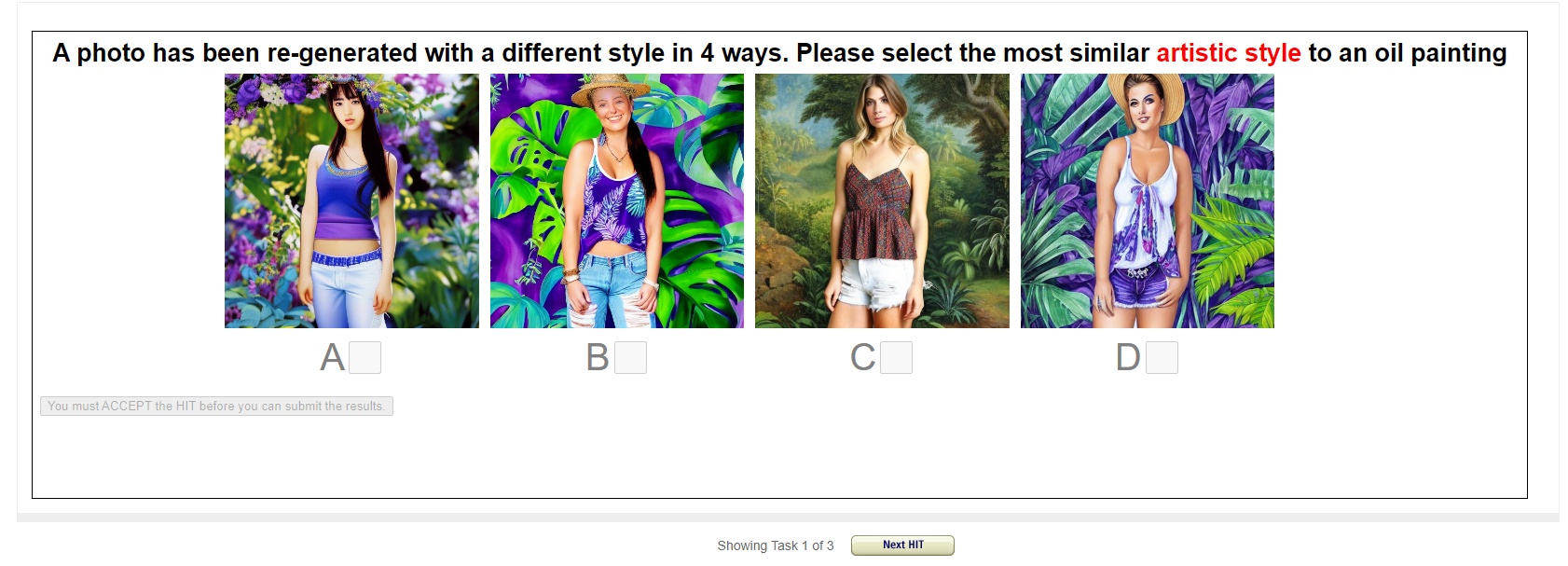}
    \caption{Screenshot of user study presented to users for evaluating the quality of the stylization against the three baselines.}
    \label{fig:mturk}
\end{figure}

We further conducted a more extensive scale user study on Amazon Mechanical Turk (AMT) to measure the real-life preferences between our model and the HIG baseline approaches. We perform a 4-way comparison, asking workers to select their best preference from randomly shuffled samples, as shown in Figure~\ref{fig:mturk}. 

\begin{table}[!ht]
\begin{center}
    \begin{adjustbox}{width=0.6\textwidth}
        \begin{tabular}{l |cccc|c }
        \toprule
        \multicolumn{1}{c|}{} & \multicolumn{5}{c}{Human Evaluation }  \\
        \toprule
         Image Styles & HumanSD & ControlNet & T2I-Adapter & \textbf{ViscoNet (Ours)}  & Ours (\%)\\
         \midrule
          Ukiyoe & 27 & 32 & 4 & \textbf{37}  & \textbf{37\%} \\
         Cyberpunk anime & 23 & 13 & 21 & \textbf{41} & \textbf{41\%} \\
         Stained glass & 0 & 32 & 23 & \textbf{45} & \textbf{45\%}\\
         Van Gogh & 2 & 13 & 9 & \textbf{76}  & \textbf{76\%}\\
         Picasso &  0 & 13 & 42 & \textbf{45} & \textbf{45\%}\\
         Oil Painting &  9 & 11 & 7 & \textbf{73} & \textbf{73\%} \\
         Disney &  5 & 23 & 5 & \textbf{67}  & \textbf{67\%}\\
         \midrule
         Total & 77  & 139  & 111  & \textbf{384} &  \\
         Average & 9.43\%  & 19.9\%  & 15.9\%  & \textbf{54.9\%} &   \\
         \bottomrule
         \end{tabular}
    \end{adjustbox}
    \caption{Our method scores the highest in human evaluation, proving its ability to generate good-quality, diverse image styles.}
\end{center}
\vspace{-7mm}
\end{table}

\end{document}